\documentclass[%
  preprint%
  ]{elsarticle}
\newif\iftodo

\usepackage[T1]{fontenc}

\usepackage{lipsum}

\usepackage{algorithm}
\usepackage[noend]{algpseudocode}

\usepackage{fancyvrb}

\usepackage{scalerel}

\usepackage{xspace}

\usepackage{hyperref}

\usepackage{amssymb,amsmath,amsthm,mathtools,proof}
\usepackage{thmtools}
\usepackage{thm-restate}
\usepackage{bm,bbm}

\usepackage{geometry}
\usepackage{wrapfig}

\usepackage[capitalise]{cleveref}
\crefname{equation}{}{}

\usepackage[usenames]{xcolor}

\usepackage[authormarkup=none,defaultcolor=blue,draft]{changes}
\definechangesauthor[color=red]{1}

\usepackage{subcaption}
\usepackage{booktabs}
\usepackage{multirow}

\usepackage[export]{adjustbox}

\usepackage{tikz}
\usetikzlibrary{decorations,arrows,shapes,automata}
\usetikzlibrary{shapes,backgrounds,mindmap,trees,arrows.meta,quotes,calc,tikzmark,fit}
\tikzset{fit margins/.style={/tikz/afit/.cd,#1,
    /tikz/.cd,
    inner xsep=\pgfkeysvalueof{/tikz/afit/left}+\pgfkeysvalueof{/tikz/afit/right},
    inner ysep=\pgfkeysvalueof{/tikz/afit/top}+\pgfkeysvalueof{/tikz/afit/bottom},
    xshift=-\pgfkeysvalueof{/tikz/afit/left}+\pgfkeysvalueof{/tikz/afit/right},
    yshift=-\pgfkeysvalueof{/tikz/afit/bottom}+\pgfkeysvalueof{/tikz/afit/top}},
    afit/.cd,left/.initial=2pt,right/.initial=2pt,bottom/.initial=2pt,top/.initial=2pt
}

\newcommand{\ak}{A\"it-Kaci\xspace}
\newcommand{\OSF}{Order-Sorted Feature\xspace}
\newcommand{\bpl}{Bousi$\sim$Prolog\xspace}

\newcommand{\osf}{OSF\xspace}

\newcommand{\red}[1]{\textcolor{red}{#1}}
\newcommand{\black}[1]{\textcolor{black}{#1}}
\newcommand{\bluebf}[1]{\textcolor{blue}{\textbf{#1}}}

\makeatletter
\newcommand{\mymathcolor}[2]{%
  \begingroup
  \colorlet{out}{.}\color{#1}#2\@ifnextchar_{\do@mathcolorsub}{\endgroup}%
}
\newcommand{\do@mathcolorsub}[2]{%
  _{{\color{out}#2}}\@ifnextchar^{\do@mathcolorsup}{\endgroup}%
}
\newcommand{\do@mathcolorsup}[2]{%
  ^{\color{out}#2}\endgroup
}
\makeatother

\newcommand\sortcol[1]{%
  \ifcolors \textcolor{sortcol}{#1}%
  \else     {#1} \fi
}
\newcommand\strcol[1]{%
  \ifcolors \textcolor{gray}{#1}%
  \else     {#1} \fi
}
\newcommand\natcol[1]{%
  \ifcolors \textcolor{red!60!black}{#1}%
  \else     {#1} \fi
}
\newcommand\tagcol[1]{%
  \ifcolors \textcolor{tagcol}{#1}%
  \else     {#1} \fi
}
\newcommand\featcol[1]{%
  \ifcolors \textcolor{featcol}{#1}%
  \else     {#1} \fi
}

\newcommand{\movie}{\s[movie]}
\newcommand{\person}{\s[person]}
\newcommand{\director}{\s[director]}
\newcommand{\strng}{\s[string]}
\newcommand{\horror}{\s[horror]}

\newcommand{\thriller}{\s[thriller]}
\newcommand{\slasher}{\s[slasher]}
\newcommand{\ttle}{\f[title]}
\newcommand{\dirby}{\f[directed\_by]}

\newcommand{\psycho}{\obj{psycho}}
\newcommand{\halloween}{\obj{halloween}}
\newcommand{\carpenter}{\obj{carpenter}}
\newcommand{\hitchcock}{\obj{hitchcock}}
\newcommand{\appdegree}{0.5}

\CustomVerbatimCommand{\inCodeStub}{Verb}{commandchars=\\\{\}}

\DeclareFontFamily{U} {MnSymbolA}{}
\DeclareFontShape{U}{MnSymbolA}{m}{n}{
  <-6> MnSymbolA5
  <6-7> MnSymbolA6
  <7-8> MnSymbolA7
  <8-9> MnSymbolA8
  <9-10> MnSymbolA9
  <10-12> MnSymbolA10
  <12-> MnSymbolA12}{}
\DeclareFontShape{U}{MnSymbolA}{b}{n}{
  <-6> MnSymbolA-Bold5
  <6-7> MnSymbolA-Bold6
  <7-8> MnSymbolA-Bold7
  <8-9> MnSymbolA-Bold8
  <9-10> MnSymbolA-Bold9
  <10-12> MnSymbolA-Bold10
  <12-> MnSymbolA-Bold12}{}
\DeclareSymbolFont{MnSyA} {U} {MnSymbolA}{m}{n}

\DeclareFontFamily{U} {MnSymbolC}{}
\DeclareFontShape{U}{MnSymbolC}{m}{n}{
  <-6> MnSymbolC5
  <6-7> MnSymbolC6
  <7-8> MnSymbolC7
  <8-9> MnSymbolC8
  <9-10> MnSymbolC9
  <10-12> MnSymbolC10
  <12-> MnSymbolC12}{}
\DeclareFontShape{U}{MnSymbolC}{b}{n}{
  <-6> MnSymbolC-Bold5
  <6-7> MnSymbolC-Bold6
  <7-8> MnSymbolC-Bold7
  <8-9> MnSymbolC-Bold8
  <9-10> MnSymbolC-Bold9
  <10-12> MnSymbolC-Bold10
  <12-> MnSymbolC-Bold12}{}
\DeclareSymbolFont{MnSyC} {U} {MnSymbolC}{m}{n}

\DeclareFontFamily{U} {MnSymbolF}{}
\DeclareFontShape{U}{MnSymbolF}{m}{n}{
  <-6> MnSymbolF5
  <6-7> MnSymbolF6
  <7-8> MnSymbolF7
  <8-9> MnSymbolF8
  <9-10> MnSymbolF9
  <10-12> MnSymbolF10
  <12-> MnSymbolF12}{}
\DeclareFontShape{U}{MnSymbolF}{b}{n}{
  <-6> MnSymbolF-Bold5
  <6-7> MnSymbolF-Bold6
  <7-8> MnSymbolF-Bold7
  <8-9> MnSymbolF-Bold8
  <9-10> MnSymbolF-Bold9
  <10-12> MnSymbolF-Bold10
  <12-> MnSymbolF-Bold12}{}
\DeclareSymbolFont{MnSyF} {U} {MnSymbolF}{m}{n}

\newcommand{\bigcdot}{\boldsymbol{\cdot}}
\DeclareMathSymbol{\oast}{\mathbin}{MnSyC}{101}
\DeclareMathSymbol{\upsquigarrow}{\mathrel}{MnSyA}{161}
\DeclareMathSymbol{\downsquigarrow}{\mathrel}{MnSyA}{163}

\DeclareMathSymbol{\cap}{\mathbin}{MnSyC}{57}
\DeclareMathSymbol{\cup}{\mathbin}{MnSyC}{56}
\let\bigcap\relax
\DeclareMathSymbol{\bigcap}{\mathop}{MnSyF}{29}
\let\bigcup\relax
\DeclareMathSymbol{\bigcup}{\mathop}{MnSyF}{31}

\DeclareMathSymbol{\capdot}{\mathbin}{MnSyC}{61}
\DeclareMathSymbol{\cupdot}{\mathbin}{MnSyC}{60}
\DeclareMathSymbol{\bigcapdot}{\mathop}{MnSyF}{33}
\DeclareMathSymbol{\bigcupdot}{\mathop}{MnSyF}{35}

\DeclareMathSymbol{\curlywedge}{\mathbin}{MnSyC}{50}
\DeclareMathSymbol{\curlyvee}{\mathbin}{MnSyC}{51}
\DeclareMathSymbol{\bigcurlywedge}{\mathop}{MnSyF}{17}
\DeclareMathSymbol{\bigcurlyvee}{\mathop}{MnSyF}{19}

\DeclareMathSymbol{\curlywedgedot}{\mathbin}{MnSyC}{52}
\DeclareMathSymbol{\curlyveedot}{\mathbin}{MnSyC}{53}
\DeclareMathSymbol{\bigcurlywedgedot}{\mathop}{MnSyF}{21}
\DeclareMathSymbol{\bigcurlyveedot}{\mathop}{MnSyF}{23}

\makeatletter
\providecommand{\leftsquigarrow}{%
  \mathrel{\mathpalette\reflect@squig\relax}%
}
\newcommand{\reflect@squig}[2]{%
  \reflectbox{$\m@th#1\rightsquigarrow$}%
}
\makeatother

\input{preamble/fuzzy}
\DeclarePairedDelimiter\fceil
  {\lceil}
  {\rceil^{\mkern-3.5mu\bigcdot}}
\DeclarePairedDelimiter\ffloor
  {\lfloor}
  {\rfloor_{\mkern-3.5mu\bigcdot}}

\newcommand{\pow}{\mathcal{P}}

\newcommand{\card}[1]{\left|#1\right|}

\makeatletter
\newcommand{\defeq}{\mathrel{\aban@defeq}}
\newcommand{\aban@defeq}{%
  \vbox{\offinterlineskip\check@mathfonts
    \ialign{\hfil##\hfil\cr
      \fontsize{\ssf@size}{\z@}\normalfont def\cr
      \noalign{\kern1\p@}
      $\m@th=$\cr
      \noalign{\kern-.5\fontdimen22\textfont2}
    }%
  }%
}
\makeatother

\newcommand{\oT}{\Leftarrow}

\newcommand{\To}{\Rightarrow}
\newcommand{\Iff}{\Leftrightarrow}

\makeatletter
\newcommand{\defiff}{\mathrel{\aban@defiff}}
\newcommand{\aban@defiff}{%
  \vbox{\offinterlineskip\check@mathfonts
    \ialign{\hfil##\hfil\cr
      \fontsize{\ssf@size}{\z@}\normalfont def\cr
      \noalign{\kern2.5\p@}
      $\m@th\Iff$\cr
      \noalign{\kern-.5\fontdimen22\textfont2}
    }%
  }%
}
\makeatother

\newcommand\restr[2]{{%
  \left.\kern-\nulldelimiterspace %
  #1 %
  \vphantom{\big|} %
  \right|_{#2} %
}}
\DeclareMathOperator*{\bigosfwith}{\scalerel*{\&}{\sum}}

\input{preamble/misc}

\newcommand{\id}[1][X]{\mathit{id}_{#1}}
\newcommand{\cfun}[1][X]{\bm{1}_{#1}}
\newcommand{\isa}{\preceq}
\newcommand{\synisa}{\trianglelefteq}

\newcommand{\isop}{\mathop{\isa}}
\newcommand{\meet}{\curlywedge}

\newcommand{\bigmeet}{\bigcurlywedge}

\NewDocumentCommand{\poset}{ O{\isa} O{X} }{\ensuremath{(#2, #1)}}

\newcommand{\lb}[1]{{#1}^{\mathit{l}}}

\newcommand{\eq}{\approx}
\NewDocumentCommand{\eqclass}{ O{\eq} O{x} }{[#2]_{#1}}

\NewDocumentCommand{\quot}{ O{\eq} O{X} }{\ensuremath{{#2}_{/{#1}}}}

\newif\ifcolors%
\newcommand{\with}{\&}
\renewcommand{\S}{\sortcol{\mathcal{S}}}
\newcommand{\F}{\featcol{\mathcal{F}}}
\newcommand{\V}{\tagcol{\mathcal{V}}}

\newcommand{\fss}{(\S, \F, \fisa)}
\newcommand{\tax}{\poset[\isa][\S]}

\newcommand\rtag{\mathop{\mathit{RootTag}}}

\definecolor{tagcol}{rgb}{0.72, 0.1, 0.26}
\definecolor{sortcol}{rgb}{0.06, 0.22, 0.61}
\definecolor{featcol}{rgb}{0.0, 0.45, 0.0}
\definecolor{relcol}{rgb}{0.8, 0.52, 0.0}
\definecolor{stringcol}{rgb}{0.1, 0.1, 0.1}

\newcommand{\s}[1][s]{\sortcol{#1}}
\newcommand{\su}[1][s]{\s[s']}
\newcommand{\cc}[1][c]{\s[c]}

\newcommand{\si}[1][1]{\s[s_{#1}]}
\newcommand{\sii}[1][1]{\s_{\sortcol{#1}}^\ii}
\newcommand{\sgg}[1][1]{\s_{\sortcol{#1}}^\gg}
\newcommand{\sic}[1][1]{\s_{\sortcol{#1}}^{\ii[D]}}
\newcommand{\fcl}[1][D]{\F^*(#1)}

\newcommand{\str}[1][s]{\strcol{\text{\lstinline|#1|}}}
\newcommand{\nat}[1][s]{\natcol{\text{\lstinline|#1|}}}
\newcommand{\bots}{\s[\bm{\bot}]}
\newcommand{\tops}{\s[\bm{\top}]}

\newcommand{\f}[1][f]{\featcol{#1}}
\newcommand{\fu}{\featcol{f'}}
\newcommand{\fj}[1][1]{\f[f_{#1}]}
\newcommand{\fjp}[1][1]{\f[f'_{#1}]}
\newcommand{\fji}[1][1]{\mymathcolor{featcol}{f}_{\mymathcolor{featcol}{#1}}^\ii}
\newcommand{\fjj}[1][1]{\mymathcolor{featcol}{f}_{\mymathcolor{featcol}{#1}}^\jj}
\newcommand{\fjg}[1][1]{\mymathcolor{featcol}{f}_{\mymathcolor{featcol}{#1}}^\gg}

\newcommand{\fto}{\xrightarrow{\f}}
\newcommand{\fti}[1][1]{%
    \ensuremath{\f[f_{#1}] \to t_{#1}}
}
\newcommand{\X}[1][X]{\tagcol{#1}\xspace}
\newcommand{\Y}[1][Y]{\X[#1]}
\newcommand{\Z}[1][Z]{\X[#1]}
\newcommand{\Xj}[1][1]{\tagcol{X_{#1}}}
\newcommand{\Yj}[1][1]{\tagcol{Y_{#1}}}
\newcommand{\Zj}[1][1]{\tagcol{Z_{#1}}}
\NewDocumentCommand{\xs}{O{X} O{s}}{%
    \ensuremath{\tagcol{#1}:\sortcol{#2}}
}
\newcommand{\osfterm}{%
    \ensuremath{\xs(\fti, \ldots, \fti[n])}
}

\newcommand{\osfwith}{\;\with\;}
\NewDocumentCommand{\osfce}{O{\X} O{\X[X']}}{%
    \ensuremath{#1 \doteq #2}
}

\newcommand{\nullv}{\ensuremath{\mathit{null}}\xspace}
\NewDocumentCommand{\osfcf}{O{\X} O{\f} O{\X[X']}}{%
    \ensuremath{#1.#2 \doteq #3}
}

\newcommand{\rulename}[1]{%
  \ifcolors \bluebf{#1}%
  \else     \textbf{#1} \fi
}
\newcommand{\osfref}[1]{\rulename{\ref{#1}}}

\newcommand{\osfrule}[4][0pt]{\ensuremath{%
\begin{array}[t]{ll}
                 & \rulename{#2}         %
\\ %
                 & #3             %
\\               \cmidrule(r{#1}){2-2}   %
                 & #4             %
\end{array}}}

\newcommand{\osfrulec}[5][0pt]{\ensuremath{
\begin{array}[t]{ll}
                 & \rulename{#3}         %
\\ %
 & \multicolumn{1}{l}{\left[#2\right]}   %
\\ \\
                 & #4\gstick             %
\\               \cmidrule(r{#1}){2-2}   %
                 & #5\gstick             %
\end{array}}}

\renewcommand{\osfrulec}[5][0pt]{\ensuremath{
\begin{array}[t]{lll}
   & \rulename{#3} &                                    %
\\
& #4     &  \multirow{2}{*}{$\left[#2\right]$} %
\\
\cmidrule(r{#1}){2-2}                                %
   & #5     &                                    %
\end{array}}}

\newcommand{\osfeq}{\mathop{\mathit{Eq}}} %
\newcommand{\solved}{\mathop{\mathit{Solved}}}

\newcommand{\tags}{\mathop{\mathit{Tags}}}
\newcommand{\osftags}{\mathop{\mathit{Tags}}}
\newcommand{\sort}{\mathbin{\mathit{Sort}}}

\newcommand{\ii}{\mathcal{I}}
\newcommand{\jj}{\mathcal{J}}
\newcommand{\kk}{\mathcal{K}}
\renewcommand{\gg}{\mathcal{G}}
\renewcommand{\jj}{\mathcal{J}}
\newcommand{\iimod}{\ii, \alpha \models}
\newcommand{\iimodb}{\ii, \alpha \models_{\beta}}
\newcommand{\dom}[1][\ii]{\varDelta^{{#1}}}
\newcommand{\obj}[1]{\mathit{#1}}
\newcommand{\osfa}{(\dom, \cdot^\ii)}
\newcommand{\osfg}{(\dom[\gg], \cdot^\gg)}
\NewDocumentCommand{\denot}{O{t} O{\ii}}{[\![#1]\!]^{#2}}
\NewDocumentCommand{\denota}{O{t} O{\alpha}}{[\![#1]\!]^{\ii,#2}}
\NewDocumentCommand{\denoti}{O{t} O{\ii} }{[\![#1]\!]^{#2,\alpha}}

\newcommand{\PhiR}{\ensuremath\Phi_{R}}
\newcommand{\osfval}{\alpha:\V\to\dom}
\newcommand{\val}{\mathit{Val}}
\newcommand{\reach}{\mathrel{\overset{\phi}{\rightsquigarrow}}}

\newcommand{\ograph}{\ensuremath(N, E, \lambda_N, \lambda_E, \X)}
\NewDocumentCommand{\tgraph}{O{f} O{g}}{G(\Z_{\f[#1],#2}:\tops)}
\NewDocumentCommand{\tvar}{O{f} O{g}}{\Z_{\f[#1],#2}}
\newcommand{\pshi}{\psi_{\phi}}
\newcommand{\psig}{\psi_{G}}
\newcommand{\canon}[1][\phi]{\gg[\dom[\gg,#1]]}
\newcommand{\gequiv}{\equiv}

\newcommand{\approximates}[1][\gg]{\sqsubseteq^{#1}}
\NewDocumentCommand{\fapproximates}{O{\gg}}{\mathop{\sqsubseteq^{#1}}}
\NewDocumentCommand{\fapprel}{O{\gg} O{\beta}}{\mathbin{\sqsubseteq^{#1}_{#2}}}

\makeatletter
\def\th@plain{%
  \thm@notefont{}%
  \itshape %
}
\def\th@definition{%
  \thm@notefont{}%
  \normalfont %
}
\makeatother

\declaretheorem[name=Proposition,numberwithin=section]{prop}
\Crefname{prop}{Proposition}{Propositions}
\declaretheorem[name=Theorem,numberlike=prop]{theorem}
\declaretheorem[name=Lemma,numberlike=prop]{lemma}
\declaretheorem[name=Corollary,numberlike=prop]{corollary}

\theoremstyle{definition}
\declaretheorem[name=Definition,numberlike=prop]{definition}
\newtheorem{remark}{Remark}
\newtheorem*{remark*}{Remark}

\newtheorem*{notation*}{Notation}
\declaretheorem[name=Example,qed={$\triangleright$},numberlike=prop]{example}

\newenvironment{subproof}[1][\proofname]{%
  \begin{proof}[#1]%
}{%
  \end{proof}%
}
\declaretheorem[name=Claim,numberlike=prop]{claim}

\newif\iffuzzy
\newif\ifallproofs
\newif\ifshortproofs
\newif\ifrefer
\newif\ifclaimproofs

\newcommand{\myrestatable}[6]{%
  \ifallproofs
  \begin{restatable}%
    [{\protect\hyperlink{#4}{#1}}]%
    {#2}{#5}
    \label{#3}
    {#6}
  \end{restatable}
  \else
  \begin{#2}[#1]
    \label{#3}
    {#6}
  \end{#2}
  \fi
}

\fuzzytrue %
\colorstrue
\shortproofstrue

\begin{document}

\begin{frontmatter}
  \title{Fuzzy order-sorted feature logic}
  \author[1]{Gian Carlo Milanese\corref{cor}}
  \ead{giancarlo.milanese@unimib.it}

  \author[1]{Gabriella Pasi}
  \ead{gabriella.pasi@unimib.it}

  \cortext[cor]{Corresponding author}
  \affiliation[1]{%
    organization={Università degli Studi di Milano-Bicocca},
    country={Italy}
  }
  \begin{abstract}
    \OSF (\osf) logic is a knowledge representation and reasoning language
based on
function-denoting \textit{feature symbols}
and
set-denoting \textit{sort symbols}
ordered in a subsumption lattice.
\osf logic allows the construction of record-like terms that represent
classes of entities and that are themselves ordered in a subsumption
relation. The unification algorithm for such structures provides an
efficient calculus of type subsumption, which has been
applied in computational linguistics and
implemented in constraint logic programming languages
such as LOGIN and LIFE
and automated reasoners such as CEDAR.
This work generalizes
\osf logic to a fuzzy setting. We give
a
flexible definition of a \textit{fuzzy subsumption} relation
which generalizes Zadeh's
inclusion between fuzzy sets. Based on this definition we define a fuzzy
semantics of \osf logic
where sort symbols and \osf terms denote fuzzy sets.
We extend the subsumption relation to \osf terms and prove that it
constitutes a fuzzy partial order
with the property that
two \osf terms are subsumed by one another in the crisp sense
if and only if
their
subsumption degree is greater than 0.
We show how to find the greatest lower bound of two \osf terms by unifying
them and how to compute the
subsumption degree between two \osf terms,
and we provide the complexity of these operations.%

  \end{abstract}
\end{frontmatter}

\section{Introduction}%
\label{cha:introduction}

Order-Sorted Feature (\osf) logic is a Knowledge Representation and
Reasoning language that originates in Hassan \ak's work \cite{AitKaci1984}
and, similarly to Description Logics (DLs), it was initially meant as a
formalization of Ron Brachmann's structured inheritance networks
\cite{DLKRHB,AitKaci2007b}.
\osf logic and related formalisms
-- e.g., feature logic \cite{Smolka1988} or the logic of typed feature
structures \cite{Carpenter1992} --
have been applied in computational linguistics \cite{Carpenter1992} and
implemented in constraint logic programming languages such as LOGIN
\cite{AitKaci1986b}, LIFE \cite{AitKaci1993b} and CIL \cite{Mukai1987} and,
more recently, in the very efficient CEDAR Semantic Web reasoner
\cite{AitKaciAmir2017,AmirAitKaci2017}.%

At the core of \osf logic are \textit{sort symbols} denoting conceptual classes such as
$\s[person]$ or $\s[student]$, and a \textit{sort subsumption relation}
that denotes inclusion between classes, comparable to the inclusion axioms
of a DL terminological box. Concept subsumption relations are a
central part of any ontology, and \osf logic and its
implementations rely on graph encoding techniques to perform Boolean
operations on a concept lattice very efficiently
\cite{AitKaciAmir2017,AitKaci1989}.
Besides sort symbols, \osf logic employs \textit{feature symbols} to
describe attributes of objects, like $\f[name]$, $\f[directed\_by]$ or
$\f[written\_by]$.
While feature symbols denote \textit{total functions}, which %
may
appear less versatile than the relational roles of DLs, versions
of \osf logic that support partial functions or relations have also been
defined
\cite{Smolka1988,Carpenter1992,AitKaci2007b}.

Together with \textit{variables}, also named \textit{coreference tags}, sort symbols and
feature symbols can be used to construct record-like structures called
\osf terms that can describe complex concepts, such as the
following, which denotes the class of movies that are written and
directed by the same person:%
\[
  \X[X] : \s[movie]
  \left(
    \begin{array}{lll}
      \f[directed\_by] & \to & \X[Y] : \s[person],\\
      \f[written\_by] & \to & \X[Y]
    \end{array}
  \right).
\]
The unification algorithm for such structures provides an efficient way
to decide whether two \osf terms are subsumed by each other, and, in
general, for finding the most general \osf term that is subsumed by both
terms, thus offering an efficient calculus of partially ordered types
\cite{AitKaci1986b}.%

This work presents a fuzzy generalization of Order-Sorted Feature (\osf)
logic in which both sorts and \osf terms
are interpreted as
\textit{fuzzy subsets} of the domain of an interpretation rather than crisp
subsets, and where the sort subsumption relation is generalized to a
\textit{fuzzy subsumption relation between sorts}, which is then
extended to a fuzzy subsumption between \osf terms.

The idea of a fuzzy generalization of \osf logic
was first introduced in \cite{AitKaciPasi2020}
where a
weaker notion of
first-order term
unification based on similarity relations between
term constructors is introduced that also allows mismatches
between functor arities and argument positions.
While the interpretation of sort symbols (concept names in DL lingo) as
fuzzy subsets of a domain and the notion of fuzzy or graded subsumption
have already been studied extensively within fuzzy DLs,
the generalization of the semantics of \osf logic to a fuzzy setting is
novel.
In particular, the way in which we interpret
fuzzy subsumption relations
departs significantly from their treatment in fuzzy DLs, which relies
on the fuzzy implication operator \cite{Straccia2014,Borgwardt2017}.

In the rest of this section we provide an informal presentation of our
definition of fuzzy sort subsumption and how it is extended to the terms
of \osf logic, and we motivate the development of fuzzy \osf logic by
outlining its possible implementation
as an extension of the CEDAR reasoner
\cite{AitKaciAmir2017,AmirAitKaci2017} that would be capable of providing
approximate solutions in retrieval applications,
and as a fuzzy logic programming language.

\paragraph{Fuzzy sort subsumption}%

Let $\S$ and $\F$ be (finite) sets of \textit{sort symbols}
  and \textit{feature symbols}, respectively, and let us fix
 an interpretation
  $\ii = \osfa$, i.e., a structure with a domain
  $\dom$ and a function $\cdot^\ii$ that provides an interpretation to the
  elements of $\S$ and $\F$\footnote{Fuzzy interpretations will be defined
  properly in \cref{cha:semantics}.}.
  In our fuzzy setting, a sort $\s\in\S$ is interpreted as a
  fuzzy subset\footnotemark{}
  $\s^\ii:\dom\to[0,1]$, while a feature symbol $\f\in\F$ is
  interpreted as a function $\f^\ii:\dom\to\dom$.
  We rely on the minimum t-norm (denoted $\land$) and the maximum
  t-conorm (denoted $\lor$).
  \footnotetext{The definitions of fuzzy sets, fuzzy binary relations and
  fuzzy orders are recalled in \cref{app:fuzzy}.}%

In the crisp setting, a subsumption relation $\mathop{\isa} \subseteq
\S\times\S$ is a binary relation that denotes set inclusion. A natural
fuzzy generalization of this notion is Zadeh's inclusion of fuzzy sets
 \cite{DuboisPrade1980}, according to which the subsumption $\s\isa\su$ has
the following meaning:
\begin{align}
  \label{eq:fuzzy_inclusion}
  \text{if}~\s\isa\su,\text{then}~\forall d\in\dom: \s^\ii(d)\leq\su^\ii(d).
\end{align}
That is, this definition of fuzzy inclusion requires
  that, whenever $d$ is an
instance of $\s^\ii$ with degree
$\s^\ii(d)=\beta\in[0,1]$, then $d$ must
also be an instance of $\su^\ii$ with a degree
\textit{greater than or equal to} $\beta$.
\begin{figure}[ht!]
  \centering
  \begin{tikzpicture}[scale=0.8]
      \node[thick,shape=ellipse,draw=sortcol] (bot) at (2, -1.5) {$\bots$};
      \node[thick,shape=ellipse,draw=sortcol] (sla) at (4, 0) {$\s[slasher]$};
      \node[thick,shape=ellipse,draw=sortcol] (per) at (0, 0) {$\s[director]$};
      \node[thick,shape=ellipse,draw=sortcol] (ttl) at (1.5, 1.5) {$\s[string]$};
      \node[thick,shape=ellipse,draw=sortcol] (dir) at (0, 3) {$\s[person]$};
      \node[thick,shape=ellipse,draw=sortcol] (thr) at (4, 1.5) {$\s[thriller]$};
      \node[thick,shape=ellipse,draw=sortcol] (hor) at (7, 1.5) {$\s[horror]$};
      \node[thick,shape=ellipse,draw=sortcol] (mov) at (4, 3) {$\s[movie]$};
      \node[thick,shape=ellipse,draw=sortcol] (top) at (2, 4.5) {$\tops$};

      \draw[->,thick] (bot.west) to[bend left=20] node[left,yshift=-1mm] {$\red{1}$} (per.south);
      \draw[->,thick] (bot.east) to[bend right=20] node[right,yshift=-1mm]   {$\red{1}$} (sla.south);
      \draw[->,thick] (bot.north) to[bend left=10] node[right]   {$\red{1}$} (ttl.south);
      \draw[->,thick] (per.north) to node[left]   {$\red{1}$} (dir.south);
      \draw[->,thick] (sla.north) to node[right]   {$\red{\appdegree}$} (thr.south);
      \draw[->,thick] (sla.east) to[bend right=20] node[right,yshift=-1mm]   {$\red{1}$} (hor.south);
      \draw[->,thick] (hor.north) to[bend right=20] node[right,yshift=1mm]   {$\red{1}$} (mov);
      \draw[->,thick] (thr.north) to node[right]   {$\red{1}$} (mov);
      \draw[->,thick] (ttl.north) to[bend left=10] node[right]   {$\red{1}$} (top.south);
      \draw[->,thick] (dir.north) to[bend left=20] node[left,yshift=1mm]   {$\red{1}$} (top.west);
      \draw[->,thick] (mov.north) to[bend right=20] node[right,yshift=1mm]   {$\red{1}$} (top.east);
  \end{tikzpicture}
  \caption{Fuzzy subsumption relation}
  \label{fig:fuzzy_sub_small}
\end{figure}
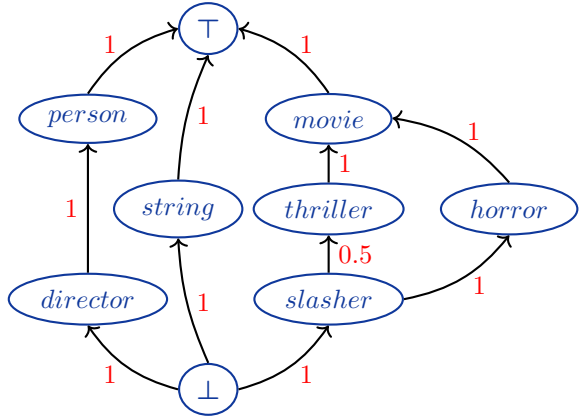

We consider a \textit{fuzzy subsumption relation}
as a way to model a weaker notion of inclusion, where the element $d$ can
be an instance of $\su^\ii$ with a degree
that \textit{may possibly be smaller than} its degree
of membership to $\s^\ii$.
Thus, we define a fuzzy subsumption relation as a fuzzy partial order,
i.e., a function $\fisa: \S^2\to[0,1]$ that associates a subsumption degree
$\beta$ with each pair of sort symbols $\si[0], \si[1]$ (and satisfying
every constraint of \cref{def:fuzzy_poset} in
\cref{app:fuzzy}) and having the following semantics:
\begin{align}
  \label{eq:fuzzy_subsumption}
  \text{if}~\fisop(\si[0], \si[1])=\beta,~\text{then}~\forall d\in\dom:
  \sii[0](d) \land \beta \leq \sii[1](d).
\end{align}
That is, any object $d$ which is an instance of $\sii[0]$ with degree
$\beta_0$ must also be an instance of $\sii[1]$ with a degree $\beta_1$
  that is \textit{greater than or equal to the minimum of}
$\beta_0$ and $\beta$.
Note that \cref{eq:fuzzy_inclusion} is a special case of this equation with
$\beta = 1$.
An example of a fuzzy sort subsumption
relation is given in \cref{fig:fuzzy_sub_small}\footnotemark{}.
\footnotetext{We represent a fuzzy subsumption relation $\fisa$ graphically
  as a weighted directed acyclic graph (DAG), of which $\fisa$ is the
reflexive and transitive closure (see \cref{def:fuzzy_transitive_closure}
in \cref{app:fuzzy}).}

\paragraph{Fuzzy \osf term subsumption and unification}%

Sort symbols are analogous to the primitive or named concepts
of DLs. Complex concepts can be expressed with
\textit{\osf terms},
structures built from \sortcol{sort symbols}, \featcol{\textit{feature
symbols}} (attributes) and \tagcol{variables}. For example, the term
\[\arraycolsep=2pt
  t_1 = \Xj[1]:\s[movie]
  \left(
    \begin{array}{lll}
      \dirby& \to & \Yj[1]:\person,\\
      \f[genre] & \to & \Zj[1]:\thriller
    \end{array}
  \right)
\]
denotes the class of movies directed by some person and whose genre is
thriller.

In this work we will show that
a fuzzy subsumption relation between sort
symbols can be extended to a fuzzy subsumption relation between \osf terms.
For example, consider the fuzzy subsumption relation depicted in
\cref{fig:fuzzy_sub_small}
and consider the term
\[
  \arraycolsep=2pt
  t_2 = \X[X_{2}]:\s[movie]
  \left(
    \begin{array}{lll}
      \f[title] & \to & \X[W_{2}]:\strng,  \\
      \f[genre] & \to & \X[Z_{2}]:\slasher, \\
      \dirby    & \to & \X[Y_{2}]:\director
    \end{array}
  \right).
\]
This term is more specific that $t_1$, as it provides
additional information by
introducing the feature title and by constraining its value
to be of sort $\strng$,
and because it defines more restrictive constraints,
by requiring
that the value of the feature $\dirby$ be of sort $\director$, which is
subsumed by $\person$, and
that the value of the feature $\f[genre]$ be of sort
$\s[slasher]$, which is subsumed by $\s[thriller]$ (with degree
\appdegree).
In this case we say that $t_1$ \textit{subsumes} $t_2$ with degree
$\appdegree$.

One of the reasoning tasks supported by \osf logic is deciding whether a
given term is subsumed by another, or in general finding the most general
term which is subsumed by two given terms. As we will see,
  this problem can be solved by their \textit{unification}, a
  process which aims to combine the constraints expressed by the two terms
  in a consistent way.
For example, $t_2$ is the unifier of term $t_1$ and the following
term:
\[ t_3 = \Xj[3]:\s[movie]
  \left(
    \begin{array}{lll}
      \dirby & \to & \Yj[3]:\s[director],\\
      \f[title] & \to & \X[W_3]:\strng,\\
      \f[genre] & \to & \Zj[3]:\s[horror]
    \end{array}
  \right).
\]
In particular, the value for the feature $\f[genre]$ in $t_2$ must be of
sort $\s[slasher]$, as this sort is subsumed by both $\s[thriller]$
and $\s[horror]$ (the values of $\f[genre]$ in $t_1$ and $t_3$), and it
is the most general one with this property. The unifier $t_2$ is
associated with a unification degree, which depends on the subsumption
degrees of its sorts with respect to the corresponding sorts in $t_1$ and
$t_3$. In this case the unification degree is $0.5$, due to
$\fisop(\slasher, \thriller) = 0.5$.

\paragraph{Application to fuzzy logic programming}%

A possible application of \osf logic with a fuzzy subsumption relation is the
implementation of a logic programming language (like LOGIN \cite{AitKaci1986b}) based
on SLD resolution and fuzzy \osf term unification, i.e., where first-order term
unification is replaced by the unification of \osf terms over a fuzzy sort
subsumption lattice.
Logic programming languages and weaker definitions of unification based on fuzzy
relations such as similarities and proximities have been researched extensively
(e.g.,
\cite{Arcelli2002,Gerla1999,Sessa2002,Iranzo2015b,Iranzo2020,Kutsia2020,AitKaciPasi2020})
and implemented in systems such as \bpl \cite{Iranzo2023} and FASILL
\cite{Iranzo2020fasill}.%

There are several potential advantages to using \osf logic in this context.
First of all, the unification
algorithm for \osf terms takes into account a (fuzzy) sort subsumption relation,
which can result in more efficient computations \cite{AitKaci1986b,Cohn1989}.
Another advantage is the flexibility provided by \osf terms,
which lack a fixed arity and can thus easily represent partial information,
and are moreover simpler to interpret thanks to their use of features rather than
positions to specify arguments \cite{AitKaci1986b}.

\paragraph{Application to similarity-based reasoning}%

Another application of our definition of fuzzy
subsumption relation is similarity-based reasoning.
A possible way to define a fuzzy subsumption relation
is by taking advantage of a given similarity relation $\mathop{\sim}:
\S\times\S\to[0,1]$ between sort symbols in order to enrich a crisp
subsumption relation $\mathop{\isa}\subseteq\S\times\S$ according to the
following intuitive inference rule, which also motivates similarity-based
approaches in logic programming (e.g., \cite{Sessa2002}):
    if   $\s\isa\su$ and $\simop(\su, \s[s'']) = \beta$,
    then $\fisop(\s, \s[s'']) = \beta$.
For example,
    if   $\s[slasher]\isa\s[horror]$
    and  $\simop(\s[horror], \s[thriller]) = \beta$,
    then $\fisop(\s[slasher], \s[thriller]) = \beta$.
This would mean, for example, that if $h$ is an instance of
$\slasher$ with degree $1$, then $h$ must also be an instance of
$\thriller$ with degree greater than or equal to $\beta$.

This approach could be behind the implementation of a fuzzy logic
programming language similar to \bpl \cite{Iranzo2023}, where
the similarity-based
or (proximity-based) first-order
term unification of this language
may be replaced by similarity-based \osf term unification, with the potential
advantages discussed above.

\paragraph{Fuzzy CEDAR}%

Finally, we plan to apply fuzzy \osf logic to define a fuzzy
  version of CEDAR\footnote{%
  More information about the CEDAR project
  can be found at \url{https://cedar.liris.cnrs.fr/},
  including reports, papers, demos and software.},
  a very efficient Semantic Web reasoner based on \osf
  logic \cite{AitKaciAmir2017,AmirAitKaci2017}. One of the
  capabilities supported by CEDAR is the optimization of a query expressed
  as an \osf term according to the knowledge expressed in a given
  ontology, which
  can significantly simplify the original query and reduce the
  instance retrieval
  search space.
  The consistency of the input query against the ontology is also ensured
  during this process, and no answer is provided if the query is
  inconsistent.
  The retrieval is further optimized thanks to a custom RDF triple indexing
  scheme based on \osf sort and attribute information
  \cite{AmirAitKaci2017}.

A fuzzy extension of CEDAR could relax the consistency
  requirement to
  provide approximate answers when the input
  query is inconsistent with the
  \textit{crisp} knowledge expressed by the ontology.
  This would be achieved, for example, by
  enriching the subsumption relation of the given ontology according to a
  similarity relation (with the procedure outlined above) in order to
  obtain a fuzzy subsumption, which would then be taken into
  account during the query optimization phase.%

\paragraph{Structure of the paper}%

After recalling its syntax in \cref{cha:syntax},
our fuzzy generalization of \osf logic begins in \cref{cha:semantics},
where we define the interpretation of the syntactic objects of this
language -- namely sorts, features, \osf terms and \osf clauses -- in
structures called fuzzy \osf interpretations, and a special
interpretation called the fuzzy \osf graph algebra, whose elements are
rooted directed labeled graphs called \osf graphs.

In \cref{sec:homomorphisms} we define structure-preserving mappings between
fuzzy \osf interpretations called fuzzy \osf homomorphisms, which are
valuable for proving several results regarding the
satisfiability of \osf clauses in fuzzy interpretations.%

Fuzzy \osf homomorphisms are then employed extensively in
\cref{sec:subsumption}, where we extend the fuzzy sort subsumption
ordering to \osf terms and prove that it constitutes a fuzzy partial order.
We provide similar orderings for \osf clauses and \osf graphs, and
illustrate how
the fuzzy \osf term subsumption ordering is related to its crisp
counterpart.

\cref{sec:unification} is devoted to unification. We show how to compute
the greatest lower bound of two \osf terms in the fuzzy \osf term
subsumption ordering through their unification, and
how this procedure can be used to find the
degree to which two \osf terms are subsumed by each other.
We also discuss the complexity of these computations.%

\cref{app:table} presents a reference table of the symbols used in this
paper.
The basic definitions of fuzzy set theory and fuzzy orders are recalled
in \cref{app:fuzzy} to fix the notation.
The proofs of the main results
are reported in
\ifallproofs
\cref{proofs:sem,proofs:hom,proofs:sub,proofs:unif}.
\else
\cref{proofs:hom,proofs:sub,proofs:unif}.
\fi

\section{Fuzzy \osf logic: syntax}%
\label{cha:syntax}

In this section we provide the definitions of fuzzy \osf signature
and of two formal languages
that are used to
represent knowledge with \osf logic: \textit{\osf terms} and \textit{\osf clauses}.
As discussed in the introduction, \osf terms are comparable to the defined
concepts of DLs, and they will be interpreted as fuzzy subsets of a domain
of interpretation. An \osf clause is an equivalent representation that can
be seen as a logical reading of an \osf term, and for which a notion of graded
satisfaction will be defined. In \osf logic, both syntactic
representation are important from an implementation perspective, as \osf
terms are the abstract syntax employed by an user, while \osf clauses
are used in the constraint normalization rules needed for \osf term
unification
\cite{AitKaci1993b}.

As far as syntax is concerned, the \textit{only difference with crisp \osf
logic} is that we consider a fuzzy sort subsumption relation rather than a
crisp one (the most significant differences concern the
  semantics, as will be seen starting from \cref{cha:semantics}). The
  definitions of \osf terms and \osf constraints and the result concerning
  their equivalence are reported from \cite{AitKaci1993b} for
  self-containment.

\begin{definition}[Fuzzy \osf signature]
\label{def:sort_signature}
A \emph{fuzzy \osf signature} is a tuple $\fss$ where
  $\S$ is a set of \emph{sort symbols},
  $\F$ is a set of \emph{feature symbols}, and
  $\ftax$ is a fuzzy finite bounded lattice with least element
    $\bots$ and greatest element $\tops$\footnotemark.
\footnotetext{See \cref{def:fuzzy_lattice} in \cref{app:fuzzy}.}
Elements of $\S$ and $\F$ will also simply be called \emph{sorts} and
\emph{features}, respectively.
The greatest lower bound (GLB) $\s\fmeet\su$ of two sorts $\s$ and $\su$ is
also called their \emph{greatest common subsort}.
\end{definition}

\begin{example}[Fuzzy \osf signature]
\label{ex:fuzzy_osf_signature}
  As an example of a fuzzy \osf signature we may take the set of sorts and
  the fuzzy subsumption relation corresponding to the graph of
  \cref{fig:fuzzy_sub_small}, and $\F = \{ \f[directed\_by], \f[title] \}$
  as the set of features.
\end{example}

\begin{definition}[\osf term \cite{AitKaci1993b}]
\label{def:osf_term}
Let $\V$ be a countably infinite set of variables (or
\emph{coreference tags}, or simply \textit{tags}). Let
$\X\in\V$, $\s\in\S$ and $\fj, \ldots,
\fj[n]\in\F$. An \emph{\osf term} is defined recursively as follows.
\begin{itemize}
    \item A sorted variable $\xs$ is an \osf term.
    \item If $t_1, \ldots, t_n$ are \osf terms, then an attributed sorted
      variable $t = \osfterm$ is an \osf term.
\end{itemize}
We let $\tags(t) \defeq \{ \X \} \cup \bigcup_{1\leq i \leq
n}\tags(t_i)$.
The variable $\X$ is called the \textit{root tag} of $t$ and is denoted
$\rtag(t)$.
\end{definition}

\begin{example}[\osf term]
  \label{ex:osf_term}
  The following \osf term shows how variables can be used to indicate
  coreference. Variables that are not used for coreference may be left
  implicit to improve readability.
    In line with \cite{AitKaci1993b}, we adhere to the convention of
    specifying the sort of each variable at most once, with the implicit
    understanding that other occurrences also refer to the same structure.
  \[
    \s[movie]
    \left(
      \begin{array}{lll}
        \f[title] & \to & \s[string],\\
        \f[directed\_by] & \to & \X[X]:\s[director]
        \left(
          \begin{array}{lll}
            \f[name] & \to & \s[string],\\
            \f[spouse] & \to & \X[Y]
          \end{array}
        \right),\\
        \f[written\_by] & \to & \X[Y]:\s[writer]
        \left(
          \begin{array}{lll}
            \f[spouse] & \to & \X[X]
          \end{array}
        \right)
      \end{array}
    \right).\qedhere
  \]
\end{example}

The definition of \osf terms given above does not rule out the presence of
redundant or even contradictory information (e.g., consider the \osf term
$\s(\f\to \si[0], \f\to\si[0], \f\to\si)$, which is contradictory if
$\si[0]\fmeet\si = \bots$). \osf terms that are well-behaved to this regard
are called
\ifrefer%
  \textit{normal \osf terms}, %
  and the reader is referred to \cite{AitKaci1993b} for their definition.
  \osf terms in normal form are also called \textit{$\psi$-terms} and
  denoted $\psi$, $\psi_i$, and so on.
  For an \osf term $\psi$ in normal form and $\X\in\tags(\psi)$, we let
  $\sort_\psi(\X)$ be the most specific sort $\s$ such that $\X:\s$ appears in
  $\psi$.
  The notation $\X\fto_\psi\Y$ indicates that there is a feature $\f$
  pointing from a subterm of $\psi$ with root tag $\X$ to a subterm of
  $\psi$ with root tag $\Y$.
  We let $\Psi$ denote the set of all normal \osf terms.%

\else
\textit{normal \osf terms} and are defined as follows \cite{AitKaci1993b}.
\begin{definition}[Normal \osf term, or $\psi$-term \cite{AitKaci1993b}]
\label{def:osf_term_normal form}
  An \osf term $t=\osfterm$ is in \textit{normal form} (or \textit{normal})
  if: (i) the root sort $\s$ is different from $\bots$,
  (ii) the features $\fj[1], \ldots, \fj[n]\in\F$ are pairwise distinct,
  (iii) each $t_i$ is in normal form, and
  (iv) for all $\Y\in\tags(t)$, there is at most one occurrence of $\Y$ in
  $t$ such that $\Y$ is the root variable of an \osf term different from
  $\Y:\tops$.

  \osf terms in normal form are also called \textit{$\psi$-terms} and
  denoted $\psi$, $\psi_i$, and so on. For an \osf term $\psi$ in normal
  form and $\X\in\tags(\psi)$, we let
  $\sort_\psi(\X)$ be the most specific sort $\s$ such that $\X:\s$ appears in
  $\psi$.
  The notation $\X\fto_\psi\Y$ indicates that there is a feature $\f$
  pointing from a subterm of $\psi$ with root tag $\X$ to a subterm of
  $\psi$ with root tag $\Y$.
  We let $\Psi$ denote the set of all normal \osf terms.
\end{definition}
\fi

A method for transforming an \osf term into a normal one is more easily
presented as a constraint normalization procedure for \osf constraints,
which are defined next.
\begin{definition}[\osf constraints and clauses \cite{AitKaci1993b}]
\label{def:osf_clause}
  An \osf constraint is an expression of the form $\xs$, $\osfce$, or
  $\osfcf$. An \osf clause $\phi$ is a conjunction of \osf constraints. The
  set of variables occurring in $\phi$ is denoted $\osftags(\phi)$, while
  $\phi[\X/\Y]$ is the \osf clause obtained by replacing all occurrences
  of $\Y$ with $\X$.
\end{definition}
Informally,
the constraint $\xs$ means that the value assigned to $\X$ is of sort $\s$;
$\osfce$ means that the same value is assigned to the variables $\X$ and $\X[X']$; while
$\osfcf$ means that applying the feature $\f$ to the value assigned to $\X$ returns the
value assigned to $\X[X']$.%

\begin{example}[\osf clause]
  \label{ex:osf_clause}
  Let $\phi$ be the following \osf clause.
  \[\arraycolsep=4pt
    \begin{array}[b]{llllllll}
      \X[X_{0}] : \s[movie] & \with & \X[X_{0}].\f[title] \doteq \X[X_{1}] & \with & \X[X_{1}] : \s[string] & \with & \X[X_{0}].\f[directed\_by] \doteq \X[X] & \with\\
      \X[X] : \s[director] & \with & \X[X].\f[name] \doteq \X[X_{2}] & \with & \X[X_{2}] : \s[string] & \with & \X[X].\f[spouse] \doteq \X[Y] & \with\\
      \X[X_{0}].\f[written\_by] \doteq \X[Y] & \with & \X[Y] : \s[writer] &
      \with & \X[Y].\f[spouse] \doteq \X[X].
    \end{array}
  \]
  Note that $\phi$ is simply a translation of the term from
  \cref{ex:osf_term} into an \osf clause.
  The variables that
  were left implicit in \cref{ex:osf_term} must be written explicitly in
  the \osf clause.
\end{example}

While every \osf term can be rewritten as an \osf clause, the converse is
in general only true for the class of
\ifrefer
\textit{rooted solved \osf clauses}, which are defined in \cite{AitKaci1993b}.
  A rooted \osf clause is denoted $\phi_{\X}$, where $\X$ is the
  root of the clause $\phi$. For an \osf clause $\phi$ and a variable
  $\X\in\osftags(\phi)$, the maximal
  subclause of $\phi$ rooted in $\X$ is denoted $\phi(\X)$.
  The set of all \osf clauses in solved form is denoted $\Phi$, and the
  subset of rooted solved \osf clauses is denoted $\PhiR$.%
\else
\textit{rooted solved \osf clauses} \cite{AitKaci1993b},
which are defined next.
\begin{definition}%
  [Rooted \osf clause and maximal subclause rooted in $\X$ \cite{AitKaci1993b}]
\label{def:rooted_clause}
  Given an \osf clause $\phi$, the binary relation ${\reach}\subseteq
  \osftags(\phi)\times\osftags(\phi)$ is defined as follows, for all $\X,
  \Y\in\osftags(\phi)$: (i) $\X\reach\X$, and (ii) $\X\reach\Y$ if there is
  a constraint $\X.\f\doteq\Z$ in $\phi$ and $\Z\reach\Y$.
  A \textit{rooted \osf clause} $\phi_{\X}$ is an \osf clause $\phi$ together
  with a distinguished variable $\X$ (its root) such that every variable
  $\Y$ occurring in $\phi$ is explicitly sorted (possibly as $\Y:\tops$)
  and reachable from $\X$ (i.e., $\X\reach\Y$).
  Given a clause $\phi$ and a variable $\X\in\osftags(\phi)$,
  the \textit{maximal subclause of $\phi$ rooted in $\X$} is denoted $\phi(\X)$.
\end{definition}

\begin{definition}[Solved \osf clause \cite{AitKaci1993b}]
\label{def:solved_clause}
  An \osf clause $\phi$ is called \textit{solved}
  (or \textit{in solved form}) if, for each variable
  $\X$, $\phi$ contains
  (i) at most one sort constraint of the form $\X:\s$ with $\s\neq\bots$,
  (ii) at most one feature constraint of the form $\X.\f\doteq \Y$ for each
  $\f$, and
  (iii) no equality constraint of the form $\X\doteq\Y$.
  The set of all \osf clauses in solved form is denoted $\Phi$, and the
  subset of rooted solved \osf clauses is denoted $\PhiR$.
\end{definition}
\fi

\begin{prop}[Equivalence of $\psi$-terms and rooted solved \osf
  clauses \cite{AitKaci1993b}]
\label{prop:bijections_terms_clauses}
  There exist bijective mappings $\phi:\Psi\to\PhiR$ and
  $\psi:\PhiR\to\Psi$, i.e., such that\footnote{%
    For a set $S$, the symbol $\id[S]$ denotes the identity function
    $\id[S]:S\to S$ defined by letting $\id[S](s) = s$ for all $s\in S$.}
  $\id[\PhiR] = \phi\circ\psi$ and
  $\id[\Psi] = \psi\circ\phi$.
\end{prop}
Note that we use the same symbol $\phi$ to denote an \osf clause and the
function mapping an \osf term to a clause (and similarly for the symbol
$\psi$), as the meaning is always clear from context.

Finally, we report the constraint normalization rules from
\cite{AitKaci1993b} that are needed to transform an \osf clause into a
solved form. Each rule is of the form
\begin{center}
\osfrulec{\text{Side condition}}{\lab{Rule name}{osf:rn}}
{\text{Premise}~\phi}{\text{Conclusion}~\phi'}\\
\end{center}
and it expresses that, whenever the (optional) side condition holds, the
premise
$\phi$ can be simplified into the conclusion $\phi'$.
The rules of \cref{fig:osf_normalization} are the same as the ones from
\cite{AitKaci1993b}, except that we are considering the GLB operation
$\fmeet$ in a fuzzy lattice rather than an ordinary one. Thanks to
\cref{prop:glbs}, this does not constitute a significant difference as far
as the properties of the normalization procedure are concerned.
An \osf term $\psi$ can be normalized by applying the constraint
normalization rules to $\phi(\psi)$ and translating the result back into an
\osf term.%

\begin{restatable}%
  [{{\osf} clause normalization \cite{AitKaci1993b}}]%
    {theorem}{propclausenormalization}
\label{prop:osf_clause_normalization}
The rules of \cref{fig:osf_normalization} are finite terminating and
confluent (modulo variable renaming). Furthermore, they always result in a
normal form that is either the inconsistent clause or an \osf clause in
solved form together with a conjunction of equality constraints.
\end{restatable}

\begin{figure}[ht!]
\centering
\begin{tabular}{ll}
\osfrule{\lab{Fuzzy Sort Intersection}{osf:si}}
{\phi \osfwith \xs \osfwith \xs[X][s']}
{\phi\osfwith\xs[X][s\fmeet s']}
&
\osfrule{\lab{Feature Functionality}{osf:ff}}
{\phi\osfwith \osfcf[\X][\f][\Y] \osfwith \osfcf[\X][\f][\Y[Y']]}
{\phi\osfwith \osfcf[\X][\f][\Y]\osfwith \osfce[\Y][\Y[Y']]}
\\ \\
\osfrule[1.4cm]{\lab{Inconsistent Sort}{osf:is}}{\phi\osfwith
\xs[\X][\bots]}{\X:\bots}
&
\osfrulec{\Y\in\osftags(\phi)}{\lab{Tag Elimination}{osf:te}}
{\phi\osfwith \osfce[\X][\Y]}{\phi[\X/\Y]\osfwith \osfce[\X][\Y]}\\
\end{tabular}
\caption{Fuzzy \osf Constraint Normalization Rules}
\label{fig:osf_normalization}
\end{figure}

\section{Fuzzy \osf interpretations}%
\label{cha:semantics}

In this section we start our development of a fuzzy semantics of \osf logic
based on the minimum t-norm and the maximum t-conorm.
We begin by defining \textit{fuzzy \osf interpretations}, the structures
used to interpret the syntax of \osf logic.
These provide the meaning of a fuzzy sort subsumption relation and
determine the denotations of sort symbols as fuzzy subsets and of feature
symbols as functions.

In \cref{cha:syntax} we have presented \osf terms and \osf
  clauses as two alternative (and syntactically equivalent) data structures
for representing knowledge with \osf logic.
We thus define
the \textit{denotation} of an \osf term as a fuzzy subset of the domain
and
the \textit{graded satisfaction} of an \osf clause,
and show that these two notions are semantically equivalent too.

There is, in fact, a third syntactic representation level, namely that of
\textit{\osf graphs}. These are rooted directed graphs whose nodes are
labeled by sorts and whose edges are labeled by features, and that are
also syntactically equivalent to $\psi$-terms and rooted solved \osf
clauses. \osf graphs play an essential role in the semantics of fuzzy \osf
logic, as they are the elements of the domain of
the \textit{fuzzy \osf graph algebra},
a fuzzy interpretation in which every solved \osf clause is
satisfiable.%

\subsection{Fuzzy \osf interpretations}%
\label{sec:osf_algebra}

Let us fix a fuzzy signature $\fss$ for the rest of the section.

\begin{definition}[Fuzzy \osf interpretation]
\label{def:osf_algebra}
A \textit{(fuzzy) \osf interpretation} (or \textit{algebra}) for $\fss$ is
a pair $\ii = \osfa$ such that\footnotemark
\begin{enumerate}
  \item $\dom$ is a set called the \textit{domain} or \textit{universe} of
    the interpretation;
  \item for each $\s\in \S$, $\s^\ii:\dom\to[0,1]$ is a fuzzy subset of
    $\dom$ (where in particular $\tops^\ii = \cf[{\dom[\ii]}]$ and
    $\bots^\ii = \cf[\emptyset]$);
  \item for each $\si[0],\si[1]\in \S$,
    $\forall d\in\dom: \sii[0](d)\land\fisop(\si[0], \si[1])\leq\sii[1](d)$;
  \item for each $\si[0],\si[1]\in \S,\forall d\in\dom$: if
    $(\sii[0]\fcap \sii[1])(d) >0$,
    then $(\si[0]\fmeet\si[1])^\ii(d)> 0$; and
  \item for each $\f\in \F$: $\f^\ii$ is a function $\f^\ii:\dom\to\dom$.
\end{enumerate}
\end{definition}
In the following we write $d\in\ii$ and $d\in\dom$ interchangeably.

\footnotetext{%
    The symbol $\cf[D]$ denotes the characteristic function $\cf[D]:
    \dom\to\{ 0,1 \}$ of the set
    $D\subseteq\dom$, which is defined by letting, for all $d\in \dom$,
    $\cf[D](d) \defeq 1$
    if $d\in D$, and
    $\cf[D](d) \defeq 0$
    otherwise. The symbol $\fcap$ denotes the intersection of
    fuzzy subsets (\cref{def:intersection_of_fuzzy_subsets}).}

The motivation behind condition (3) has been discussed in the
\nameref{cha:introduction}.
In order to motivate condition (4), let us momentarily switch to a crisp
setting, where $\s^\ii$ denotes a (crisp) subset of the domain
$\dom[\ii]$, $\isop\subseteq \S^2$ is a binary (subsumption) relation and
$\meet$ is the GLB operation on $(\S, \isa)$. The definition of an \osf
algebra \cite{AitKaci1993b} requires that
$\sii[0]\cap\sii[1] \subseteq (\si[0]\meet\si[1])^\ii$ holds for every
$\si[0], \si[1]\in\S$, i.e., whenever an element belongs to the
intersection of the denotations of two sorts, then it must also belong to
the denotation of their GLB.
Our fuzzy generalization of this requirement is very flexible
as it does not impose restrictive constraints on the membership degree of
$d$ to
$(\si[0]\fmeet\si[1])^\ii$, but it simply requires that whenever an object is
an instance of two sorts $\si[0]$ and $\si[1]$ with a degree greater than
$0$, then it must also be an instance of their GLB with a degree greater
than $0$.

\begin{example}[Fuzzy \osf interpretation]
\label{ex:fuzzy_osf_interpretation}
  Consider the fuzzy \osf signature of \cref{ex:fuzzy_osf_signature} and
  let $\ii = \osfa$ be defined as follows.
  \begin{itemize}
    \item The domain is $\dom[\ii] = \{
      \psycho,
      \halloween,
      \hitchcock,
      \carpenter,
      \obj{``Psycho''},
      \obj{``Halloween''},
      \nullv
    \}$.
    \item The interpretation of the sort symbols is defined by letting
    \begin{itemize}
      \item $\thriller^\ii(\halloween) = 0.5$ and
        $\horror^\ii(\halloween) = \slasher^\ii(\halloween) = 1$;
      \item $\thriller^\ii(\psycho) = \horror^\ii(\psycho)
        = 1$ and $\slasher^\ii(\psycho) = 0.7$;
      \item $\movie^\ii(\psycho) =\movie^\ii(\halloween) = 1$;
      \item $\strng^\ii(\obj{``Psycho''}) =
        \strng^\ii(\obj{``Halloween''}) = 1$;
      \item $\person^\ii(\hitchcock) =
        \person^\ii(\carpenter) = 1$;
      \item $\director^\ii(\hitchcock) =
        \director^\ii(\carpenter) = 1$;
      \item $\tops^\ii(x) = 1$ for every $x\in\dom[\ii]$; and
      \item all the remaining membership degrees are equal to 0.
    \end{itemize}
    \item The interpretation of the feature symbols is defined by letting
    \begin{itemize}
      \item $\dirby^\ii(\psycho) = \hitchcock$ and
        $\dirby^\ii(\halloween) = \carpenter$;
      \item $\ttle^\ii(\psycho) = \obj{``Psycho''}$ and
        $\ttle^\ii(\halloween) = \obj{``Halloween''}$; and
      \item all the remaining feature applications are equal to \nullv.
    \end{itemize}
  \end{itemize}
  This is easily verified to satisfy all constraints of
  \cref{def:osf_algebra}. In particular $\thriller^\ii(\halloween) =
  0.5 \geq 1\land 0.5 = \slasher^\ii(\halloween)\land\fisa(\slasher,
  \thriller)$.
\end{example}
Since features are interpreted as \textit{total} functions, in
the last example the feature $\ttle$ had to be defined also for elements of
sort $\person$ such as $\hitchcock$. While we can circumvent this issue by
assigning
$\ttle^\ii(\hitchcock)=\nullv$, there are versions of \osf logic that
interpret features as \textit{partial}
functions instead \cite{Smolka1988,Carpenter1992,AitKaci2007b}. An
analogous extension of fuzzy \osf logic
is left for future work.

We now define subalgebras of fuzzy \osf algebras.
\begin{definition}[$\F$-closure]
  \label{def:fclosure}
  Let $\ii$ be an \osf interpretation.
  For each feature composition $w = \fj[1]\ldots\fj[n]\in\F^*$ let $w^\ii =
  \fji[n]\circ \ldots\circ\fji[1]$ be the corresponding function composition
  on $\dom$. For any non-empty subset $D$ of $\dom$ the
  \textit{$\F$-closure} of $D$ is the set
  \[
    \F^*(D) \defeq \bigcup\limits_{w\in \F^*}w^\ii(D) =
    \bigcup\limits_{w\in \F^*}\{ w^\ii(d)\mid d\in D \}.
  \]
  In other words $\F^*(D)$ is the smallest set containing $D$ and closed
  under feature application.
\end{definition}

\begin{definition}[Fuzzy \osf subalgebra]
  \label{def:fuzzy_osf_subalgebra}
  Let $\ii$ and $\jj$ be fuzzy \osf interpretations.
  Then $\ii$ is a subalgebra of $\jj$ if
  $\dom[\ii]\subseteq\dom[\jj]$ and for all
  $d\in\dom[\ii]$, all $\s\in\S$ and all $\f\in\F$:
  $\s^\ii(d) = \s^\jj(d)$ and
  $\f^\ii(d) = \f^\jj(d)$.
\end{definition}

\begin{definition}[$\osf$ subalgebra generated by a set]
  \label{def:osf_subalgebra}
  Let $\ii = \osfa$ be a fuzzy \osf interpretation and $D\subseteq\dom$ be
  nonempty. Then the fuzzy $\osf$ subalgebra generated by $D$ is the structure
  $\ii[D] = (\F^*(D), \cdot^{\ii[D]})$ such that, for each $\s\in\S$,
  $\s^{\ii[D]} = \s^\ii\fcap\cf[\F^*(D)]$, and for each $\f\in\F$, $\f^{\ii[D]}$
  is the restriction of $\f^\ii$ to $\F^*(D)$.
\end{definition}
When $D=\{ d \}$ is a singleton we write $\ii[d]$ instead of $\ii[\{ d
\}]$.

\myrestatable%
  {Least $\osf$ subalgebra generated by a set} %
  {prop} %
  {prop:osf_subalgebra} %
  {proof:subalgebra} %
  {propsubalgebra} %
  {%
  Let $\ii = \osfa$ be an \osf interpretation. For any
  non-empty subset $D\subseteq\dom$ the structure $\ii[D]$ is the least
  fuzzy subalgebra of $\ii$ \mbox{containing $D$.}%
  }

\subsection{Denotation of an \osf term}%
\label{sec:denotation_of_an_osf_term_in_an_osf_algebra}

We now define the meaning of an \osf term in a fuzzy \osf interpretation. Let
$\val(\ii)$ be the set of all variable assignments $\osfval$.

\begin{definition}[Denotation of an \osf term]
\label{def:osf_term_denotation}
  Let $t = \osfterm$ be an \osf term, let $\ii = \osfa$ be a fuzzy \osf
  interpretation, and
  let $\alpha: \V\to\dom$ be a variable assignment. The denotation of
  $t$ in the interpretation $\ii$ under the assignment $\alpha$ is the fuzzy
  subset of $\dom$ defined by letting, for all $d\in\dom$:
  \[
    \denota[t](d) \defeq
    \cfun[\{ \alpha(\X) \}](d) \land
    \s^\ii(d)
    \land\bigwedge\limits_{1\leq i\leq n} \denota[t_i](\fji[i](d)).
  \]
  The denotation of $t$ in the interpretation $\ii$ is defined
  as\footnote{%
    The symbol $\fcup$ denotes the union of fuzzy subsets
  (\cref{def:intersection_of_fuzzy_subsets}).}
  \[
    \denot = \fbigcup\limits_{\alpha:\V\to\dom}\denota.
  \]
\end{definition}
\cref{def:osf_term_denotation} is a direct generalization of the crisp
denotation of an \osf term in an \osf interpretation under an assignment
$\alpha$
\cite{AitKaci1993b}. Note that the crisp denotation of a term is always a
singleton or the empty set, while $\denota[t]$ has a value grater than 0
for at most one element $d\in\dom$.

\begin{example}[Denotation of an \osf term]
\label{ex:osf_term_denotaion}
  Continuing from \cref{ex:fuzzy_osf_interpretation}, let $t$ be the term
  $\X:\thriller \left( \dirby \to \Y:\director \right)$
  and $\alpha$ be an assignment such that $\alpha(\X) = \halloween$ and
  $\alpha(\Y) = \carpenter$. Then (with $h \defeq \halloween$ and $c \defeq
  \carpenter$):
  \[%
    \begin{array}[b]{lllllllll}
      \denota[t](h) & = & \cfun[\{ \alpha(\X) \}](h) & \land & \thriller^\ii(h) & \land & \multicolumn{3}{l}{\denota[\Y:\director](\dirby^\ii(h))} \\
                    & = & 1                          & \land & 0.5              & \land & \multicolumn{3}{l}{\denota[\Y:\director](c)}             \\
                    & = & 1                          & \land & 0.5              & \land & \cfun[\{ \alpha(\Y) \}](c) & \land & \director^\ii(c) \\
                    & = & 1                          & \land & 0.5              & \land & 1                          & \land & 1 = 0.5.
    \end{array}
  \]
  Now consider the term $t'=
  \X:\movie \left( \dirby \to \Y:\director, \dirby\to \Y:\strng \right)$
  and note that $\director\fmeet\strng = \bots$. It follows that the
  denotation of $t'$ in any fuzzy \osf interpretation $\ii$ is always equal
  to $0$, i.e., the term $t'$ is contradictory.
  Indeed, suppose towards a contradiction that
  $\denota[t'](d)> 0$ for some \osf interpretation $\ii=\osfa$, $d\in\dom$
  and valuation $\alpha$. Then it must be the case that
  $\director^\ii(d') \land \strng^\ii(d') > 0$ or, equivalently,
  $(\director^\ii\fcap\strng^\ii)(d') > 0$,
  where $d' = \alpha(\Y)$.
  From \cref{def:osf_algebra} it follows that
  $(\director\fmeet\strng)^\ii(d') = \bots^\ii(d') > 0$,
  which is impossible since $\bots^\ii = \cf[\emptyset]$.%
\end{example}

\begin{remark}[Denotation of an \osf term]
  \label{rem:osf_term_denotation}
  Let $t=\osfterm$ be an \osf term and $\Xj[i] = \rtag(t_i)$ for each $i$.
  Let $\ii$ be an \osf interpretation and $d\in\dom$. It is easy to
  see that, if $\alpha_0$ and $\alpha_1$ are assignments such that
  \[
    \begin{array}{lll}
      \denota[t][\alpha_0](d) = \beta_0 >0 &
      \text{and} &
      \denota[t][\alpha_1](d) = \beta_1 >0,
    \end{array}
  \]
  then it must be the case that $\alpha_0(\Y) = \alpha_1(\Y)$ for all
  $\Y\in\tags(t)$, and thus $\beta_0 = \beta_1$.
  Indeed, it must be the case that $\alpha_0(\X) = \alpha_1(\X) = d$, and
  so for each $\fj[i]$ is must hold that $\alpha_0(\Xj[i]) = \fji[i](d) =
  \alpha_1(\Xj[i])$, and so on for all subterms.
  Hence the set $\{ \beta \mid \denota(d) = \beta > 0 \text{ for some }
  \alpha:\V\to\dom\}$ is a singleton and if $\denot(d) = \beta>0$, then there exists
  some $\alpha$ such that $\denota(d) = \beta$.
\end{remark}

\subsection{Satisfaction of \osf clauses in an \osf algebra}
\label{sec:constraint_semantics}

We now define the graded satisfaction of an \osf clause in a fuzzy \osf
interpretation, generalizing the crisp definition from \cite{AitKaci1993b}.
\begin{definition}[Satisfaction of an \osf clause in an \osf algebra]
\label{def:osf_clause_satisfaction}
  If $\ii = (\dom, \cdot^\ii)$ is a fuzzy \osf interpretation and $\osfval$
  is an assignment, then the satisfaction of an \osf clause $\phi$ to a degree
  $\beta\in[0,1]$ in the
  interpretation $\ii$ under the assignment $\alpha$ (notation:
  $\ii,\alpha\models_\beta\phi$) is defined recursively as follows:
  \[
    \begin{array}{lll}
      \ii,\alpha\models_\beta \xs & \Leftrightarrow &
      \s^\ii(\alpha(\X))\geq\beta,\\
      \addlinespace[1mm]
      \ii,\alpha\models_\beta \X\doteq\Y & \Leftrightarrow &
      \cf[\{\alpha(\X)\}](\alpha(\Y))\geq\beta,\\
      \addlinespace[1mm]
      \ii,\alpha\models_\beta \X.\f\doteq\Y & \Leftrightarrow &
      \cf[\{\f^\ii(\alpha(\X))\}](\alpha(\Y))\geq\beta,\\
      \addlinespace[1mm]
      \ii,\alpha\models_\beta \phi\osfwith\phi'& \Leftrightarrow &
      \ii,\alpha\models_\beta\phi \text{ and }
      \ii,\alpha\models_\beta\phi'.
    \end{array}
  \]
  If $\iimodb\phi$, then $\alpha$ is called a \textit{solution}
  in $\ii$
  for the clause $\phi$ with degree $\beta$, and $\phi$ is said to be
  \textit{satisfiable} in $\ii$ with degree $\beta$. The clause $\phi$ is
  said to be satisfiable in $\ii$ if there is some $\alpha:\V\to\dom$ and
  $\beta>0$ such that $\iimodb\phi$, and $\phi$ is said to be satisfiable
  if there is some $\ii$ such that $\phi$ is satisfiable in $\ii$.
\end{definition}
Note that $\ii,\alpha\models_0\phi$ always holds for any fuzzy \osf
interpretation $\ii$, assignment $\alpha$ and \osf clause $\phi$ (the
satisfaction degree of a clause for a given assignment and interpretation
is always greater than or equal to $0$). Moreover,
for any
$\beta\in(0,1]$,
\[
  \begin{array}{lllllll}
    \iimodb\X\doteq\Y    & \Iff & \cf[\{\alpha(\X)\}](\alpha(\Y))\geq\beta         & \Iff & \alpha(\X) = \alpha(\Y)         & \Iff & \iimod_1\X\doteq\Y,\\
    \iimodb\X.\f\doteq\Y & \Iff & \cf[\{\f^\ii(\alpha(\X))\}](\alpha(\Y))\geq\beta & \Iff & \f^\ii(\alpha(\X)) = \alpha(\Y) & \Iff & \iimod_1\X.\f\doteq\Y,
  \end{array}
\]
and
\[
  \begin{array}{lllll}
    \alpha(\X)\neq\alpha(\Y) & \To & \cf[\{\alpha(\X)\}](\alpha(\Y)) = 0 & \To & \ii,\alpha\models_0 \X\doteq\Y,\\
    \f^\ii(\alpha(\X))\neq\alpha(\Y) & \To & \cf[\{\f^\ii(\alpha(\X))\}](\alpha(\Y)) = 0 & \To & \ii,\alpha\models_0 \X.\f\doteq\Y.
  \end{array}
\]

\myrestatable%
  {Denotation and satisfaction in subalgebras}
  {prop}
  {prop:subalgebrasem}
  {proof:subalgebra_sem}
  {propsubalgebrasem}
  {%
    Let $\ii$ be a subalgebra of $\jj$. Then, for every \osf term $t$,
    every \osf clause $\phi$, every assignment $\alpha:\V\to\dom[\ii]$ and
    every $\beta\in[0,1]$:
    (i) $\denota[t](d) = \denoti[t][\jj](d)$ for all $d\in\dom$, and (ii)
    $\ii, \alpha\models_\beta
    \phi$ if and only if $\jj, \alpha\models_\beta \phi$.%
  }

We now provide the connection between the denotation of an \osf term $t$ in a
fuzzy interpretation and the degree of satisfaction of the corresponding \osf
clause $\phi(t)$.
\myrestatable%
  {Equivalence of terms denotation and constraints satisfaction} %
  {prop} %
  {prop:equivalence_terms_constraints} %
  {proof:equivalence_constraints} %
  {propequivalencetermsconstraints} %
  {%
  For every \osf term $t$ (with root variable $\X$),
  every interpretation $\ii$,
  every assignment $\alpha$,
  and every $\beta\in[0,1]$:
    $\denota[t](\alpha(\X)) \geq \beta \;\Iff\; \iimodb \phi(t)$.
  Therefore
    $\denota[t](\alpha(\X)) = \sup(\{\beta\mid \ii,\alpha\models_{\beta} \phi(t)\})$
  and thus, for all $d\in\dom$:
    $\denot(d) =
    \sup(\{\beta\mid\alpha\in\val(\ii)~\text{and}~\alpha(\X)=d~\text{and}~\ii,
    \alpha\models_{\beta} \phi(t) \})$.%
  }
The constraint normalization rules of \cref{fig:osf_normalization} preserve the
satisfiability of an \osf clause.
\myrestatable%
  {Solution-preservation of {\osf} clause normalization} %
  {prop} %
  {prop:osf_clause_normalization_semantics} %
  {proof:preservation} %
  {propsolutionpreservation} %
  {%
  For any rule of \cref{fig:osf_normalization} with premise
  $\phi$ and conclusion $\phi'$,
  for every fuzzy \osf algebra $\ii$ and assignment $\alpha$:
  $\ii,\alpha\models_\beta \phi$ for some $\beta>0$ if and only if $\ii,
  \alpha\models_{\beta'}\phi'$ for some $\beta'>0$.%
}
Note that the degrees $\beta$ and $\beta'$ in the last proposition may be
not be the same, so that the degree of satisfaction of an \osf
constraint in an algebra $\ii$ according to some assignment $\alpha$ may be
different from the degree of satisfaction of its solved form in the same
algebra and according to the same assignment. Together with
\cref{prop:equivalence_terms_constraints}, this implies
that the denotation of an \osf term $t$ in a fuzzy \osf interpretation may
not coincide
with the denotation of its normal form $\psi$ obtained by
applying the constraint normalization procedure to $\phi(t)$. This
constitutes a significant departure from crisp \osf logic, where in any
\osf algebra
every \osf clause is equivalent to its solved form,
and
every \osf term has the same denotation as its normal form.

\subsection{\osf graphs and fuzzy \osf graph algebra}

We introduce yet another class of objects that are in 1-to-1 correspondence
with normal \osf terms and rooted solved \osf clauses, namely that of \osf
graphs, originally defined in \cite{AitKaci1993b}.

\begin{definition}[\osf graph \cite{AitKaci1993b}]
\label{def:osf_graph}
  An \textit{\osf graph} is a directed labeled graph
    $g = \ograph$
  such that
  \begin{itemize}
    \item $N\subseteq\V$ and $\X\in N$ is a distinguished node called the
      \textit{root} of $g$;
    \item $\lambda_N: N\to\S$ is a node labeling function such that each
      node of $g$ is labeled by a non-bottom sort, i.e.,
      $\lambda_N(N)\subseteq\S\setminus\{ \bots \}$;
    \item $\lambda_E: E\to \F$ is an edge labeling function that assigns a
      feature to each edge $(\Y, \Z)\in E$ in such a way that
      no two edges outgoing from the same node are labeled by the same
      feature, i.e., if $\lambda_E(\Y, \Z) = \lambda_E(\Y, \Z[Z'])$ then
      $\Z = \Z[Z']$; and
    \item every node lies on a directed path starting at the root.
  \end{itemize}
\end{definition}

\begin{prop}[Equivalence of $\psi$-terms, rooted solved \osf clauses and
  \osf graphs \cite{AitKaci1993b}]
\label{prop:bijections}
  There exist bijective mappings
  $\pshi : \PhiR\to\Psi$, $\psig : \dom[\gg]\to\Psi$
  $G : \Psi\to\dom[\gg]$, and $\phi : \Psi\to\PhiR$,
i.e., such that
  $\id[\PhiR] = \phi\circ\pshi$,
  $\id[\Psi]  = \pshi\circ\phi = \psi_G\circ G$, and
  $\id[{\dom[\gg]}] = G\circ\psi_G$.
\end{prop}
We will write $\phi(g)$ instead of $\phi(\psi_G(g))$, or $G(\phi)$ instead
of $G(\pshi(\phi))$. We will also simply write $\psi(g)$ for $\psig(g)$, and
$\psi(\phi)$ for $\pshi(\phi)$. We refer the reader to
\cite{AitKaci1993b} for the definitions of these mappings.

\begin{example}[\osf graph]
\label{ex:osf_graph}
  Let $g = (N, E, \lambda_N, \lambda_E, \Xj[0])$ be such that
  \begin{itemize}
    \item $N = \{ \Xj[0], \Xj[1], \Xj[2], \X, \Y \}$;
    \item $E = \{
        (\Xj[0], \Xj[1]),
        (\Xj[0], \X),
        (\Xj[0], \Y),
        (\X, \Y),
        (\X, \Xj[2]),
        (\Y, \X)
      \}$;
    \item $\lambda_N = \{
        (\Xj[0], \s[movie]),
        (\Xj[1], \s[string]),
        (\Xj[2], \s[string]),
        (\X, \s[director]),
        (\Y, \s[writer])
      \}$; and
    \item $\lambda_E = \left\{
        \begin{array}{lll}
          ((\Xj[0], \Xj[1]), \f[title]),&
          ((\Xj[0], \X), \f[directed\_by]),&
        ((\Xj[0], \Y), \f[written\_by]),\\
          ((\X, \Y), \f[spouse]),&
          ((\X, \Xj[2]), \f[name]),&
        ((\Y, \X), \f[spouse])
        \end{array}
      \right\}$.
  \end{itemize}
  The \osf graph $g$ is depicted in \cref{fig:osf_graph}, where the root
  node is identified with a double ellipse, and the node
  identifiers (i.e., the variables) are omitted. Let $\psi$ be the \osf
  term from \cref{ex:osf_term} and $\phi$ be the \osf clause from
  \cref{ex:osf_clause}: then $g = G(\phi) = G(\psi)$, $\phi = \phi(g)$ and
  $\psi = \psi(g)$.
\end{example}

\newcommand{\mov}{$\s[movie]$}
\newcommand{\mst}{$\s[string]$}
\newcommand{\dir}{$\s[director]$}
\newcommand{\wri}{$\s[writer]$}
\newcommand{\dst}{$\s[string]$}
\newcommand{\dby}{$\f[directed\_by]$}
\newcommand{\wby}{$\f[written\_by]$}
\newcommand{\ttl}{$\f[title]$}
\newcommand{\nme}{$\f[name]$}
\newcommand{\spo}{$\f[spouse]$}
\newcommand{\xl}{-1.4}
\newcommand{\xc}{3}
\newcommand{\xr}{7}
\newcommand{\yl}{-2}
\newcommand{\yc}{0}
\newcommand{\yr}{2}

\begin{figure}[ht!]
  \centering
  \begin{tikzpicture}[node distance={20mm},thick,
                      main/.style = {thick,shape=ellipse,draw=sortcol},
                      root/.style = {thick,shape=ellipse,draw=sortcol,accepting},
                      box/.style = {draw,red,inner sep=10pt,rounded corners=5pt}]

    \node[root] (mov) at (\xl,\yc) {\mov};
    \node[main] (mst) at (\xc,\yr) {\mst};
    \node[main] (dir) at (\xc,\yc) {\dir};
    \node[main] (wri) at (\xc,\yl) {\wri};
    \node[main] (dst) at (\xr,\yc) {\dst};

    \draw[->,featcol] (mov) to[bend left=20] node[above,yshift=2mm] {\ttl} (mst);
    \draw[->,featcol] (mov) to node[above] {\dby} (dir);
    \draw[->,featcol] (mov) to[bend right=20] node[below,yshift=-2mm] {\wby} (wri);
    \draw[->,featcol] (dir) to node[above] {\nme} (dst);
    \draw[->,featcol] (dir) to[bend left=30] node[right] {\spo} (wri);
    \draw[->,featcol] (wri) to[bend left=30] node[left] {\spo} (dir);

  \end{tikzpicture}
  \caption{\osf graph of \cref{ex:osf_graph}}
  \label{fig:osf_graph}
\end{figure}

\osf graphs play a fundamental role in the semantics  of crisp \osf logic
\cite{AitKaci1993b} and,
as we will see, in that of our fuzzy generalization of this language. In
particular, they are the elements of the domain of the fuzzy \osf graph
algebra, a fuzzy \osf interpretation which will be essential for proving many
results about fuzzy \osf logic. This fuzzy interpretation is defined next.

\begin{definition}[Fuzzy \osf graph algebra]
\label{def:osf_graph_algebra}
  The \textit{fuzzy \osf graph algebra} is the pair $\gg = \osfg$ defined
  as follows.
  \begin{enumerate}
    \item The domain $\dom[\gg]$ is the set of all \osf graphs.
    \item For each $\s\in\S$ and for each graph $g = \ograph\in\dom[\gg]$:
      $\s^\gg(g) \defeq \fisop(\lambda_N(\X), \s)$.
    \item For each $\f\in\F$, the function $\f^\gg:\dom[\gg]\to\dom[\gg]$
      is defined by letting, for any $g = \ograph$:%
      \[
        \f^\gg(g) \defeq
        \begin{cases}
          \restr{g}{\Y} & \text{if}~\exists \Y\in N%
                          ~\text{such that}~\lambda_E(\X, \Y) = \f,\\
          \tgraph & \text{otherwise,}
        \end{cases}
      \]
      where
      \begin{itemize}
        \item $\restr{g}{\Y}$ is the maximally connected subgraph of $g$
          rooted at $\Y$; and
        \item $\tgraph$ denotes the trivial \osf graph
          $(\{ \tvar \}, \emptyset, \{ (\tvar, \tops) \}, \emptyset, \tvar)$
          whose only node is the new variable $\tvar\in\V\setminus N$ --
          labeled $\tops$ -- which is uniquely determined by the feature
          $\f$ and the graph $g$, i.e., if $\f\neq\f[f']$ or $g\neq g'$,
          then $\tvar\neq \tvar[f'][g']$.
      \end{itemize}
  \end{enumerate}
\end{definition}
This definition generalizes the \osf graph algebra of \cite{AitKaci1993b}: the
interpretation of feature symbols is the same, but the denotation of a sort
symbol $\s$ is now a fuzzy set whose value for a graph $g$ with root $\X$ is
the subsumption degree of the root sort $\lambda_N(\X)$ with respect to $\s$.

\begin{example}[\osf graph algebra]
  Consider the fuzzy subsumption relation corresponding to the graph of
  \cref{fig:fuzzy_sub_small}.
  Then, for example, $\thriller^\gg$ is the function such that:
  \begin{itemize}
    \item $\thriller^\gg(g) = 1$ for any \osf graph $g$ whose root is labeled
      by $\thriller$ or by $\bots$;%
    \item $\thriller^\gg(g) = \appdegree$ for any \osf graph $g$ whose root is
      labeled by $\slasher$; and
    \item $\thriller^\gg(g) = 0$ for any other \osf graph $g$.
  \end{itemize}

  \cref{fig:osf_graph_algebra} shows how features are interpreted in the
  fuzzy \osf graph algebra. \osf graphs are represented inside boxes.
  For a feature $\f$, the application of $\f^\gg$ to a graph $g$ is represented
  as an arrow originating from the box containing $g$
  and pointing at the box containing $\f^\gg(g)$.
  The figure shows the iterative application of a few
  features to the graph
  $g$ corresponding to the term
  \[
    \s[s_0]
    \left(
      \begin{array}{lll}
        \f[f_{2}] & \to & \s[s_1],\\
        \f[f_{0}] & \to & \X[X]:\s[s_2]
        \left(
          \begin{array}{lll}
            \f[f_{4}] & \to & \X[Y],\\
            \f[f_{3}] & \to & \s[s_1]
          \end{array}
        \right),\\
        \f[f_{1}] & \to & \X[Y]:\s[s_3]
        \left(
          \begin{array}{lll}
            \f[f_{4}] & \to & \X[X]
          \end{array}
        \right)
      \end{array}
    \right).
  \]
  Note that the result of applying the function $\f^\gg$ to a graph that
  does not contain this feature results in a trivial graph (e.g., the
  application of $\fjg[3]$ to $g$ or to $\fjg[1](g)$).
\end{example}
\renewcommand{\mov}{$\si[0]$}
\renewcommand{\mst}{$\si[1]$}
\renewcommand{\dir}{$\si[2]$}
\renewcommand{\wri}{$\si[3]$}
\renewcommand{\dst}{$\si[1]$}
\renewcommand{\dby}{$\fj[0]$}
\renewcommand{\wby}{$\fj[1]$}
\renewcommand{\ttl}{$\fj[2]$}
\renewcommand{\nme}{$\fj[3]$}
\renewcommand{\spo}{$\fj[4]$}
\renewcommand{\xl}{0}
\renewcommand{\xc}{2}
\renewcommand{\xr}{4}
\renewcommand{\yl}{-1.5}
\renewcommand{\yc}{0}
\renewcommand{\yr}{1.5}
\newcommand{\nfjg}[1]{\mymathcolor{featcol}{f}_{\mymathcolor{featcol}{#1}}^{\black{\gg}}}

\begin{figure}[ht!]
  \centering
  \begin{tikzpicture}[node distance={20mm},thick,
                      main/.style = {thick,shape=ellipse,draw=sortcol},
                      root/.style = {thick,shape=ellipse,draw=sortcol,accepting},
                      box/.style = {draw,red,inner sep=5pt,rounded corners=5pt}]

    \node[root] (mov) at (\xl,\yc) {\mov};
    \node[main] (mst) at (\xc,\yr) {\mst};
    \node[main] (dir) at (\xc,\yc) {\dir};
    \node[main] (wri) at (\xc,\yl) {\wri};
    \node[main] (dst) at (\xr,\yc) {\dst};

    \draw[->,featcol] (mov) to[bend left=20] node[above] {\ttl} (mst);
    \draw[->,featcol] (mov) to node[above] {\dby} (dir);
    \draw[->,featcol] (mov) to[bend right=20] node[below] {\wby} (wri);
    \draw[->,featcol] (dir) to node[above] {\nme} (dst);
    \draw[->,featcol] (dir) to[bend left=30] node[right] {\spo} (wri);
    \draw[->,featcol] (wri) to[bend left=30] node[right] {\spo} (dir);

    \node[box,fit=(mov)(mst)(dir)(wri)(dst)] (g)
    [label=above:$g$]
    {};

    \node[root] (dir0) at (\xc+5,\yc) {\dir};
    \node[main] (wri0) at (\xc+5,\yl) {\wri};
    \node[main] (dst0) at (\xr+5,\yc) {\dst};

    \draw[->,featcol] (dir0) to node[above] {\nme} (dst0);
    \draw[->,featcol] (dir0) to[bend left=30] node[right] {\spo} (wri0);
    \draw[->,featcol] (wri0) to[bend left=30] node[right] {\spo} (dir0);

    \node[box,fit=(dir0)(wri0)(dst0)] (gdir)
    [label=above:$\nfjg{0}(g)$]
    {};
    \draw[->,featcol] (g)   to node[above]  {$\nfjg{0}$} (gdir.west);

    \node[main] (dir1) at (\xc,\yc-4) {\dir};
    \node[root] (wri1) at (\xc,\yl-4) {\wri};
    \node[main] (dst1) at (\xr,\yc-4) {\dst};

    \draw[->,featcol] (dir1) to node[above] {\nme} (dst1);
    \draw[->,featcol] (dir1) to[bend left=30] node[right] {\spo} (wri1);
    \draw[->,featcol] (wri1) to[bend left=30] node[right] {\spo} (dir1);

    \node[box,fit=(dir1)(wri1)(dst1)] (gwri)
    [label=below:$\nfjg{1}(g)$]
    {};
    \draw[->,featcol] (g)   to node[left] {$\nfjg{1}$} (gwri.north);

    \node[root] (dst2) at (\xr+5,\yc-4) {\dst};
    \node[box,fit=(dst2)] (gdst)
    [label=below:$\nfjg{3}(\nfjg{0}(g))$]
    {};
    \draw[->,featcol] (gdir)   to node[right]  {$\nfjg{3}$} (gdst);

    \node[root] (wst1) at (\xr+2.5,\yc-5) {$\tops$};
    \node[box,fit=(wst1)] (wdstb)
    [label=below:$\nfjg{3}(\nfjg{1}(g))$]
    {};
    \draw[->,featcol] (gwri)   to node[above]  {$\nfjg{3}$} (wdstb);

    \node[root] (triv) at (\xl,\yc-4) {$\tops$};
    \node[box,fit=(triv)] (trivb)
    [label=below:$\nfjg{3}(g)$]
    {};
    \draw[->,featcol] (g)   to node[left]  {$\nfjg{3}$} (trivb);

    \node[root] (ttlmst) at (\xc+5,\yr) {\mst};
    \node[box,fit=(ttlmst)] (gttl)
    [label=above:$\nfjg{2}(g)$]
    {};

    \draw[->,featcol] (g)   to node[above]  {$\nfjg{2}$} (gttl.west);

    \draw[->,featcol] (gwri)   to[bend left=10] node[above] {$\nfjg{4}$} (gdir);
    \draw[->,featcol] (gdir)   to[bend left=10] node[below] {$\nfjg{4}$} (gwri);
  \end{tikzpicture}
  \caption{Feature application in the fuzzy \osf graph algebra.}
  \label{fig:osf_graph_algebra}
\end{figure}
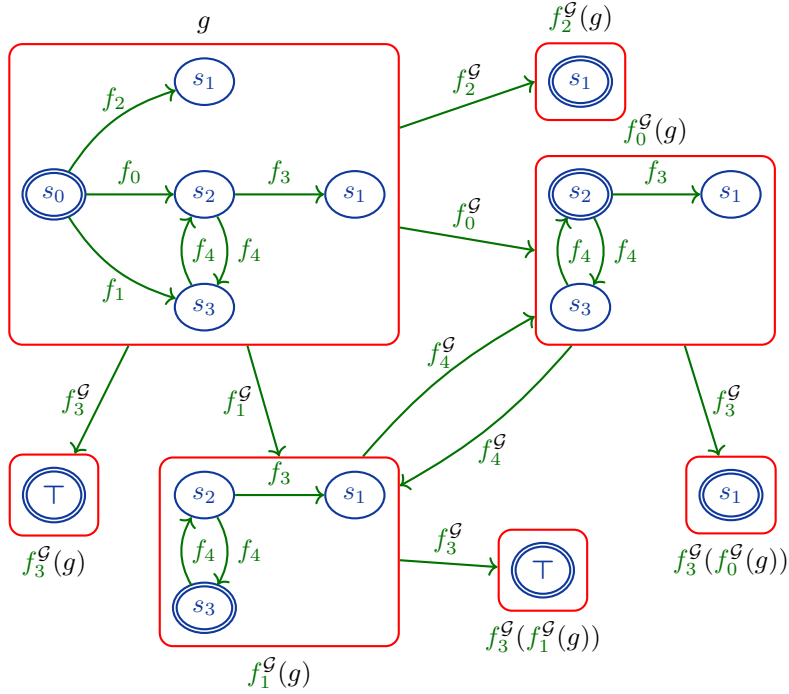

\myrestatable%
  {Fuzzy \osf graph algebra} %
  {prop} %
  {prop:graphalgebra} %
  {proof:fuzzy_osf_graph_algebra} %
  {propgraphalgebra} %
  {%
    The fuzzy \osf graph algebra of \cref{def:osf_graph_algebra} is a fuzzy
    \osf interpretation in the sense of \cref{def:osf_algebra}.%
  }

Every solved \osf clause $\phi$ is satisfiable in
a subalgebra of the fuzzy \osf graph algebra, namely the subalgebra %
generated by the graphs corresponding to the maximal rooted subclauses of
$\phi$.
\begin{definition}[Canonical graph algebra]
\label{def:canonical_graph_algebra}
  Let $\phi$ be a solved \osf clause. The subalgebra $\gg[\dom[\gg,\phi]]$ of
  the \osf algebra $\gg$ generated by
  $\dom[\gg,\phi] \defeq \{ G(\phi(\X)) \mid \X\in\osftags(\phi) \}$
  is called the \textit{canonical graph algebra} induced by $\phi$.
\end{definition}
\myrestatable%
  {Satisfiability} %
  {prop} %
  {thm:satisfiability} %
  {proof:satisfiability} %
  {thmsatisfiability} %
  {%
  Any solved clause $\phi\in\Phi$ is satisfiable in $\canon$
  with degree $1$ under any assignment $\alpha:\V\to\dom[\canon]$ such that
  $\alpha(\Y) = G(\phi(\Y))$ for all $\Y\in\tags(\phi)$.%
}

The following is a corollary of
\cref{thm:satisfiability,prop:subalgebrasem}.
\begin{corollary}[Canonical solution]
  \label{coro:canonical}
  Any solved \osf clause $\phi$ is satisfiable in the \osf graph
  algebra $\gg$ with degree $1$ under any assignment
  $\alpha:\V\to\dom[\gg]$ such that, for each
  $\Y\in\osftags(\phi(\X))$, $\alpha(\Y) = G(\phi(\Y))$.%
\end{corollary}

\section{Fuzzy \osf algebra homomorphisms}%
\label{sec:homomorphisms}

In this section we introduce fuzzy $\beta$-homomorphisms, mappings between
fuzzy \osf interpretations that preserve feature applications
and, to some degree,
the sorts of the elements of the domain.
Fuzzy $\beta$-morphisms allow us to prove several
results regarding the satisfaction of \osf clauses in fuzzy \osf
interpretations. In particular, we prove that the \osf graph algebra $\gg$
is canonical
in the sense that any \osf clause is satisfiable if and only if it is
satisfiable in $\gg$. We also prove that the denotation of a normal \osf
term $\psi$ in a fuzzy \osf interpretation can be characterized through the
existence of fuzzy homomorphisms from the subalgebra of $\gg$ generated by
$G(\psi)$.
In the next section, $\beta$-morphisms will be used to define a fuzzy
ordering on the domain of any fuzzy \osf interpretation,
which will eventually lead to the fuzzy subsumption ordering
between \osf terms.

\begin{definition}[Fuzzy \osf algebra $\beta$-homomorphism]
\label{def:osf_algebra_homomorphism}
  A \textit{$\beta$-morphism}
  (or \textit{$\beta$-homomorphism})
  $\gamma:\ii\to\jj$
  between two fuzzy \osf interpretations $\ii$
  and $\jj$ is a function $\gamma:\dom\to\dom[\jj]$ such that,
    for all $\f\in\F$, all $\s\in\S$, and all $d\in\dom$:
    $\gamma(\f^\ii(d)) = \f^\jj(\gamma(d))$ and
      $\s^\ii(d)\land \beta \leq \s^\jj(\gamma(d))$.
\end{definition}

Our definition of fuzzy \osf algebra $\beta$-morphism generalizes the
analogous crisp one from \cite{AitKaci1993b}. The condition on features
is unchanged: a $\beta$-morphism must preserve the structure of the input
algebra $\ii$, i.e., it must commute with function
applications. The original
crisp condition on sorts states that, whenever $d$ is an element of the
interpretation of $\s$ in $\ii$, then its image $\gamma(d)$ must be an
element of the interpretation of $\s$ in $\jj$. A direct fuzzy
generalization of this statement would specify that, for all
$\s\in\S$ and $d\in\dom[\ii]$,
\begin{align}
  \label{eq:morphism_direct}
  \s^\ii(d)\leq\s^\jj(\gamma(d)).
\end{align}
Similarly to our definition of a fuzzy subsumption relation,
we further generalize
this constraint by requiring that, in order for $\gamma$ to be a
$\beta$-morphisms, whenever $d$ is a member of $\s^\ii$ with degree
$\beta_0$, then $\gamma(d)$ must be a member of $\s^\jj$ with degree
greater than or equal to the minimum of $\beta$ and $\beta_0$. Clearly
\cref{eq:morphism_direct} is recovered simply by setting $\beta = 1$.

\begin{example}[Fuzzy \osf algebra $\beta$-homomorphism]
\label{ex:algebra_morphism}
  Consider
  the fuzzy interpretation $\ii$,
  the assignment $\alpha$
  and
  the term $t = \X:\thriller \left( \dirby \to \Y:\director \right)$
  from \cref{ex:osf_term_denotaion}. Consider the
  subalgebra of $\gg$ generated from the element $G(t)$ and define a
  function $\gamma:\dom[{\gg[G(t)]}] \to \dom[\ii]$ by setting, for $g =
  w^\gg(G(t))\in\dom[{\gg[G(t)]}]$, $\gamma(g) \defeq w^\ii(\halloween)$,
  where $w\in \F^*$. In particular we have $\gamma(G(t)) = \halloween$ and
  $\gamma(G(\Y:\director)) = \carpenter$. This is easily verified to be a
  $\appdegree$-morphism.
  This is depicted in \cref{fig:algebra_morphism} (where trivial graphs are
  not shown, and some names have been shortened).
\end{example}
\newcommand{\basez}{0}
\newcommand{\baseo}{-1.5}
\colorlet{intcol}{orange}
\begin{figure}[ht!]
  \centering
  \begin{tikzpicture}[node distance={20mm},thick,
                      main/.style = {thick,shape=ellipse,draw=sortcol},
                      elem/.style = {},
                      root/.style = {thick,shape=ellipse,draw=sortcol,accepting},
                      box/.style = {draw,red,inner sep=5pt,rounded corners=5pt},
                      outbox/.style = {draw,intcol,fit margins,rounded corners=5pt}]

    \node[root] (1) at (0,\basez) {$\thriller$};
    \node[main] (2) at (3.5,\basez) {$\director$};
    \node[root] (6) at (8,\basez) {$\director$};

    \node[elem] (3) at (0.5,\baseo) {$\halloween$};
    \node[elem] (4) at (2,\baseo-1.5) {$\carpenter$};
    \node[elem] (8) at (4.5,\baseo) {$\psycho$};
    \node[elem] (9) at (6,\baseo-1.5) {$\hitchcock$};
    \node (10) at (8,\baseo-0.75) {$\cdots$};

    \draw[->,featcol] (1) to node[above] {$\f[dir\_by]$} (2);
    \draw[->,featcol] (3) to node[left,yshift=-1mm] {$\f[dir\_by]^\ii$} (4);
    \draw[->,featcol] (8) to node[left,yshift=-1mm] {$\f[dir\_by]^\ii$} (9);

    \node (gn) at (1.75, 0.77) {$g$};
    \node (fgn) at (8, 0.77) {$\featcol{dir\_by}^\gg(g)$};
    \node[box,fit=(1)(2)] (g) {};
    \node[box,fit=(6)] (fg) {};
    \node[outbox,fit=(g)(fg)(gn)] (gg) [label=above:${\textcolor{intcol}{\gg{}[G(t)]}}$] {};
    \node[outbox,fit=(3)(4)(8)(9)(10),fit margins] (ii) [label=right:$\textcolor{intcol}{\ii}$] {};

    \draw[->,featcol] (g) to node[above] {$\f[dir\_by]^{\black{\gg}}$} (fg);
    \draw[->] (g) to node[right] {$\gamma$} (3);
    \draw[->] (fg) to[bend right=20] node[near end,left] {$\gamma$} (4);
  \end{tikzpicture}
  \caption{Morphism of \cref{ex:algebra_morphism}}
  \label{fig:algebra_morphism}
\end{figure}

\begin{restatable}%
  [{\protect\hyperlink{proof:homs}{Homomorphisms}}]%
  {prop}{prophoms}
\label{prop:homs}
Let $\gamma:\ii\to\jj$ be a $\beta$-morphism.
\begin{enumerate}
  \item If $\gamma':\jj\to\kk$ is a $\beta'$-morphism, then
    $\gamma'\circ\gamma:\ii\to\kk$ is a $\beta\land\beta'$-morphism.
  \item For all $\beta'\leq\beta$: $\gamma$ is a $\beta'$ morphism.
  \item There is a maximum $\beta'$ such that $\gamma$ is a
    $\beta'$-morphism.
\end{enumerate}
\end{restatable}

In general it is not true that, for any fuzzy
interpretations $\ii$ and $\jj$,
there is a maximum $\beta$ such that there exists a $\beta$-morphism
$\gamma:\ii\to\jj$.
This property -- which will be valuable later  -- holds
however for specific
homomorphisms relating subalgebras generated by singletons.
\myrestatable%
  {{Homomorphisms originating from singletons}} %
  {prop} %
  {prop:homsel} %
  {proof:homsel} %
  {prophomsel} %
  {%
  Let $\ii$ and $\jj$ be fuzzy \osf interpretations and fix $d\in\dom$. Let
  $\gamma:\ii[d]\to\jj$ be a $\beta$-morphism.
  \begin{enumerate}
    \item
      For every $d'\in\ii[d]$, $\gamma(d')$ is an element of
      the domain of $\jj[\gamma(d)]$, i.e.,
      $\gamma:\ii[d]\to\jj[\gamma(d)]$.
    \item For any $\beta'$-homomorphism $\gamma':\ii[d]\to\jj$: if
      $\gamma'(d) = \gamma(d)$, then $\gamma = \gamma'$, i.e., for all
      $d'\in\dom[{\ii[d]}]$, $\gamma(d') = \gamma'(d')$.
    \item
      There is a maximum
      $\beta'$ such that there exists a $\beta'$-morphism
      $\gamma':\ii[d]\to\jj$ that satisfies $\gamma'(d) = \gamma(d)$.%
  \end{enumerate}%
  }

As should be expected, the satisfiability of an \osf clause is preserved
(modulo degree $\beta$) under \osf algebra $\beta$-morphisms.

\myrestatable%
  {Extending solutions}
  {prop}
  {thm:extending}
  {proof:extending}
  {thmextending}
  {%
    Let $\ii$ and $\jj$ be two fuzzy \osf
    interpretations and $\gamma:\ii\to\jj$
    be a $\beta$-morphism. For every \osf clause $\phi$ and assignment
    $\alpha:\V\to\dom$, if
    $\iimod_{\beta_\ii}\phi$, then
    $\jj,\gamma\circ\alpha\models_{\beta_\ii\land\beta}\phi$.%
  }

An interesting property that carries over from crisp \osf logic
\cite{AitKaci1993b} is that it is always possible to define a
$\beta$-morphism from any fuzzy interpretation $\ii$ into the
fuzzy \osf graph algebra for some positive \mbox{degree
$\beta$.}

\begin{restatable}%
  [{\protect\hyperlink{proof:finality}
    {Fuzzy morphisms into {$\gg$}}}]%
  {theorem}{thmfinality}
\label{thm:weak_finality}
For any fuzzy \osf interpretation $\ii$ there exists a
$\beta$-homomorphism into the fuzzy \osf graph algebra $\gg$ for some
$\beta\in(0,1]$.
\end{restatable}

The following corollary follows directly from the last result and
\cref{thm:extending}.
\myrestatable%
  {Canonicity of {$\gg$}}
  {corollary}
  {cor:canonicity}
  {proof:canonicity}
  {corcanonicity}
  {%
  An \osf clause is satisfiable if and only if it is satisfiable in the
  fuzzy \osf graph algebra.%
  }

Another interesting property is that
any solution $\alpha$ for a clause $\phi$ in any fuzzy
interpretation $\ii$
can be obtained as a homomorphism from the canonical graph algebra induced
by $\phi$, as stated next.
An example of the application of this theorem was given in
\cref{ex:algebra_morphism}.

\begin{restatable}%
  [{\protect\hyperlink{proof:extracting}{Extracting solutions}}]%
  {theorem}{thmextracting}
\label{thm:extracting}
  For any solved \osf clause $\phi$, fuzzy interpretation
  $\ii$, assignment
  $\alpha:\V\to\dom$ and $\beta\in(0,1]$ such that $\iimodb\phi$
  there exists an \osf algebra
  $\beta$-homomorphism $\gamma:\canon\to\ii$ such that $\alpha(\X) =
  \gamma(G(\phi(\X)))$ for each $\X\in\osftags(\phi)$.
\end{restatable}

Thanks to \cref{prop:equivalence_terms_constraints,thm:extracting} we can
show that the denotation
of a normal \osf term $\psi$ in a fuzzy \osf interpretation can be
characterized through the existence of fuzzy homomorphisms from the
canonical graph algebra induced by $G(\psi)$.
\begin{restatable}%
  [{\protect\hyperlink{proof:interpretability}
  {Denotation of $\psi$-terms via fuzzy morphisms}}]%
  {theorem}{thminterpretability}
\label{thm:interpretability}
  Let $\psi$ be a normal \osf term, let $\phi = \phi(\psi)$ and
  let $\ii$ be a fuzzy \osf interpretation. Then, for all
  $d\in\dom$
    and $\beta\in(0,1]$:
  \[
    \denot[\psi](d) \geq\beta \;\Iff\;~\text{there is a $\beta$-morphism}~
    \gamma:\canon\to\ii~\text{such that}~d=\gamma(G(\psi))
  \]
  and thus
    $\denot[\psi](d) =
    \sup(\{ \beta\in[0,1]
    \mid \exists\text{$\beta$-morphism}~
  \gamma:\canon\to\ii~\text{such that}~d=\gamma(G(\psi)) \})$.
\end{restatable}

\section{Fuzzy \osf orderings and subsumption}%
\label{sec:subsumption}

In this section we define a fuzzy approximation ordering between \osf
graphs, a fuzzy subsumption ordering between \osf terms and a graded
implication ordering between \osf clauses. We show that these fuzzy
relations are equivalent and that they constitute fuzzy partial
orders%
\footnote{To be more precise, these fuzzy orders are antisymmetric on
  \textit{equivalence classes} of \osf graphs, normal \osf terms and rooted
  solved \osf clauses, as detailed later.}
on, respectively, \osf graphs, normal \osf terms and rooted solved \osf
clauses.
We give both a \textit{semantic} and \textit{syntactic} definition of fuzzy
subsumption of \osf terms, and prove that they are equivalent. We also
provide a link between the crisp definition of subsumption of \osf terms
\cite{AitKaci1993b} and our fuzzy generalization.

We begin by defining a fuzzy preorder on the elements
of any fuzzy \osf interpretation.

\renewcommand{\basez}{0}
\renewcommand{\baseo}{-3.5}
\colorlet{intcol}{orange}
\begin{figure}[ht!]
  \centering
  \begin{tikzpicture}[node distance={20mm},thick,
                      main/.style = {thick,shape=ellipse,draw=sortcol},
                      elem/.style = {},
                      root/.style = {thick,shape=ellipse,draw=sortcol,accepting},
                      box/.style = {draw,red,inner sep=5pt,rounded corners=5pt},
                      outbox/.style = {draw,intcol,fit margins,rounded corners=5pt}]

    \node[root] (1) at (0,\basez) {$\thriller$};
    \node[main] (2) at (3.5,\basez) {$\director$};
    \node[root] (6) at (8,\basez) {$\director$};
    \node[root] (7) at (12,\basez) {$\tops$};

    \node[root] (3) at (0,\baseo) {$\slasher$};
    \node[main] (4) at (3.5,\baseo+1) {$\director$};
    \node[main] (5) at (3.5,\baseo-1) {$\s[string]$};
    \node[root] (8) at (8,\baseo) {$\director$};
    \node[root] (9) at (12,\baseo) {$\s[string]$};

    \draw[->,featcol] (1) to node[above] {$\f[dir\_by]$} (2);
    \draw[->,featcol] (3) to node[sloped,above] {$\f[dir\_by]$} (4);
    \draw[->,featcol] (3) to node[sloped,above] {$\f[title]$} (5);

    \node (gn) at (1.75, 0.77) {$g_0$};
    \node (fgn) at (8, 0.77) {$\featcol{dir\_by}^\gg(g_0)$};
    \node[inner sep=0pt] (topsn) at (12.1, 0.77) {$\featcol{title}^\gg(g_0)$};
    \node[inner sep=0pt] (titlen) at (7.42,1.4) {$\f[title]^{\black{\gg}}$};
    \node[box,fit=(1)(2)] (g) {};
    \node[box,fit=(6)] (fg) {};
    \node[box,fit=(7)] (tops) {};
    \node[outbox,fit=(g)(fg)(gn)(topsn)(titlen)] (gg) [label=above:${\textcolor{intcol}{\gg{}[g_0]}}$] {};
    \node[inner sep=0pt] (g0n) at (1.75,\baseo-1.77) {$g_1$};
    \node (fg0n) at (8,\baseo-0.83) {$\featcol{dir\_by}^\gg(g_1)$};
    \node (tops0n) at (12,\baseo-0.83) {$\featcol{title}^\gg(g_1)$};
    \node[box,fit=(3)(4)(5)] (g0) {};
    \node[box,fit=(8)] (g0dir) {};
    \node[box,fit=(9)] (g0str) {};
    \node[outbox,fit=(g0)(g0dir)(g0str)(g0n),fit margins] (ii) [label=above:$\textcolor{intcol}{{\gg[g_1]}}$] {};

    \draw[->,featcol] (g0) to node[above] {$\f[dir\_by]^{\black{\gg}}$} (g0dir);
    \draw[->,featcol] (g) to node[above] {$\f[dir\_by]^{\black{\gg}}$} (fg);
    \draw[->,featcol] (g) to[bend left=15] node[above] {} (tops.north west);
    \draw[<-] (g0) to node[left,yshift=1mm] {$\gamma$} (g);
    \draw[<-] (g0dir) to[near end] node[left] {$\gamma$} (fg);
    \draw[<-] (g0str) to[near end] node[left] {$\gamma$} (tops);
    \draw[->,featcol] (g0.south east) to[bend right=10] node[below] {$\f[title]^{\black{\gg}}$} (g0str);
  \end{tikzpicture}
  \caption{Morphism of \cref{ex:endomorphic_approximation}}
  \label{fig:approx}
\end{figure}

\begin{definition}[Endomorphic approximation]
\label{def:endomorphic_approximation}
On each fuzzy \osf interpretation $\ii$ a fuzzy binary relation
$\fapproximates[\ii]:\dom\times\dom\to[0,1]$ is defined by letting, for all
$d, d'\in\dom$:
\[
  \fapproximates[\ii](d, d') =
  \sup(\{ \beta \mid
  \gamma(d) = d'~\text{for some $\beta$-homomorphism}~\gamma:\ii[d]\to\ii[d']
\}).
\]
If $\fapproximates[\ii](d, d') = \beta$ (abbreviated as
$d\fapprel d'$) we say that
$d$ \textit{approximates} $d'$ with degree $\beta$.
\end{definition}

\begin{example}[Endomorphic approximation]
\label{ex:endomorphic_approximation}
Let
\begin{align*}
  g_0 &= G(\Xj[0]:\thriller \left( \dirby \to \Yj[0]:\director \right))~\text{and}\\
  g_1 &= G(\Xj[1]:\slasher \left(\dirby \to \Yj[1]:\director, \ttle\to\Zj[1]:\s[string] \right)).
\end{align*}
Define $\gamma:\dom[{\gg[g_0]}]\to\dom[{\gg[g_1]}]$ by letting, for each $g
= w^\gg(g_0)$ in $\dom[{\gg[g_0]}]$ (with $w\in\F^*$): $\gamma(g) =
w^\gg(g_1)$. This is depicted in \cref{fig:approx} (where not all
trivial graphs are shown, and some names are shortened).
In particular $\gamma(g_0) = g_1$. The function
$\gamma$ is a $\appdegree$-morphism witnessing
$\fapproximates(g_0, g_1) = 0.5$.
\end{example}

\begin{remark}
  Let $\ii$ be a fuzzy \osf algebra and $d,d'\in\dom[\ii]$. Suppose that
  $\fapproximates[\ii](d, d') = \beta' > 0$. Then \cref{prop:homsel}
  guarantees that there exists a $\beta'$-morphism
  $\gamma':\ii[d]\to\ii[d']$ such that $\gamma'(d) = d'$, i.e., that
  $\beta'=\sup(\{ \beta \mid \gamma(d) = d'~\text{for some
    $\beta$-morphism}~\gamma:\ii[d]\to\ii[d']\})$ is a maximum.
\end{remark}

\begin{restatable}%
  [{\protect\hyperlink{proof:proppreorder}{Endomorphic approximation fuzzy preorder}}]%
  {prop}{proppreorder}
  \label{prop:preorder}
  For all fuzzy $\osf$ interpretations $\ii$, the fuzzy binary relation
  $\fapproximates[\ii]$ is a fuzzy preorder.
\end{restatable}

Like the subsumption lattice of first-order terms
\cite{Reynolds1970,Plotkin1969},
which is antisymmetric modulo variable renaming, our fuzzy partial orders
on \osf graphs, \osf terms and \osf clauses will only satisfy antisymmetry
between equivalence classes of these objects.
Our situation is slightly complicated by the fact that features are
interpreted as total functions, and thus, for instance, there exist \osf
terms that are semantically equivalent (i.e., they have the same denotation
in every fuzzy \osf interpretation)
without being variable renamings of each other.
An example is given by the terms
$\psi_0 = \Xj[0]:\s(\f\to\Yj[0]:\su)$
and
$\psi_1 = \Xj[1]:\s(\f\to\Yj[1]:\su, \f[f']\to\Zj[1]:\tops)$.
We now define an equivalence relation on \osf graphs that takes
care of this issue.

\begin{definition}[\osf graph equivalence]
\label{def:osf_graph_equivalence}
  Two \osf graphs $g_0$ and $g_1$ are \textit{equivalent} (notation: $g_0
  \gequiv g_1$) if there are
  1-morphisms
  $\gamma_0: \gg[g_0]\to\gg[g_1]$ and
  $\gamma_1: \gg[g_1]\to\gg[g_0]$ such that
  $\gamma_0(g_0) = g_1$
  and
  $\gamma_1(g_1) = g_0$.
\end{definition}
Informally, two graphs $g_0$ and $g_1$ are equivalent if they are
essentially the same graph \textit{modulo trivial subgraphs}, i.e., $g_0$
may contain subgraphs consisting of a single node labeled $\tops$ that are
not in $g_1$, and vice versa.

\begin{example}[\osf graph equivalence]
  \label{ex:graph_equivalence}
  Consider the $\osf$ terms
  \begin{align*}
    \psi_0 = \Xj[0]:\s(\f\to\Yj[0]:\su)~\text{and}~
    \psi_1 = \Xj[1]:\s(\f\to\Yj[1]:\su, \f[f']\to\Zj[1]:\tops)
  \end{align*}
  and let
  $g_0 = G(\psi_0)$ and
  $g_1 = G(\psi_1)$.
  Let $\gamma_0: \gg[g_0]\to\gg[g_1]$ be defined by letting, for all
  $g\in\gg[g_0] = w^\gg(g_0)$ for some $w\in\F^*$, $\gamma_0(g) =
  w^\gg(g_1)$. Similarly,
  let $\gamma_1: \gg[g_1]\to\gg[g_0]$ be defined by letting, for all
  $g\in\gg[g_1] = w^\gg(g_1)$ for some $w\in\F^*$, $\gamma_1(g) =
  w^\gg(g_0)$. In particular $\gamma_0(g_0) = g_1$ and $\gamma_1(g_1) =
  g_0$. These are easily verified to be $1$-morphisms, and thus $g_0\gequiv
  g_1$. This is depicted in \cref{fig:graph_equivalence}.
\end{example}

\renewcommand{\basez}{0}
\renewcommand{\baseo}{-3}
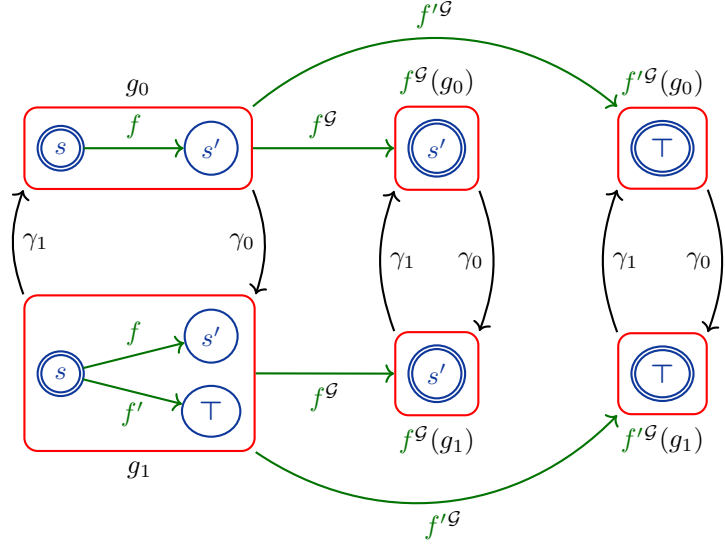
\begin{figure}[ht!]
  \centering
  \begin{tikzpicture}[node distance={20mm},thick,
                      main/.style = {thick,shape=ellipse,draw=sortcol},
                      root/.style = {thick,shape=ellipse,draw=sortcol,accepting},
                      box/.style = {draw,red,inner sep=5pt,rounded corners=5pt}]

    \node[root] (1) at (0,\basez) {$\s$};
    \node[main] (2) at (2,\basez) {$\su$};
    \node[root] (6) at (5,\basez) {$\su$};
    \node[root] (7) at (8,\basez) {$\tops$};

    \node[root] (3) at (0,\baseo) {$\s$};
    \node[main] (4) at (2,\baseo+.5) {$\su$};
    \node[main] (5) at (2,\baseo-.5) {$\tops$};
    \node[root] (8) at (5,\baseo) {$\su$};
    \node[root] (9) at (8,\baseo) {$\tops$};

    \draw[->,featcol] (1) to node[above] {$\f$} (2);
    \draw[->,featcol] (3) to node[above] {$\f$} (4);
    \draw[->,featcol] (3) to node[below] {$\fu$} (5);

    \node[box,fit=(1)(2)] (g0) [label=above:$g_0$] {};
    \node[box,fit=(6)] (fg0) [label=above:$\f^\gg(g_0)$] {};
    \node[box,fit=(7)] (fpg0) [label=above:$\fu^\gg(g_0)$] {};
    \node[box,fit=(3)(4)(5)] (g1) [label=below:$g_1$] {};
    \node[box,fit=(8)] (fg1) [label=below:$\f^\gg(g_1)$] {};
    \node[box,fit=(9)] (fpg1) [label=below:$\fu^\gg(g_1)$] {};

    \draw[->,featcol] (g0) to node[above] {$\f^{\black{\gg}}$} (fg0);
    \draw[->,featcol] (g1) to node[below] {$\f^{\black{\gg}}$} (fg1);
    \draw[->,featcol] (g0.north east) to[bend left=40] node[above] {$\fu^{\black{\gg}}$} (fpg0.north west);
    \draw[->,featcol] (g1.south east) to[bend right=40] node[below] {$\fu^{\black{\gg}}$} (fpg1.south west);

    \draw[->] (g0.south east)   to[bend left=20] node[left]  {$\gamma_0$} (g1.north east);
    \draw[->] (g1.north west)   to[bend left=20] node[right] {$\gamma_1$} (g0.south west);
    \draw[->] (fg0.south east)  to[bend left=20] node[left]  {$\gamma_0$} (fg1.north east);
    \draw[->] (fg1.north west)  to[bend left=20] node[right] {$\gamma_1$} (fg0.south west);
    \draw[->] (fpg0.south east) to[bend left=20] node[left]  {$\gamma_0$} (fpg1.north east);
    \draw[->] (fpg1.north west) to[bend left=20] node[right] {$\gamma_1$} (fpg0.south west);

  \end{tikzpicture}
  \caption{\osf graphs and morphisms of \cref{ex:graph_equivalence}}
  \label{fig:graph_equivalence}
\end{figure}

We can thus prove that
$\fapproximates$
is a fuzzy partial ordering on \osf graphs modulo \osf graph equivalence.
\begin{restatable}%
  [{\protect\hyperlink{proof:proppartial}{Endomorphic graph approximation fuzzy partial order}}]%
  {theorem}{proppartial}
  \label{prop:partial}
  The fuzzy binary relation
  $\fapproximates[\gg]$ is a fuzzy partial order on (equivalence classes
  of) \osf graphs, i.e., if
  $\fapproximates[\gg](g_0 , g_1) >0$ and
  $\fapproximates[\gg](g_1 , g_0)>0$, then $g_0\gequiv g_1$.
\end{restatable}

The equivalence relation on \osf graphs induces analogous
equivalence relations on normal \osf terms and rooted solved \osf
constraints by letting
$\psi_0\gequiv\psi_1 \Iff G(\psi_0)\gequiv G(\psi_1)$, and
$\phi_0\gequiv\phi_1 \Iff G(\phi_0)\gequiv G(\phi_1)$.
As expected, in every fuzzy \osf interpretation
equivalent normal \osf terms have the same denotation, and
  equivalent rooted solved \osf clauses are satisfied with the same
degree.
\myrestatable%
  {Denotation of equivalent \osf terms} %
  {prop} %
  {prop:equivalence} %
  {proof:equivalence} %
  {propequivalence} %
  {%
  If the normal \osf terms $\psi_0$ and $\psi_1$ are
  equivalent,
  then
  $\denot[\psi_0] = \denot[\psi_1]$
  for every fuzzy interpretation $\ii$.%
  }

\myrestatable%
  {Satisfaction of equivalent \osf clauses} %
  {prop} %
  {prop:equivalence_clause} %
  {proof:equivalence_clause} %
  {propequivalenceclause} %
  {%
  If the rooted solved \osf clauses $\phi_0$ and $\phi_1$ are
  equivalent,
  then
  for every fuzzy interpretation $\ii$
  and every $\beta\in[0,1]$:
  $\ii,\alpha_0\models_\beta\phi_0$ for some assignment $\alpha_0$ if and only if
  $\ii,\alpha_1\models_\beta\phi_1$ for some assignment $\alpha_1$.%
  }

We now give a semantic definition of fuzzy \osf term subsumption.
Recall that the fuzzy sort subsumption
$\fisop(\slasher, \thriller) = \beta$ means that, on any interpretation
$\ii$,
every object that is an instance of $\slasher^\ii$ with degree $\beta'$
must also be an instance of $\thriller^\ii$ with degree greater than or
equal to $\beta\land\beta'$.
Along these lines, we may say that an \osf term $t_1$ is subsumed by a term
$t_2$ with degree $\beta$ if, on any interpretation $\ii$, every object in
the denotation of $t_1$ with degree $\beta'$ also belongs to the
denotation of $t_2$ with degree
greater than or equal to
$\beta'\land\beta$. We thus define
fuzzy $\osf$ term subsumption as the fuzzy relation that assigns to each
pair of $\osf$ terms the supremum of all degrees $\beta\in[0,1]$ that
satisfy this property.

\begin{definition}[Semantic \osf term subsumption]
\label{def:sem_osf_term_subsumption}
  The sort subsumption relation $\fisop:\S^2\to[0,1]$ is extended to a
  fuzzy binary relation on \osf terms by letting,
  for all \osf terms $t_1$ and $t_2$:
    $\fisop(t_1, t_2) = \sup(\{ \beta \mid \forall \ii, \forall
    d\in\dom[\ii]: \denot[t_1](d)\land\beta\leq\denot[t_2](d) \})$.
  We abbreviate $\fisop(\psi_1, \psi_2) = \beta$ by writing
  $\psi_1\fisa_\beta\psi_2$.
\end{definition}
We provide an analogous definition for the graded implication of $\osf$
clauses, which generalizes the crisp one from \cite{AitKaci1993b}.
\begin{definition}[\osf clause implication]
\label{def:osf_clause_implication}
  The \textit{\osf clause $\phi_1$ implies the \osf clause $\phi_2$ at
  degree $\beta$} if,
  for all fuzzy \osf interpretations $\ii$ and assignments $\alpha$ such that
  $\ii,\alpha\models_{\beta_\ii} \phi$, there exists an assignment
  $\alpha'$ such that:
  (i) $\forall \X\in\tags(\phi_1)\cap\tags(\phi_2)$: $\alpha'(\X) =
      \alpha(\X)$, and
  (ii) $\ii, \alpha'\models_{\beta_\ii\land\beta} \phi_2$.
  The fuzzy binary relation $\models$ on \osf clauses is defined by letting
    $\mathop{\models}(\phi_1, \phi_2) = \sup(\{\beta \mid
    \phi_1~\text{implies}~\phi_2~\text{at degree}~\beta\})$.
  We abbreviate $\mathop{\models}(\phi_1, \phi_2) = \beta$ by writing
  $\phi_1\models_\beta\phi_2$.
\end{definition}

\begin{definition}[Rooted \osf clause implication]
\label{def:rooted_osf_clause_implication}
  Let $\phi_{\X}$ and $\phi'_{\Y}$ be two rooted \osf clauses with no common
  variables. Then:
  $\phi_{\X}\models_\beta \phi'_{\Y}$ if and only if
  $\phi\models_\beta \phi'[{\X}/{\Y}]$.
\end{definition}

We now prove that the fuzzy relations $\fapproximates$ on \osf graphs,
$\fisop$ on normal \osf term and $\models$ on rooted solved \osf
constraints are equivalent.
\begin{restatable}%
  [{\protect\hyperlink{proof:transparencybis}
  {Equivalence of fuzzy \osf orderings}}]%
  {theorem}{thmtransparencybis}
\label{thm:transparencybis}
  If
  the normal \osf terms $\psi$ and $\psi'$
  (with roots $\Y$ and $\X$, respectively, and no common variables),
  the \osf graphs $g$ and $g'$, and
  the rooted solved \osf clauses $\phi_{\Y}$ and $\phi'_{\X}$
  respectively correspond to one another though the syntactic mappings,
  then the following are equivalent:
  (1) $g \fapprel g'$,
  (2) $\psi'\fisa_\beta \psi$, and
  (3) $\phi'_{\X}\models_\beta \phi_{\Y}$.
\end{restatable}

The fact that $\fisop$ and $\models$ are fuzzy partial orders on normal
\osf terms and rooted solved \osf clauses (modulo \osf term and \osf clause
equivalence), respectively, is obtained as a corollary of
\cref{prop:partial,thm:transparencybis}.
\begin{corollary}
  The fuzzy relation $\fisa$ is a fuzzy partial order on normal \osf terms
  modulo \osf term equivalence. The fuzzy relation $\models$ is a fuzzy
  partial order on rooted solved \osf clauses modulo \osf clause
  equivalence.
\end{corollary}

Next, we provide a \textit{syntactic} definition of \osf term subsumption
and prove that it is equivalent to the semantic one of
\cref{def:sem_osf_term_subsumption}.
The syntactic definition
-- which was originally presented in \cite{Milanese2022} --
will be useful in \cref{sec:unification} for
  the computation of the subsumption degree between two \osf terms.

\begin{definition}[Syntactic \osf term subsumption]
\label{def:syn_osf_term_subsumption}
  \textit{The normal \osf term $\psi_0$ is (syntactically) subsumed by
    the normal \osf term $\psi_1$ with degree $\beta$} (denoted
    $\psi_0\fsynisa_\beta \psi_1$) if
    there is a mapping
    $h: \tags(\psi_1) \to \tags(\psi_0)$
    such that
  \begin{enumerate}
    \item $h(\rtag(\psi_1)) = \rtag(\psi_0)$\label{osfsubitm:1};
    \item if $\X\fto_{\psi_1}\Y$, then
      $h(\X)\fto_{\psi_0}h(\Y)$\label{osfsubitm:3}; and
    \item  $\beta = \min\{ \fisop(\sort_{\psi_0}(h(\X)),
      \sort_{\psi_1}(\X)) \mid \X\in\tags(\psi_1)\}$\label{osfsubitm:2}.
  \end{enumerate}
  We write $\fsynisop(\psi_0, \psi_1)\geq\beta$
  to express that
  $\psi_0\fsynisa_{\beta'} \psi_1$ and $\beta'\geq\beta$.
\end{definition}

\begin{remark}
  Syntactic \osf term subsumption is well-defined. Indeed, if
  $h: \tags(\psi_1) \to \tags(\psi_0)$ and
  $h': \tags(\psi_1) \to \tags(\psi_0)$
  are two mappings that satisfy
  \begin{enumerate}
    \item $h(\rtag(\psi_1)) = \rtag(\psi_0) = h'(\rtag(\psi_1))$, and
    \item if $\X\fto_{\psi_1}\Y$, then $h(\X)\fto_{\psi_0}h(\Y)$ and
      $h'(\X)\fto_{\psi_0}h'(\Y)$,
  \end{enumerate}
  then necessarily $h = h'$, which means that the value $\beta$ in
  \cref{def:syn_osf_term_subsumption} is unique.
\end{remark}

\begin{restatable}%
  [{\protect\hyperlink{proof:synsem}{Semantic and syntactic subsumption}}]%
  {theorem}{propsynsem}
  \label{prop:synsem}
  Let $\psi_0$ and $\psi_1$ be two
  normal $\osf$ terms. Then, for all $\beta\in(0,1]$:
  $\psi_0\fisa_\beta\psi_1$ if and only if there are two (normal) \osf terms
  $\psi_0'$ and $\psi_1'$ such that
    $\psi_0 \gequiv \psi_0'$,
    $\psi_1 \gequiv \psi_1'$, and
    $\psi_0'\fsynisa_\beta \psi_1'$.
\end{restatable}

Let $\isop \defeq \supp[\fisop]$ and recall that $\isa$ is a (crisp)
subsumption relation on $\S$ such that GLBs in $(\S, \isa)$ correspond to
GLBs in $(\S, \fisa)$ (see \cref{prop:glbs}).
The next theorem provides the connection between the crisp subsumption
$\isa$ and the fuzzy subsumption $\fisa$ on \osf terms.
\begin{restatable}%
  [{\protect\hyperlink{proof:crispfuzzy}{Crisp and fuzzy subsumption}}]%
  {theorem}{thmcrispfuzzy}
\label{thm:crispfuzzy}
For all normal \osf terms $\psi_1$ and $\psi_2$:
$\psi_1\isa\psi_2$ if and only if $\fisop(\psi_1, \psi_2)>0$.
\end{restatable}

The next corollary follows from \cref{thm:crispfuzzy} and the
analogous result for crisp \osf logic.
\begin{corollary}[Fuzzy \osf term lattice]
  The fuzzy binary relation $\fisop$ is a fuzzy lattice on (equivalence
  classes) of normal \osf terms.
\end{corollary}

\section{Fuzzy \osf term unification}%
\label{sec:unification}

Unification is an essential operation in automated reasoning. In crisp \osf
logic, the unification of two \osf terms allows to find their GLB in the
\osf term subsumption lattice. This operation is at the core of the
application of \osf logic and its variants in computational linguistics
\cite{Carpenter1992} and, more recently, it has enabled the implementation
of the very efficient CEDAR Semantic Web reasoner
\cite{AitKaciAmir2017,AmirAitKaci2017}.
In this section we prove that computing the GLB of two \osf terms in the
fuzzy subsumption lattice is no more difficult than computing
it in the crisp setting, as the same unification procedure can be employed,
and we also provide the complexity of computing the subsumption degree of
one term with respect to another.

\begin{definition}[Fuzzy \osf term unification]
\label{def:fuzzy_osf_term_unification}
  The \textit{unifier} of two normal \osf terms $\psi_1$ and $\psi_2$ is
  their GLB in the \osf term fuzzy subsumption lattice (modulo \osf term
  equivalence) and is denoted $\psi_1\fmeet\psi_2$.
  The \textit{unification degree} of $\psi_1$ and $\psi_2$ is defined as
  $\min(\fisop(\psi_1\fmeet\psi_2, \psi_1), \fisop(\psi_1\fmeet\psi_2,
  \psi_2))$.
  We write $\psi = \psi_1\fmeet_\beta\psi_2$ if $\psi$ is the
  unifier of $\psi_1$ and $\psi_2$ with unification degree $\beta$.
\end{definition}

Recall that applying the \osf constraint normalization rules
to an \osf clause results in a normal form $\phi$ that is either
the inconsistent clause $\X:\bots$, or an \osf clause in solved form
together with a conjunction of equality constraints
(\cref{prop:osf_clause_normalization}). The subclause of
$\phi$ in solved form is denoted $\solved(\phi)$.
The following theorem is an immediate consequence of the analogous result
for crisp \osf logic \cite{AitKaci1993b}, \cref{prop:glbs,thm:crispfuzzy}.

\begin{theorem}[Fuzzy \osf term unification]
\label{thm:fuzzy_osf_term_unification}
  Let $\psi_1$ and $\psi_2$ be \osf terms with no common variables, and let
  $\phi$ be the \osf clause obtained by non-deterministically applying any
  applicable \textit{constraint normalization rule}
  (\Cref{fig:osf_normalization}) to the clause $\phi(\psi_1) \osfwith
  \phi(\psi_2) \osfwith \rtag(\psi_1) \doteq \rtag(\psi_2)$
  until none applies. Then, $\phi$ is the inconsistent clause iff the GLB
  of $\psi_1$ and $\psi_2$ is $\X:\bots$. If $\phi$ is not the
  inconsistent clause, then $\psi_1\fmeet \psi_2 = \psi(\solved(\phi))$.
\end{theorem}

\begin{algorithm}[t!]
    \caption{Compute the unifier of $\psi_1$ and $\psi_2$ and their unification degree}
    \label{alg:unif}
    \hspace*{\algorithmicindent} \textbf{Input} $\psi_1, \psi_2\in \Psi$\\
    \hspace*{\algorithmicindent} \textbf{Output}
    $(\psi,\beta)$ such that $\psi = \psi_1 \fmeet_\beta \psi_2$
    \begin{algorithmic}[1]
        \Procedure{Unify}{$\psi_1,\psi_2$}
        \State $\phi \gets
                     \phi(\psi_1) \osfwith
                     \phi(\psi_2) \osfwith
                     \rtag(\psi_1) \doteq \rtag(\psi_2)$\label{line:2}
        \While{any rule $r$ of \cref{fig:osf_normalization} is applicable to
        $\phi$}\label{line:3}
          \State $\phi \gets $ result of applying $r$ to $\phi$\label{line:4}
        \EndWhile
        \If{$\phi$ is the inconsistent clause}\label{line:5}
          \State \textbf{return} $\X:\bots$ with unification degree $1$\label{line:6}
          \Else \label{line:7}
        \State $\phi' \gets \solved(\phi)$\label{line:8}
        \State $\osfeq = \{ (\X, \Y) \mid \X\doteq\Y~\text{or}~\Y\doteq\X~\text{is a conjunct of}~\phi\}^*$\label{line:9}
        \State $\quot[\osfeq][\osftags(\phi)] = \{ [\X] \mid \X\in \tags(\phi) \}$\label{line:10}
        \For{$\X\in\tags(\phi')$}\label{line:11}
          \State $\phi' \gets \phi'[\Z_{[\X]}/\X]$\label{line:12}
          \Comment{$\Z_{[\X]}$ is a new variable in $\V\setminus\tags(\phi)$}
        \EndFor
        \State $\psi\gets\psi(\phi')$\label{line:13}
          \Comment{Unifier $\psi = \psi_1\fmeet\psi_2$}
        \State $\beta_1 \gets \min\{
                \fisop(\sort_{\psi}(\Z_{[\X]}), \sort_{\psi_1}(\X))
              \mid \X\in\tags(\psi_1)\}$\label{line:14}
        \State $\beta_2 \gets \min\{ \fisop(\sort_{\psi}(\Z_{[\X]}),
                \sort_{\psi_2}(\X)) \mid \X\in\tags(\psi_2)\}$\label{line:15}
        \State $\beta \gets \min(\beta_1, \beta_2)$\label{line:16}
          \Comment{Unification degree $\beta\in[0,1]$}
        \State \textbf{return} ($\psi$, $\beta$)
        \EndIf
        \EndProcedure
    \end{algorithmic}
\end{algorithm}

  \cref{alg:unif} shows a procedure to unify two normal \osf terms and to
compute their unification degree\footnote{%
    The following notation is adopted in \cref{alg:unif}.
    The reflexive and transitive closure of a binary relation $R$ is
    denoted $R^*$.
    The expression
    $\quot[\osfeq][\tags(\phi)]$
    denotes the quotient of the set $\osftags(\phi)$ modulo the equivalence
    relation $\osfeq\subseteq\tags(\phi)^2$. For a variable $\X\in \tags(\phi)$, its equivalence
    class with respect to the relation
    $\osfeq$ is $[\X] \defeq \{ \Y \mid (\X, \Y)\in\osfeq \}$.}.
  Given two normal \osf terms $\psi_1$ and $\psi_2$, the algorithm proceeds
as follows.
\begin{itemize}
  \item
    \cref{line:2,line:3,line:4} involve the application of the
    constraint normalization rules of \cref{fig:osf_normalization} to the
    clause $\phi(\psi_1) \osfwith \phi(\psi_2) \osfwith \rtag(\psi_1)
    \doteq \rtag(\psi_2)$, resulting in an \osf clause $\phi$ in normal
  form.
  \item
    If this normal form is (equivalent to) the inconsistent clause,
  then $\X:\bots$ is returned with unification degree $1$\footnotemark{}
    on \cref{line:6}. Otherwise, the algorithm proceeds with the
    computation of the unification degree.
    \footnotetext{%
        Recall that by \cref{def:fuzzy_lattice} $\bots$ is subsumed by
    every sort with degree $1$.}
  \item
    In \cref{line:8,line:9,line:10,line:11,line:12}
    the set $\tags(\phi)$ is partitioned into equivalence classes
    according to the equality constraints contained in $\phi$, and each
    variable $\X$ in $\phi' = \solved(\phi)$ is uniformly renamed with a new variable
    $\Z_{[\X]} \in \V\setminus \tags(\phi)$ corresponding to its
    equivalence class. This step allows each term to maintain its own
    variable scope, and facilitates the computation of the unification
    degree in the next lines.
  \item
    Finally,
    the GLB $\psi = \psi_1\fmeet\psi_2$ is obtained on \cref{line:13}
    as the \osf term corresponding to $\phi'$,
    while \cref{line:14,line:14,line:15,line:16} deal with the
    computation of the unification degree $\beta$ of the two \osf terms.
\end{itemize}
The output of the algorithm is a pair $(\psi, \beta)$ such that
$\psi=\psi_1\fmeet_\beta\psi_2$.
Note that the unifier of the two terms can already be obtained on
\cref{line:8} as $\psi(\solved(\phi))$, and the subsequent steps are
only concerned with the computation of the unification degree.
The same algorithm can also be employed to decide whether the two terms
subsume each other, and to what degree:
if $\psi$ is equivalent to $\psi_1$, then $\fisop(\psi_1, \psi_2) =
\beta$; alternatively, if $\psi$ is equivalent to $\psi_2$, then
$\fisop(\psi_2, \psi_1) = \beta$.

\begin{restatable}%
  [{\protect\hyperlink{proof:algo}{Correctness of \cref{alg:unif}}}]%
  {theorem}{thmalgo}
\label{thm:algo}
  Let $\psi_1$ and $\psi_2$ be normal \osf terms.
  If $(\psi, \beta)$ is the output of the procedure
  $\textsc{Unify}(\psi_1, \psi_2$) of \cref{alg:unif},
  then $\psi = \psi_1\fmeet_\beta\psi_2$.
\end{restatable}

The following example clarifies each step of \cref{alg:unif}.

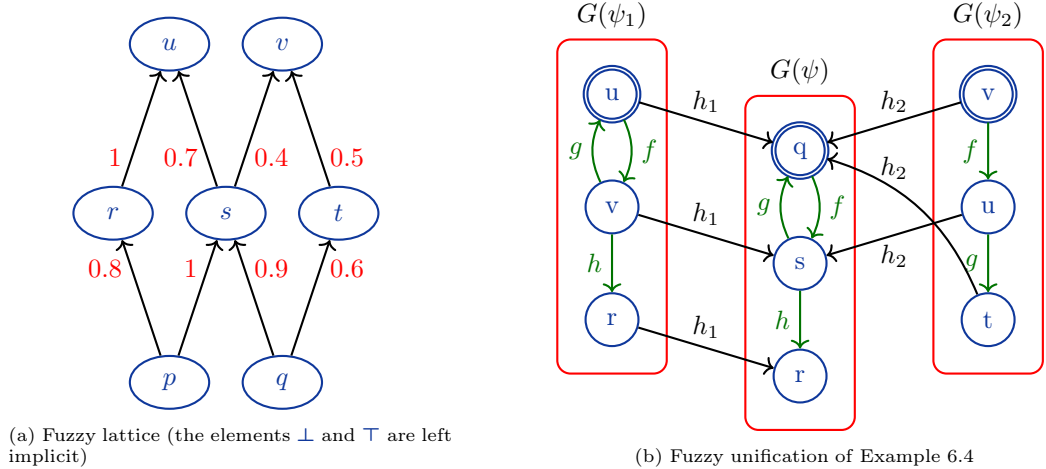
\begin{figure}[ht!]
  \centering
  \begin{subfigure}[b]{0.39\textwidth}
    \centering
    \begin{tikzpicture}[scale=1.5,
                    main/.style = {thick,
                                   shape=ellipse,
                                   draw=sortcol,
                                   minimum height=20pt,
                                   minimum width=30pt}]
    \node[main] (p) at (1, 0) {$\s[p]$};
    \node[main] (q) at (2, 0) {$\s[q]$};

    \node[main] (r) at (0.5, 1.5) {$\s[r]$};
    \node[main] (s) at (1.5, 1.5) {$\s[s]$};
    \node[main] (t) at (2.5, 1.5) {$\s[t]$};

    \node[main] (u) at (1, 3) {$\s[u]$};
    \node[main] (v) at (2, 3) {$\s[v]$};

    \draw[->,thick] (p) to node[left , near end]  {$\red{0.8}$} (r);
    \draw[->,thick] (p) to node[left , near end]  {$\red{1}$}   (s);
    \draw[->,thick] (q) to node[right, near end]  {$\red{0.9}$} (s);
    \draw[->,thick] (q) to node[right, near end]  {$\red{0.6}$} (t);
    \draw[->,thick] (r) to node[left , near start] {$\red{1}$}   (u);
    \draw[->,thick] (s) to node[left , near start] {$\red{0.7}$} (u);
    \draw[->,thick] (s) to node[right, near start] {$\red{0.4}$} (v);
    \draw[->,thick] (t) to node[right,near start]  {$\red{0.5}$} (v);
\end{tikzpicture}
    \caption{Fuzzy lattice
    (the elements $\bots$ and $\tops$ are left implicit)}
    \label{fig:wdag}
  \end{subfigure}
  \hfill
  \begin{subfigure}[b]{0.6\textwidth}
    \centering
    \begin{tikzpicture}[node distance={20mm},thick,
                    main/.style =
                    {thick,shape=ellipse,draw=sortcol,minimum height=20pt, minimum width=20pt},
                    root/.style = {thick,shape=ellipse,draw=sortcol,accepting,minimum height=20pt, minimum width=20pt},
                    box/.style = {draw,red,inner sep=10pt,rounded corners=5pt}]

  \node[root] (x0) at (5, 3)   {\s[v]};
  \node[main] (x1) at (5, 1.5) {\s[u]};
  \node[main] (x2) at (5, 0)   {\s[t]};

  \node[root] (z0) at (2.5, 2.25)   {\s[q]};
  \node[main] (z1) at (2.5, 0.75) {\s[s]};
  \node[main] (z2) at (2.5, -0.75)   {\s[r]};

  \node[root] (y0) at (0, 3)   {\s[u]};
  \node[main] (y1) at (0, 1.5) {\s[v]};
  \node[main] (y2) at (0, 0)   {\s[r]};

  \node[box,fit=(x0)(x1)(x2)] (gx) [label=above:$G(\psi_2)$] {};
  \node[box,fit=(y0)(y1)(y2)] (gy) [label=above:$G(\psi_1)$] {};
  \node[box,fit=(z0)(z1)(z2)] (gz) [label=above:$G(\psi)$] {};

  \draw[->,featcol] (x0) to node[left] {$\f$} (x1);
  \draw[->,featcol] (x1) to node[left] {$\f[g]$} (x2);

  \draw[->,featcol] (z0) to[bend left=25] node[right] {$\f$}    (z1);
  \draw[->,featcol] (z1) to[bend left=25] node[left] {$\f[g]$} (z0);
  \draw[->,featcol] (z1) to               node[left] {$\f[h]$} (z2);

  \draw[->,featcol] (y0) to[bend left=25] node[right] {$\f$}   (y1);
  \draw[->,featcol] (y1) to[bend left=25] node[left] {$\f[g]$} (y0);
  \draw[->,featcol] (y1) to               node[left] {$\f[h]$} (y2);

  \draw[->] (x0) to node[above] {$h_2$} (z0);
  \draw[->] (x1) to node[below] {$h_2$} (z1);
  \draw[->] (x2) to[bend right=25, pos=.66] node[above] {$h_2$} (z0);

  \draw[->] (y0) to node[above] {$h_1$} (z0);
  \draw[->] (y1) to node[above] {$h_1$} (z1);
  \draw[->] (y2) to node[above] {$h_1$} (z2);
\end{tikzpicture}
    \caption{Fuzzy unification of \Cref{ex:unif}}%
    \label{fig:exunif}
  \end{subfigure}
  \caption{Fuzzy subsumption relation and unification of \cref{ex:unif}}
  \label{fig:exunifmain}
\end{figure}

\newcommand{\zo}{\Z_{[\Xj[0]]}}
\newcommand{\zi}{\Z_{[\Xj[1]]}}
\newcommand{\zd}{\Z_{[\Yj[2]]}}
\begin{example}[Fuzzy \osf term unification]
  \label{ex:unif}
  Consider the fuzzy lattice of \Cref{fig:wdag} and
  the \osf terms
  $\psi_1 = \Yj[0] : \s[u] \left(
      \f[f] \to \Yj[1] : \s[v] \left(
        \f[g] \to \Yj[0],
        \f[h]\to \Yj[2]:\s[r] \right) \right)$
  and
  $\psi_2 = \Xj[0] : \s[v] \left(
      \f[f] \to \Xj[1] : \s[u] \left(
        \f[g] \to \Xj[2] : \s[t] \right) \right)$.
  After \cref{line:2,line:3,line:4},
  an application of the rules of \Cref{fig:osf_normalization} to
  $\phi(\psi_1)\osfwith\phi(\psi_2)\osfwith \Xj[0] \doteq \Yj[0]$
  yields the \osf clause in normal form
  \[
    \begin{array}{llllllll}
      \phi & = & \Xj[0]:\s[q]         & \osfwith & \Xj:\s                  & \osfwith & \Yj[2] : \s[r]     & \osfwith \\
           &   & \Xj[0].\f \doteq \Xj & \osfwith & \Xj.\f[g] \doteq \Xj[0] & \osfwith & \Xj.\f[h] = \Yj[2] & \osfwith \\
           &   & \Xj[0] \doteq\Yj[0]  & \osfwith & \Xj[0] \doteq \Xj[2]    & \osfwith & \Xj \doteq \Yj
    \end{array}
  \]
  or an equivalent clause.
  In \cref{line:9,line:10} the set $\osftags(\phi)$ is thus
  partitioned into the equivalence classes
  $[\Xj[0]] = \{ \Xj[0], \Xj[2], \Yj[0] \}$,
  $[\Xj[1]] = \{ \Xj[1], \Yj \}$
  and
  $[\Yj[2]] = \{ \Yj[2] \}$.
  The solved part of $\phi$ is renamed accordingly (\cref{line:11,line:12})
  and translated on \cref{line:13} into the \osf term
  \[
    \psi = \zo:\s[q] (\f\to\zi:\s(\f[g]\to\zo, \f[h]\to\zd:\s[r])).
  \]
  The subsumption degree $\fisa(\psi, \psi_1)$ is then computed on
  \cref{line:14} as
  $\beta_1 =
  \min\{ \fisa(\s[q], \s[u]),\allowbreak
         \fisa(\s, \s[v]),\allowbreak
         \fisa(\s[r], \s[r]) \} = 0.4$.
  In particular,
  $\fisop(\sort_{\psi}(\zi), \sort_{\psi_1}(\Yj[1])) =
  \fisop(\s[s], \s[v]) = 0.4$.
  Similarly, $\fisa(\psi, \psi_2)$ is computed on \cref{line:15} as
  $\beta_2 =
  \min\{ \fisa(\s[q], \s[v]),
         \fisa(\s, \s[u]),
         \fisa(\s[q], \s[t]) \} = 0.5$.
  Overall, the unification degree of $\psi_1$ and $\psi_2$ is thus $0.4$
  (\cref{line:16}).
  For $i\in \{ 1,2 \}$, the mapping $h_i:\tags(\psi_i)\to\tags(\psi)$
witnessing the (syntactic) subsumption $\psi \fsynisa_{\beta_i} \psi_i$
  is defined by letting $h_i(\X) = \Z_{[\X]}$ for all $\X\in\tags(\psi_i)$.
  The unification is depicted in \Cref{fig:exunif}, where the mappings
  $h_1$ and $h_2$ are depicted as arrows relating the nodes of $G(\psi_1)$,
  $G(\psi_2)$ and $G(\psi)$ corresponding to the variables of their
  respective \osf terms.
\end{example}

Finally, let us analyze \cref{alg:unif} in terms of complexity.
Finding the unifier of the two normal \osf terms $\psi_1$ and $\psi_2$ with
respect to a fuzzy sort subsumption lattice
has the same complexity of deciding the problem in the crisp setting.
This is due to the fact that GLBs in a fuzzy lattice and its crisp
counterpart can be computed in the same way (also see \cref{prop:glbs}),
and the rules for fuzzy \osf constraint normalization and crisp \osf
constraint normalization are essentially the same.
The algorithm from \cite{AitKaci1986b} is based
on the union-find problem \cite{Aho1974} and has a worst-case complexity of
$O(mG(m))$, where $m = \card{\tags(\psi_1)\cup\tags(\psi_2)}$ and
the growth rate of the function $G$ is of the order of an inverse of the
Ackermann function ($G(m)\leq 5$ for all practical purposes)
\cite{AitKaci1986b}.

Partitioning the set $\tags(\phi)$ according to the equality constraints in
$\phi$ (\cref{line:9,line:10}) is an application of the union-find problem.
The complexity is thus $O(mG(m))$, where $G$ is as above and $m =
\card{\tags(\psi_1)\allowbreak\cup\allowbreak\tags(\psi_2)}$.

 Computing $\fisop(\s, \su)$ for two sorts $\s,\su\in\S$ can be performed
 in $O(\card{\S}+e)$ time -- where $e$ is the number of edges in the DAG
 representation of the fuzzy sort subsumption relation -- with an approach
 that is analogous to solving the shortest paths problem in a DAG
 \cite{Milanese2021a}.
 The overall complexity of the computation of the subsumption degrees
 $\beta_1$ and $\beta_2$ in \cref{line:14,line:15} is thus
 $O(m(\card{\S}+e))$, where $m = \card{\tags(\psi_1)\cup\tags(\psi_2)}$.

\section{Conclusion}%
\label{sec:conclusion}

We have defined a fuzzy semantics of \osf logic in which sort
symbols denote fuzzy subsets of a domain of interpretation.
This semantics is based on a fuzzy sort subsumption lattice
$\fisop:\S\times\S\to[0,1]$ whose interpretation generalizes Zadeh's
inclusion of fuzzy sets by requiring that, whenever $\fisop(\s, \su) =
\beta$, then for any interpretation $\ii = \osfa$ and any element $d\in\dom$
the value of $\su^\ii(d)$ must be greater than or equal to the minimum of
$\s^\ii(d)$ and $\beta$.
As argued in the introduction, this notion of fuzzy subsumption may be
applied, for example, in fuzzy logic programming languages based on fuzzy
\osf term unification, or in similarity-based reasoning, where a crisp
subsumption relation $\isop\subseteq\S^2$ may be extended to a fuzzy
subsumption relation $\fisop$ according to a given similarity
$\simop:\S^2\to[0,1]$.%

The generalization to a fuzzy semantics provides \osf logic with
the capability to perform approximate reasoning.
In particular, we have shown how to decide whether two \osf terms are
subsumed by each other, and to which degree, via their unification.
The fact that the same unification procedure as crisp \osf logic can be
used to find the GLB of two \osf terms in the fuzzy subsumption lattice
constitutes a benefit in terms of developing fuzzy \osf logic
reasoners, since it would be possible to take advantage of existing
implementation techniques.%

  There are several avenues for future work. For instance, it would be
  interesting to study the semantics of fuzzy \osf logic under other
  t-norms besides the
  minimum t-norm adopted in this paper. Another direction could consist in
  developing a version of fuzzy \osf logic that supports a more expressive
  language
  that also includes negation and disjunction, and possibly
  in which features are
  interpreted as partial functions rather than total functions.
  Ultimately, the goal of this research would be to
  develop a fuzzy version of the CEDAR reasoner
  \cite{AitKaciAmir2017,AmirAitKaci2017} which is able to provide
  approximate answers to queries posed to a knowledge base by relying on a
  fuzzy sort subsumption relation or a sort similarity relation.%

\section*{Acknowledgements}%

The authors express their gratitude and a fond thought to Hassan \ak, who
with Gabriella Pasi set out to define a fuzzy version of OSF logic. This
paper originates from their work on the definition of similarity-based
unification for OSF terms, extending the approach of
\cite{AitKaciPasi2020}.

\appendix
\section*{Appendix}
\renewcommand{\thesection}{\Alph{section}}
\section{Table of symbols}
\label{app:table}

The main symbols and notation used in this paper are displayed in
\cref{tab:symbols} for reference.

\begin{figure}[ht!]
  \centering
  {\renewcommand{\arraystretch}{1.1}
  \begin{tabular}[t]{|l|l|}
    \cline{1-2}
    $\s, \su, \si[i], \ldots$                                     & Sort symbols in $\S$                                                \\ \cline{1-2}
    $\f, \fu, \fj[i], \f[g], \f[l] \ldots$                        & Feature symbols in $\F$                                             \\ \cline{1-2}
    $w$ and $w^\ii$                                               & Element $w\in\F^*$ and corresponding function composition on $\dom$ \\ \cline{1-2}
    $\X, \Y, \Z, \Xj[i], \X[X'], \ldots$                          & Variables (\osf tags) in $\V$                                       \\ \cline{1-2}
    \iffuzzy
    $\fisa: \S^2 \to [0,1]$                                       & Fuzzy subsumption partial order                                     \\ \cline{1-2}
    $\fmeet: \S^2 \to \S$                                         & Greatest lower bound operation on $\ftax$                           \\ \cline{1-2}
    \else
    $\mathop{\isa} \subseteq \S\times\S$                          & Subsumption partial order                                           \\ \cline{1-2}
    $\meet: \S^2 \to \S$                                          & Greatest lower bound operation on $\tax$                            \\ \cline{1-2}
    \fi
    $t, t_i, t', \ldots$                                          & \osf terms                                                          \\ \cline{1-2}
    $\psi, \psi_i, \psi', \ldots$                                 & Normal \osf terms ($\psi$-terms)                                    \\ \cline{1-2}
    $\Psi$                                                        & The set of all normal \osf terms                                    \\ \cline{1-2}
    $\phi, \phi_i, \phi', \ldots$                                 & \osf constraints and clauses                                        \\ \cline{1-2}
    $\phi(\X)$                                                    & Maximal subclause of $\phi$ rooted in $\X\in\tags(\phi)$            \\ \cline{1-2}
    $\Phi$                                                        & The set of all solved \osf clauses                                  \\ \cline{1-2}
    $\PhiR$                                                       & The set of all rooted solved \osf clauses                           \\ \cline{1-2}
    $g = \ograph, g_i, g', \ldots$                                & \osf graphs                                                         \\ \cline{1-2}
    $\psi_\phi:\PhiR\to\Psi$                                      & Bijection from rooted solved clauses to $\psi$-terms                \\ \cline{1-2}
    $\psi_G:\dom[\gg]\to\Psi$                                     & Bijection from \osf graphs to $\psi$-terms                          \\ \cline{1-2}
    $\phi:\Psi\to\PhiR$                                           & Bijection from $\psi$-terms to rooted solved clauses                \\ \cline{1-2}
    $G:\Psi\to\dom[\gg]$                                          & Bijection from $\psi$-terms to \osf graphs                          \\ \cline{1-2}
    $\alpha, \alpha_i, \alpha', \ldots$                           & Variable assignments $\alpha:\V\to\dom$                             \\ \cline{1-2}
    $\ii = \osfa$, $\jj = (\dom[\jj], \cdot^\jj)$, \ldots         & \iffuzzy{Fuzzy}~\fi \osf interpretations                            \\ \cline{1-2}
    $\ii[D]$                                                      & Subalgebra of $\ii$ generated by $D\subseteq\dom$                   \\ \cline{1-2}
    $\ii[d]$                                                      & Subalgebra of $\ii$ generated by $d\in\dom$                         \\ \cline{1-2}
    $\gg$                                                         & \iffuzzy Fuzzy~\fi  \osf graph algebra                              \\ \cline{1-2}
    $\Delta^{\gg,\phi} = \{ G(\phi(\X)) \mid \X\in\tags(\phi) \}$ & Set of all maximally connected subgraphs of $G(\phi)$               \\ \cline{1-2}
    $\canon$                                                      & Canonical graph algebra induced by $\phi$                           \\ \cline{1-2}
    $\approximates[\ii]$                                          & Endomorphic approximation ordering                                  \\ \cline{1-2}
    \iffuzzy
    $\land$                                                       & Minimum T-norm                                                      \\ \cline{1-2}
    $\fcup, \fcap$                                                & Union and intersection of fuzzy subsets                             \\ \cline{1-2}
    $\beta, \beta_i, \beta', \ldots$                              & Real values in $[0, 1]$                                             \\ \cline{1-2}
    $\cf[X]$                                                      & Characteristic function of the set $X$                              \\ \cline{1-2}
    $\id[X]$                                                      & Identity function on the set $X$                                    \\ \cline{1-2}
    \fi
    \cline{1-2}
  \end{tabular}
  }
  \caption{Table of symbols}%
  \label{tab:symbols}
\end{figure}

\section{Fuzzy set theory and fuzzy orders: notation and definitions}%
\label{app:fuzzy}

We recall the basic definitions of fuzzy set theory
  \cite{DuboisPrade1980} to fix the
notation.
Whenever possible, we use the same notation for fuzzy sets and fuzzy orders as the one
used for ordinary sets and orders, but with the addition of a small dot $(\cdot)$ to avoid
ambiguity. For example, the symbol for the intersection of two crisp sets
is $\cap$, while the one for the intersection of
two fuzzy sets is $\fcap$. We adopt the minimum t-norm (denoted
$\land$) and the maximum t-conorm (denoted $\lor$).

\subsection{Fuzzy sets}%

\begin{definition}[Fuzzy subset]
\label{def:fuzzy_subset}
A \emph{fuzzy subset} $F$ of a (crisp) set $X$ is determined by its
membership function $\mu_F : X \to [0, 1]$.
\end{definition}

\begin{definition}[Intersection and union of fuzzy subsets]
\label{def:intersection_of_fuzzy_subsets}
The \emph{intersection $\fbigcap \mathcal{F}$ of a set $\mathcal{F}$ of
fuzzy subsets of a set $X$} is defined by letting $\mu_{\fbigcap
\mathcal{F}}(x) \defeq \inf(\{ \mu_{F}(x) \mid F\in \mathcal{F} \})$.
The \emph{union $\fbigcup \mathcal{F}$ of a set $\mathcal{F}$ of fuzzy
subsets of a set $X$} is defined by letting $\mu_{\fbigcup \mathcal{F}}(x)
\defeq \sup(\{ \mu_{F}(x) \mid F\in \mathcal{F} \})$.
\end{definition}

\begin{definition}[Support]
\label{def:support}
  The \emph{support} of a fuzzy subset $F$ of $X$ is
  defined as $\supp \defeq \{ x\in X \mid \mu_F(x) > 0 \}$.
\end{definition}

\subsection{Fuzzy Binary Relations}%
\label{sec:fuzzy_binary_relations}

\begin{notation*}
  From now on the membership function of a fuzzy subset $F$ of $X$ will
  simply be written $F:X\to[0,1]$ instead of $\mu_F:X\to[0,1]$.
\end{notation*}

\begin{definition}[Fuzzy binary relation]
\label{def:fuzzy_binary_relation}
  A \emph{fuzzy binary relation} $R$ on a set $X$ is a fuzzy subset of
  $X\times X$, i.e., it is a function $R: X\times X \to [0, 1]$.
\end{definition}

\begin{definition}[Fuzzy preorder]
\label{def:fuzzy_preorder}
A fuzzy binary relation $R$ on a set $X$ is called a \emph{fuzzy preorder} if it
satisfies:
\begin{align*}
  \tag{Fuzzy Reflexivity} \label{eq:fuzzy_reflexivity}
  \forall x\in X,~&R(x, x) = 1,\\
  \tag{Max-Min Transitivity} \label{eq:fuzzy_transitivity}
    \forall x, y, z \in X,~&
    R(x, z) \geq R(x, y) \land R(y, z).
\end{align*}
\end{definition}

\begin{definition}[Fuzzy partial order]
\label{def:fuzzy_poset}
A fuzzy binary relation $R$ on a set $X$ is called a \emph{fuzzy partial order} if it
satisfies
\cref{eq:fuzzy_reflexivity,eq:fuzzy_transitivity} and
\begin{align*}
  \tag{Strong Fuzzy Antisymmetry} \label{eq:fuzzy_antisymmetry}
    \forall x, y \in X,~&
      \text{if}~R(x, y) > 0~\text{and}~R(y, x) > 0,
      \text{then}~x = y.
\end{align*}
The pair $\poset[R]$ is called a \emph{fuzzy partially ordered set} (\emph{fuzzy poset}).
\end{definition}

\begin{definition}[Composition of fuzzy binary relations]
\label{def:fuzzy_composition}
  The \emph{(max-min) composition} of two fuzzy binary relations $R$ and $Q$ on
  a finite set $X$ is the fuzzy binary relation $R\fcomp Q$ defined by the
  membership function
  \[
    {R\fcomp Q}(x, z) \defeq \bigvee_{y\in X}(R(x, y) \land Q(y, z)).
  \]
  The $n$-ary composition of a fuzzy binary relation $R$ with itself is defined by letting
  $R^1 \defeq R$ and $R^n \defeq R\fcomp R^{n-1}$ for $n>1$.
\end{definition}

\begin{definition}[Reflexive and transitive closure of a fuzzy binary relation]
\label{def:fuzzy_transitive_closure}
  The \emph{transitive closure} of a fuzzy binary relation $R$ %
  is defined as $R^\oplus \defeq \fbigcup_{m \geq 1}R^m$.
  The \emph{reflexive and transitive closure} $R^\oast$ of a fuzzy binary
  relation $R$ is obtained by letting $R^\oast(x, y) \defeq 1$ if $x=y$ and
  $R^\oast(x, y) \defeq R^\oplus(x, y)$ otherwise.
\end{definition}

\begin{notation*}
  From now on fuzzy preorders and fuzzy partial orders will be denoted by
  $\fisa$, and infix notation will also be used, i.e.,
  $x\fisa y$ will be written instead of $\fisop(x, y)>0$.%
\end{notation*}

We adopt the definitions of lower bounds and greatest lower bounds from
\cite{Chon2009,Mezzomo2013,Mezzomo2016}.

\begin{definition}[Lower bounds in a fuzzy poset]
\label{def:lower_bounds_in_a_fuzzy_poset}
  Let $\ftax$ be a fuzzy poset and $S \subseteq \S$. The set of
  \emph{(fuzzy) lower bounds of $S$} is defined as
  $\flb{S} \defeq
   \{ \s\in\S \mid \forall \su\in S, \s\fisa \su \}$.
\end{definition}

\begin{definition}[Fuzzy greatest lower bound]
\label{def:fuzzy_glb}
  Let $\ftax$ be a fuzzy poset and $S \subseteq \S$.
  The \emph{greatest lower bound (GLB) of $S$} is the unique
  $\s\in\flb{S}$ such that, for all $\su\in\flb{S}$, $\su\fisa \s$.
  If the GLB of $S$ exists, it is denoted
  $\fbigmeet{S}$, or simply $\s\fmeet\su$ in case $S = \{ \s, \su \}$.
\end{definition}

\begin{definition}[Fuzzy lattice and bounded lattice]
\label{def:fuzzy_lattice}
  A fuzzy poset $(\S, \fisa)$ is a \emph{fuzzy lattice} if every pair of
  elements has a GLB.
  A fuzzy lattice $(\S, \fisa)$ is \textit{bounded} if there are elements
  $\bots, \tops \in \S$ such that, for all $\s\in \S$, $\fisop(\bots, \s) =
  1$ and $\fisop(\s, \tops) = 1$.
\end{definition}

\begin{prop}[Fuzzy and crisp lattices]
\label{prop:glbs}
Let $(\S, \fisa)$ be a fuzzy lattice. Then $(\S, \supp[\fisop])$
 is a (crisp) lattice on $\S$.
Moreover, if $\meet$ is the GLB operation for $(\S, \supp[\fisop])$,
then
$\fbigmeet S = \bigmeet S$
for every subset $S\subseteq \S$.
\end{prop}
\ifallproofs
\begin{proof}
  Let $\isop \defeq \supp[\fisa]$.
  If $(\S, \fisa)$ is fuzzy poset, then $\isa$ is a reflexive,
  antisymmetric and transitive binary relation on $\S$:
  \begin{itemize}
      \item (Reflexivity) Since $\fisop(\s,\s) = 1$ for all $\s\in \S$,
        then $\s\isa\s$ for all $\s\in \S$;
      \item (Antisymmetry) If $\s\isa\s[s']$ and $\s[s']\isa \s$, then
        $\fisop(\s,\s[s']) > 0$ and $\fisop(\s[s'], \s) > 0$, and the
        antisymmetry of $\fisa$ gives $\s = \s[s']$;
      \item (Transitivity) If $\si[1]\isa\si[2]$ and $\si[2]\isa \si[3]$,
        then $\fisop(\si[1],\si[2]) > 0$ and $\fisop(\si[2], \si[3]) > 0$,
        so that max-min transitivity of $\fisa$ gives
        $\fisop(\si[1],\si[3])\geq \min(\fisop(\si[1],\si[2]),
        \fisop(\si[2],\si[3])) > 0$ and thus $\si[1]\isa\si[3]$.
  \end{itemize}
  Let $\lb{S} \defeq \{ \s \in \S \mid \s\isa \s[s'], \forall \s[s']\in S
  \}$ denote the set of lower bounds of $S\subseteq \S$ in
  $(\S, \isa)$,
  and note that $\lb{S} =\flb{S}$ for any $S\subseteq \S$.
  Let $S\subseteq \S$. Then
  \[
    \begin{array}[b]{lll}
      \s = \fbigmeet S & \Iff &
      \text{(i)}~\s\in\flb{S}~\text{and (ii)}~\forall\su\in\flb{S},\su\fisa\s\\
                        & \Iff &
      \text{(i)}~\s\in\lb{S}~\text{and (ii)}~\forall\su\in\lb{S},\su\isa\s\\
                        & \Iff & \s = \bigmeet S.
    \end{array}\qedhere
  \]
\end{proof}
\fi

\ifallproofs
\section{Proofs for \cref{cha:semantics}: \nameref{cha:semantics}}%
\label{proofs:sem}
\fi

\ifallproofs
\propsubalgebra*
\begin{proof}[Proof of \cref{prop:osf_subalgebra}]%
  \hypertarget{proof:subalgebra}{\mbox{}}%
  $\F^*(D)$ contains $D$ and is closed under feature application by
  construction.
  Clearly $\F^*(D)$ must be a subset of the domain of any subalgebra of
  $\ii$ containing $D$.

  Now we show that $\ii[D]$ satisfies all the conditions of
  \cref{def:osf_algebra}. Let
  $\si[0], \si[1]\in\S$ be arbitrary:
  \begin{itemize}
    \item Suppose that $\fisop(\si[0],\si[1]) = \beta$. Since $\ii$ is a
      fuzzy \osf interpretation it holds that, for any
      $d\in\dom[{\ii[D]}]\subseteq\dom$:
      $\sii[0](d)\land\beta\leq\sii[1](d)$.
      Thus
      $\sic[0](d)\land\beta = \sii[0](d)\land\cfun[\fcl](d) \land\beta \leq
      \sii[1](d)\land\cfun[\fcl](d) = \sic[1](d)$.
    \item Let $d\in \dom[{\ii[D]}] = \fcl$. %
      Then:
      \[
        \begin{array}[b]{lll}
          (\sic[0]\fcap \sic[1])(d) >0 & \To & \sic[0](d) > 0~\text{and}~\sic[1](d) >0 \\
                                       & \To & \sii[0](d) > 0~\text{and}~\sii[1](d)>0~\text{and}~\cfun[\fcl](d)>0\\
                                       & \To & (\sii[0]\fcap \sii[1])(d)>0~\text{and}~\cfun[\fcl](d)>0\\
                                       & \To & (\si[0]\fmeet\si[1])^\ii(d)> 0~\text{and}~\cfun[\fcl](d)>0\\
                                       & \To & (\si[0]\fmeet\si[1])^\ii(d)\land\cfun[\fcl](d)>0\\
                                       & \To & (\si[0]\fmeet \si[1])^{\ii[D]}(d) >0
        \end{array}\qedhere
      \]
  \end{itemize}
\end{proof}
\fi

\ifallproofs
\propsubalgebrasem*
\begin{proof}[Proof of \cref{prop:subalgebrasem}]%
  \hypertarget{proof:subalgebra_sem}{\mbox{}}%
    The first part is proved by induction on the structure of $t$.
    \begin{itemize}
      \item Let $t = \X:\s$ and let $d\in\dom$ be arbitrary. Then
        \[
          \begin{array}{llll}
            \denota[t](d) & = & \cfun[\{ \alpha(\X) \}](d) \land \s^\ii(d) &  \\
                          & = & \cfun[\{ \alpha(\X) \}](d) \land \s^\jj(d) & \text{(By \cref{def:fuzzy_osf_subalgebra})} \\
                          & = & \denoti[t][\jj](d)
          \end{array}
        \]
      \item Now let $t = \osfterm$ and let $d\in\dom$ be arbitrary.
        Then
        \[
          \begin{array}{llll}
            \denota[t](d) & = & \cfun[\{ \alpha(\X) \}](d) \land \s^\ii(d) \land\bigwedge\limits_{1\leq i\leq n} \denota[t_i](\fji[i](d)) & \\
                          & = & \cfun[\{ \alpha(\X) \}](d) \land \s^\jj(d) \land\bigwedge\limits_{1\leq i\leq n} \denoti[t_i][\jj](\fjj[i](d)) & \text{(By IH and \cref{def:fuzzy_osf_subalgebra})} \\
                          & = & \denoti[t][\jj]
          \end{array}
        \]
    \end{itemize}
    For the second part, note that $\alpha(\X)\in\dom[\ii]$ for all
    $\X\in\tags(\phi)$, and, for all $d\in\dom[\ii]$ and $\f\in\F$:
    $\f^\ii(d)\in\dom$. Then:
    \begin{itemize}
      \item $\ii,\alpha\models_\beta \X:\s
        \Iff
        \s^\ii(\alpha(\X))\geq\beta
        \Iff
        \s^\jj(\alpha(\X))\geq\beta
        \Iff
        \jj,\alpha\models_\beta \X:\s$.
      \item $\ii,\alpha\models_\beta \X.\f\doteq\Y
        \Iff
        \cf[\{\f^\ii(\alpha(\X))\}](\alpha(\Y)) \geq\beta
        \Iff
        \cf[\{\f^\jj(\alpha(\X))\}](\alpha(\Y)) \geq\beta
        \Iff
        \jj,\alpha\models_\beta \X.\f\doteq\Y$.
      \item $\ii,\alpha\models_\beta \X\doteq\Y
        \Iff
        \cf[\{\alpha(\X)\}](\alpha(\Y)) \geq\beta
        \Iff
        \jj,\alpha\models_\beta \X\doteq\Y$.\qedhere
    \end{itemize}
\end{proof}
\fi

\ifallproofs
\propequivalencetermsconstraints*
\begin{proof}[Proof of \cref{prop:equivalence_terms_constraints}]%
  \hypertarget{proof:equivalence_constraints}{\mbox{}}%
  \ifallproofs
  By induction on $t$. Let $t = \X:\s$ and thus $\phi(t) = \X:\s$. Then:
  \[
    \begin{array}{lllll}
     \denota[\X:\s](\alpha(\X))\geq\beta & \Iff & \s^\ii(\alpha(\X))\geq\beta
                                         & \Iff & \iimodb \X:\s.
    \end{array}
  \]
  Now let $t = \osfterm$, with $\Xj[i] = \rtag(t_i)$ for each $1\leq i\leq
  n$,
  so that $\phi(t) =
  \X:\s \osfwith
  \bigosfwith\limits_{1\leq i\leq n}
  (\phi(t_i) \osfwith \X.\fj[i] \doteq \Xj[i])$.
  Suppose that, for each $1\leq i\leq n$:
  \[
    \denota[t_i](\alpha(\Xj[i])) \geq \beta \;\Iff\; \iimodb \phi(t_i)~%
    (\text{induction hypothesis}).
  \]
  Hence
  \[
    \begin{array}{lll}
     \denota(\alpha(\X))\geq \beta & \Iff &
     \s^\ii(\alpha(\X))\geq\beta
     \text{ and }\\
                                   & & \forall 1\leq i \leq n:
     \cf[{\{\fji[i](\alpha(\X))\}}](\alpha(\Xj[i]))\geq\beta
     \text{ and }\denota[t_i](\alpha(\Xj[i]))\geq\beta\\
                                   & \Iff & \iimodb \X:\s \text{ and }
                                   \forall 1\leq i\leq n: \iimodb \X.\fj[i]
                                   \doteq \Xj[i] \osfwith \phi(t_i)\\
                                   & \Iff & \iimodb \phi(t).
    \end{array}
  \]
  \else
  The fact that $\denota(\alpha(\X))\geq \beta \Iff \iimodb \phi(t)$ holds
  can be proved by induction on $t$.
  \fi
  For the next part, note that
  \[\arraycolsep=3pt
    \begin{array}{lllll}
      \denota(\alpha(\X)) & = & \sup(\{\beta\in[0,1] \mid \denota(\alpha(\X)) \geq \beta\})
                          & = & \sup(\{\beta\in[0,1]\mid \ii,\alpha\models_{\beta} \phi(t) \}).
    \end{array}
  \]
  Finally, note that $\denot(d)$ can be rewritten as follows:
  \[
    \begin{array}[b]{lll}
      \denot(d) & = & \fbigcup\limits_{\alpha:\V\to\dom}\denota(d)\\
                & = & \sup(\{\beta\mid\alpha\in\val(\ii)~\text{and}~\denota(d) = \beta \})\\
                & = & \sup(\{\beta\mid\alpha\in\val(\ii)~\text{and}~\alpha(\X)=d~\text{and}~\denota(d) = \beta \})\\
                & = & \sup(\{\beta\mid\alpha\in\val(\ii)~\text{and}~\alpha(\X)=d~\text{and}~\denota(d) \geq \beta \})\\
                & = & \sup(\{\beta\mid\alpha\in\val(\ii)~\text{and}~\alpha(\X)=d~\text{and}~\ii, \alpha\models_{\beta} \phi(t) \}).
    \end{array}\qedhere
  \]
\end{proof}
\fi

\ifallproofs
\propsolutionpreservation*
\begin{proof}[Proof of \cref{prop:osf_clause_normalization_semantics}]%
  \hypertarget{proof:preservation}{\mbox{}}%
  The only rule worth considering is \osfref{osf:si}.
  Suppose that
  $\iimodb \X:\si[0]\osfwith \X:\si[1]$
  for some $\beta>0$.
  It follows that
  $\sii[0](\alpha(\X)) > 0$
  and
  $\sii[1](\alpha(\X))> 0$
  so that
  $(\sii[0]\fcap\sii[1])(\alpha(\X))>0$.
  By
  \cref{def:osf_algebra}
  then
  $(\si[0]\fmeet\si[1])^\ii(\alpha(\X))>0$,
  so that
  $\iimod_{\beta'} \X:\si[0]\fmeet\si[1]$ for some $\beta'>0$.

  With respect to the other direction,
  let $\s = \si[0]\fmeet\si[1]$ for convenience and
  suppose that
  $\iimod_{\beta} \X:\s$
  for some $\beta>0$, so that
  $\s^\ii(\alpha(\X)) > 0$.
  Let $\beta_0 = \fisop(\s, \si[0])>0$
  and $\beta_1 = \fisop(\s, \si[1])>0$. By
  \cref{def:osf_algebra} then
  $\s^\ii(\alpha(\X))\land\beta_0 \leq \sii[0](\alpha(\X))$ and
  $\s^\ii(\alpha(\X))\land\beta_1 \leq \sii[1](\alpha(\X))$, so that
  $\s^\ii(\alpha(\X))\land\beta_0\land\beta_1 \leq
  \sii[0](\alpha(\X)) \land \sii[1](\alpha(\X))$. Thus $\sii[0](\alpha(\X))
  \land \sii[1](\alpha(\X))>0$, so that $\iimod_{\beta'}
  \X:\si[0]\osfwith\X:\si[1]$ for some $\beta'>0$.
\end{proof}
\fi

\ifallproofs
\propgraphalgebra*
\begin{proof}[Proof of \cref{prop:graphalgebra}]%
  \hypertarget{proof:fuzzy_osf_graph_algebra}{\mbox{}}%
  We focus on conditions 3 and 4 of \cref{def:osf_algebra}.

  Let $\si[0],\si[1]\in \S$ and let $g=\ograph$ be an \osf graph.
  Then $\sgg[0](g)\land \fisop(\si[0], \si[1]) = \fisop(\lambda_N(\X),
  \si[0]) \land \fisop(\si[0], \si[1]) \leq \fisop(\lambda_N(\X), \si[1]) =
  \sgg[1](g)$ by transitivity of $\fisop$.

  Now suppose $(\sgg[0]\fcap \sgg[1])(g) >0$, so that
  $\sgg[i](g) = \fisop(\lambda_N(\X), \si[i]) > 0$ for $i\in\{ 0,1 \}$.
  Because $\ftax$ is a fuzzy lattice then %
  $\fisop(\lambda_N(\X), \si[0]\fmeet\si[1])>0$, so that
  $(\si[0]\fmeet\si[1])^\gg(g)> 0$.
\end{proof}
\fi

\ifallproofs
\thmsatisfiability*
\begin{proof}[Proof of \cref{thm:satisfiability}]%
  \hypertarget{proof:satisfiability}{\mbox{}}%
  If $\phi$ contains a constraint of the form $\Y:\s$, then
  $\canon,\alpha\models_1\Y:\s\Iff\s^{\canon}(\alpha(\Y)) = 1$, which holds
  because $\alpha(\Y) = G(\phi(\Y))$ is an \osf graph with root labeled by
  $\s$ by construction, so $\s^{\canon}(\alpha(\Y)) = \fisop(\s, \s) = 1$.

  If $\phi$ contains a constraint of the form $\Y.\f \doteq \Z$, then
  $\canon,\alpha\models_1\Y.\f\doteq\Z
  \Iff
  \f^{\canon}(\alpha(\Y)) = \alpha(\Z)
  \Iff
  \f^{\canon}(G(\phi(\Y))) = G(\phi(\Z))$, which is true by construction of
  $G(\phi(\Y))$
  and
  $G(\phi(\Z))$.
\end{proof}
\fi

\section{Proofs for \cref{sec:homomorphisms}: \nameref{sec:homomorphisms}}%
\label{proofs:hom}

\newcommand{\homsfirst}{%
  Let $\f\in\F$ and $d\in\dom$. Then
    $\gamma'(\gamma(\f^\ii(d))) = \gamma'(\f^\jj(\gamma(d))) =
    \f^\kk(\gamma'(\gamma(d)))$.
  Now let $\s\in\S$ and $d\in\dom$. Since
    $\s^\ii(d)\land\beta\leq\s^\jj(\gamma(d))$
  and
    $\s^\jj(\gamma(d))\land\beta'\leq\s^\kk(\gamma'(\gamma(d)))$
  then
  $\s^\ii(d)\land\beta\land\beta'\leq\s^\jj(\gamma(d))\land\beta'\leq\s^\kk(\gamma'(\gamma(d)))$.
}
\newcommand{\homssecond}{%
  Since $\gamma$ is a $\beta$-morphism, then $\forall d'\in\dom$:
  $\s^\ii(d')\land\beta\leq\s^\jj(\gamma(d'))$.
  Since $\beta'\leq\beta$, then, $\forall d'\in\dom$:
  $\s^\ii(d')\land\beta'\leq
  \s^\ii(d')\land\beta\leq\s^\jj(\gamma(d'))$.%
}
\newcommand{\homsclaimfirstproof}{%
  \begin{subproof}[Proof of \cref{claim:b}]
    Let $\beta_s\defeq\sup(B)$.
    A few facts used in the rest of the proof:
    \begin{enumerate}
      \item For all $\beta\in B$ and $\beta'\in[0,1]$: if $\beta'\leq \beta$,
        then $\beta'\in B$.
      \item For all $\beta'\in[0,1]$: if $\beta'\notin B$, then $\beta'$ is
        an upper bound of $B$, and thus $\beta_s\leq \beta'$.
      Indeed, suppose $\beta'\notin B$, so let $x_0\in X$ be such that
      $\min(f(x_0), \beta') > g(x_0)$. Let $\beta\in B$ be arbitrary, and
      we want to show that $\beta \leq \beta'$. Suppose that $\beta >
      \beta'$ towards a contradiction. Thus $\min(f(x_0), \beta')\leq
      \min(f(x_0), \beta) \leq g(x_0)$, which is the desired contradiction.
      \item For all $\beta'\in[0,1]$, if $\beta'<\beta_s$, then $\beta'\in B$.
    \end{enumerate}
    Suppose that $\forall x\in X$: $f(x) \leq g(x)$. Then
    $\min(f(x), \beta) \leq g(x)$
    is satisfied for all $\beta\in[0,1]$
    and all $x\in X$, so that $B = [0, 1]$ and $\sup(B) =
    1\in B$.

    Otherwise, let $C \defeq \{ x\in X \mid f(x) > g(x) \}\neq \emptyset$.
    Let $g(C) \defeq \{ g(x) \mid x\in C \}\subseteq[0,1]$. Let $\beta_g
    \defeq \inf(g(C))$. Then
      $\min(f(x), \beta_g)\leq g(x)$ holds
      for all $x\in X$, since:
    \begin{itemize}
      \item This is true for all $x$ such that $f(x)\leq g(x)$.
      \item If $x$ is such that $f(x) > g(x)$, then $x\in C$ and $g(x)\in
        g(C)$, thus $\beta_g\leq g(x)$.
    \end{itemize}
    This means that $\beta_g\in B$.
    Now we show that $\beta_g = \beta_s$:
    \begin{itemize}
      \item Clearly $\beta_g\leq\beta_s$, since $\beta_g\in B$ and $\beta_s =
        \sup(B)$.
      \item Suppose towards a contradiction that $\beta_g<\beta_s$. Then
        there is an $\epsilon>0$ such that $\beta_g+\epsilon<\beta_s$, so that
        $\beta_g+\epsilon\in B$.
        Then there exists an $x_0\in C$ such that $g(x_0)<\beta_g+\epsilon$,
        as otherwise $\beta_g+\epsilon\leq g(x)$ for all $x\in C$, which
        contradicts the fact that $\beta_g$ is the $\inf$ of $g(C)$.
        Then we have that
          $\beta_g\leq g(x_0) < \beta_g+\epsilon<\beta_s$.
        Since $\beta_g+\epsilon\in B$, then
          $\min(f(x_0), \beta_g+\epsilon)\leq g(x_0)$,
        but this is absurd, since $f(x_0) > g(x_0)$ (recall that $x_0\in C$)
        and $\beta_g+\epsilon > g(x_0)$.\qedhere
    \end{itemize}
  \end{subproof}
}
\newcommand{\homsclaimsecondproof}{%
  \begin{subproof}[Proof of \cref{claim:c}]
    Let $\beta_s\defeq\sup(B)$.
    A few facts used in the rest of the proof:
    \begin{enumerate}
      \item For all $\beta\in B$ and $\beta'\in[0,1]$: if $\beta'\leq \beta$,
        then $\beta'\in B$.
      \item For all $\beta'\in[0,1]$: if $\beta'\notin B$, then $\beta'$ is
        an upper bound of $B$, and thus $\beta_s\leq \beta'$.
      Indeed, suppose $\beta'\notin B$, so let $x_0\in X$ and $1\leq
      i_0\leq n$ be such that
      $\min(f_{i_0}(x_0), \beta') > g_{i_0}(x_0)$. Let $\beta\in B$ be arbitrary, and
      we want to show that $\beta \leq \beta'$. Suppose that $\beta >
      \beta'$ towards a contradiction. Thus $\min(f_{i_0}(x_0), \beta')\leq
      \min(f_{i_0}(x_0), \beta) \leq g_{i_0}(x_0)$, which is the desired contradiction.

      \item For all $\beta'\in[0,1]$, if $\beta'<\beta_s$, then $\beta'\in B$.
    \end{enumerate}
    For each $1\leq i \leq n$ let
    $B_i \defeq
      \{ \beta\in[0,1] \mid \forall x\in X: \min(f_i(x), \beta) \leq g_i(x) \}$
    and $\beta_i = \sup(B_i)$. By \cref{claim:b} it holds that $\beta_i\in B_i$
    for each $i$. Also note that $B\subseteq B_i$ for all $1\leq i\leq n$.

    Let $\beta_m = \min(\{ \beta_1, \ldots, \beta_n \})$.
    Note that it holds that
    $\min(f_i(x), \beta_m) \leq g_i(x)$
    for all $1\leq i\leq n$ and all $x\in X$.
    Indeed, for each $1\leq i\leq n$, $\beta_m\leq\beta_i$, and so, for each
    $x\in X$, $\min(f_i(x), \beta_m)\leq\min(f_i(x), \beta_i)\leq g_i(x)$.
    Thus, $\beta_m\in B$, directly implying that $\beta_m\in B_i$ for all
    $1\leq i\leq n$.

    Note that $\beta_s\leq\beta_i$ for all $i$, for suppose otherwise: then
    there is an $i$ such that $\beta_i < \beta_s$, so that there is a
    $\epsilon>0$ such that $\beta_i +\epsilon< \beta_s$ and thus
    $\beta_i+\epsilon\in B$. But then $\beta_i+\epsilon\in B_i$, which
    contradicts the fact that $\beta_i$ is the sup of $B_i$. Thus
    $\beta_s\leq\beta_i$ for all $i$, which means that $\beta_s\leq\min(\{
    \beta_1, \ldots, \beta_n \})=\beta_m$.
    Since $\beta_m\in B$ and $\beta_s = \sup(B)$, then $\beta_m \leq \beta_s$. Thus
    $\beta_m=\beta_s\in B$.\qedhere
  \end{subproof}
}
\newcommand{\homsthird}{%
  Let $B \defeq \{ \beta\in[0,1]\mid \gamma:\ii\to\jj~\text{is
  a}~\beta\text{-morphism} \}$ and $\beta_s = \sup(B)$.
  Note that
  \[
    \begin{array}{lll}
      B & \defeq & \{ \beta\in[0,1]\mid \gamma:\ii\to\jj~\text{is
      a}~\beta\text{-morphism} \}\\
        & = & \{ \beta\in[0,1]\mid \forall \s\in \S,\forall d\in\dom:
        \s^\ii(d)\land \beta\leq\s^\jj(\gamma(d)) \}
    \end{array}
  \]
  and that $B\neq \emptyset$ by assumption. The fact that $\beta_s\in
  B$ and thus $\beta_s$ is the maximum $\beta'$ such that $\gamma$ is a
  $\beta'$-morphism follows from the following
  \ifclaimproofs
  claims.
  \else
  claim.
  \fi
  \ifclaimproofs\begin{claim}
    \label{claim:b}
    Let $f:X\to[0,1]$ and $g:X\to[0,1]$ be functions and
    \[
      B \defeq \{ \beta\in[0,1] \mid \forall x\in X: \min(f(x), \beta)
      \leq g(x) \}.
    \]
    Then $\sup(B) \in B$.
  \end{claim}
  \homsclaimfirstproof{}\fi
  \begin{claim}
    \label{claim:c}
      For each $1\leq i\leq n$, let $f_i: X\to [0,1]$ and $g_i:
      X\to[0,1]$ be functions. Let
      \[
        B \defeq \{ \beta\in[0,1] \mid \forall x\in X, \forall 1\leq i\leq n:
        \min(f_i(x), \beta) \leq g_i(x) \}.
      \]
      Then $\sup(B)\in B$.\ifclaimproofs\else\qedhere\fi
  \end{claim}
  \ifclaimproofs\homsclaimsecondproof{}\fi
}

\prophoms*
\begin{proof}[Proof of \cref{prop:homs}]%
  \hypertarget{proof:homs}{\mbox{}}%
  \ifallproofs
    \begin{enumerate}
      \item \homsfirst{}
      \item \homssecond{}
      \item \homsthird{}
    \end{enumerate}
  \else
  The first two points are easy to show.

  \homsthird{}
  \fi
\end{proof}

\ifallproofs
\prophomsel*
\begin{proof}[Proof of \cref{prop:homsel}]%
  \hypertarget{proof:homsel}{\mbox{}}%
  \begin{enumerate}
    \item Let $d'\in\ii[d]$, i.e., $d' = w^\ii(d)$ for some $w\in\F^*$.
      Then $\gamma(d') = \gamma(w^\ii(d)) =
      w^\jj(\gamma(d)) \in\jj[\gamma(d)]$.
    \item Let $\gamma'$ be a $\beta'$-morphism $\gamma':\ii[d]\to\jj$ such
      that $\gamma'(d) = \gamma(d)$. Let
      $d'\in\ii[d]$, so that $d'=w^\ii(d)$ for some $w\in\F^*$. Then
      $\gamma(d') = \gamma(w^\ii(d)) = w^\jj(\gamma(d))
      =w^\jj(\gamma'(d)) = \gamma'(w^\ii(d)) = \gamma'(d')$.
    \item
      By the previous point
      \[
        \begin{array}{lll}
          B & \defeq & \{ \beta' \mid \exists~\text{$\beta'$-morphism}~\gamma':\ii[d]\to\jj~\text{such that}~\gamma'(d)=\gamma(d) \}\\
            & = & \{ \beta' \mid \gamma:\ii[d]\to\jj~\text{is a $\beta'$-morphism}\}
        \end{array}
      \]
      By \cref{prop:homs} the maximum $\beta_m$ of $B$ exists, and $\gamma$
      is a $\beta_m$-morphism.
      \qedhere
  \end{enumerate}
\end{proof}
\fi

\ifallproofs
\thmextending*
\begin{proof}[Proof of \cref{thm:extending}]%
  \hypertarget{proof:extending}{\mbox{}}%
  Let $\alpha' \defeq \gamma\circ\alpha$.

  Suppose that $\iimod_{\beta_\ii} \X:\s$, i.e., $\s^\ii(\alpha(\X))\geq
  \beta_\ii$. By \cref{def:osf_algebra_homomorphism}, then
  $\s^\jj(\alpha'(\X)) = \s^\jj(\gamma(\alpha(\X))) \geq \s^\ii(\alpha(\X))
  \land \beta \geq \beta_\ii\land \beta$, hence
  $\jj,\alpha'\models_{\beta_\ii\land\beta}\X:\s$.

  Suppose $\iimod_{\beta_\ii}\X.\f \doteq \Y$.
  If $\f^\ii(\alpha(\X)) \neq \alpha(\Y)$ then $\beta_\ii = 0$ and the result
  immediately follows.
  If $\f^\ii(\alpha(\X)) = \alpha(\Y)$ then
  $\f^\jj(\alpha'(\X)) = \f^\jj(\gamma(\alpha(\X)))) =
  \gamma(\f^\ii(\alpha(\X))) =
  \gamma(\alpha(\Y)) = \alpha'(\Y)$ and thus
  $\jj,\alpha' \models_{\beta_\ii\land\beta}\X.\f \doteq \Y$.

  Suppose. $\iimod_{\beta_\ii}\X \doteq \Y$.
  If $\alpha(\X) \neq \alpha(\Y)$ then $\beta_\ii=0$ and the result
  immediately follows.
  If $\alpha(\X) = \alpha(\Y)$ then
  $\alpha'(\X) = \gamma(\alpha(\X))) =
  \gamma(\alpha(\Y)) = \alpha'(\Y)$ and thus
  $\jj, \alpha' \models_{\beta_\ii\land\beta}\X \doteq \Y$.
\end{proof}
\fi

\newcommand{\suppd}{\supp[d]^{\ii}_{\S}}

\thmfinality*
\begin{proof}[Proof of \cref{thm:weak_finality}]%
  \hypertarget{proof:finality}{\mbox{}}%
  For each element $d\in\dom$ we construct an \osf graph
  $\gamma(d)\in\dom[\gg]$.

  Let us start with a few preliminary definitions. Let $d\in\dom$:
  \begin{itemize}
    \item the set of sorts to which $d$ belongs
      to with degree greater than $0$ is denoted\footnotemark
      \[
        \suppd\defeq \{ \s\in\S \mid \s^\ii(d)>0 \};
      \]
    \item the most specific sort to which $d$ belongs to with degree
      greater than $0$ is denoted
        $\s_{d} \defeq \fbigmeet\suppd$.
  \end{itemize}
  We construct a labeled graph $g = (N, E, \lambda_N, \lambda_E)$ with
  $N\subseteq \V$ as follows:
  \begin{itemize}
    \item for each $d\in\dom$ we choose a variable $\X_d\in\V$ to denote
      a node in $g$;
    \item the label of $\X_d$ is defined as $\lambda_N(\X_d) \defeq
      \s_{d}$;
    \item for each $\f\in\F$, if $\f^\ii(d) = d'$, then $g$ contains the
      edge $(\X_d, \X_{d'})$ labeled $\f$.
  \end{itemize}
  \footnotetext{Recall that we assume that $\tax$ is finite, and thus
  $\suppd$ is a finite subset of $\S$.}
  For each $d\in\dom$, let $\beta_d = \min_{\s\in\suppd}\fisop(\s_d, \s)$.
  Note that $\beta_d > 0$ for any $d\in\dom$, and that
  \ifshortproofs
  because the set $\S$ of sorts is finite, then the set $\{ \beta_d\mid
  d\in\dom \} = \{ \min_{\s\in\suppd}\fisop(\s_d, \s) \mid d \in\dom \}$ is also finite.
  \else
  the sets $\S$, $\S\times\S$ and $\pow(\S\times\S)$ are finite. Thus:
  \begin{itemize}
    \item for each $d\in\dom$, the sets $\suppd\subseteq\S$ and
    $E_d\defeq\{ (\s_d, \s) \mid \s\in\suppd \}\subseteq \S\times\S$ are
    finite;
    \item for each $d\in\dom$, the set $F_d\defeq\{ \fisop(\s, \su)
      \mid (\s, \su)\in E_d \} = \{ \fisop(\s_d, \s)
      \mid \s\in\suppd \} \subseteq (0,1]$ is finite;
    \item the set $\{ E_d \mid d\in\dom \}\subseteq \pow(\S\times\S)$ is
      finite, and thus so is the set $\{ F_d \mid d\in \dom \}$;
    \item thus, the set $\{ \beta_d\mid d\in\dom \} =
      \{ \min_{\s\in\suppd}\fisop(\s_d, \s) \mid d \in\dom \} = \{ \min(F_d)
      \mid d\in\dom \}$ is finite.
  \end{itemize}
  \fi
  Let $\beta = \min \{ \beta_d \mid d\in\dom \}>0$ and define
  $\gamma(d)$ to be the maximally connected subgraph of $g$ rooted in
  $\X_d$. It is left to show that $\gamma: \ii\to\gg$ is a $\beta$-homomorphism.
  \begin{itemize}
    \item Let $\f\in\F$ and $d\in\dom$. If $\f^\ii(d) = d'$, then
      $\gamma(d)$ has an edge labeled $\f$ from
      $\X_d$ to $\X_{d'}$, which is the root of $\gamma(d')$. Hence
      $\gamma(\f^\ii(d))  =\f^\gg(\gamma(d))$.
    \item Let $\s\in\S$ and $d\in\dom$.
      If $\s^\ii(d) = 0$, then
      $\s^\gg(\gamma(d)) \geq \beta\land \s^\ii(d)$. Otherwise,
      $\s\in\suppd$ and thus
      $\s^\gg(\gamma(d)) =
      \fisop(\s_d, \s)\geq\min_{\su\in\suppd}\fisop(\s_d,
      \su)=\beta_d\geq\beta\geq \beta\land \s^\ii(d)$.\qedhere
  \end{itemize}
\end{proof}

\ifallproofs
\corcanonicity*
\begin{proof}[Proof of \cref{cor:canonicity}]%
  \hypertarget{proof:canonicity}{\mbox{}}%
  Suppose $\phi$ is satisfiable in some \osf interpretation $\ii$, so let
  $\alpha:\V\to\dom$ and $\beta_\ii>0$ be such that
  $\ii,\alpha\models_{\beta_\ii}\phi$. By
  \cref{thm:weak_finality} let $\gamma:\ii\to\gg$ be a $\beta$-morphism
  for some $\beta>0$. By \cref{thm:extending}
  then $\gg,\gamma\circ\alpha\models_{\beta_\ii\land\beta}\phi$,
  i.e, $\phi$ is satisfiable in $\gg$. The other direction is obvious.
\end{proof}
\fi

\thmextracting*
\begin{proof}[Proof of \cref{thm:extracting}]
  \hypertarget{proof:extracting}{\mbox{}}%
  Let us abbreviate $G(\phi(\X))$ as $g_{\X}$.
  The $\beta$-morphism $\gamma:\canon\to\ii$ is defined as follows. Let
  $g\in\canon$, so that $g = w^\gg(g_{\X})$ for some $w\in \F^*$ and
  $\X\in\tags(\phi)$: we define $\gamma(g) = \gamma(w^\gg(g_{\X})) \defeq
  w^\ii(\alpha(\X))$.
  Note that $\gamma(g_{\X}) = \alpha(\X)$ for each $\X\in\tags(\phi)$.
  First of all, we need to show that $\gamma$ is well-defined. Let
  $g\in\canon$ and note that there are two cases to consider.
  \begin{itemize}
    \item Suppose $g = g_{\X}$ for some $\X\in\osftags(\phi)$ and let $w,
      w'\in\F^*$ and $\Y, \Z\in\osftags(\phi)$ be such that $g =
      w^\gg(g_{\Y}) =
      w'^\gg(g_{\Z})$. We want to prove that $w^\ii(\alpha(\Y)) =
    w'^\ii(\alpha(\Z))$. By construction of $g_{\X}$
    it must be the case that $\phi$ contains constraints of
    shape
    \[
      \begin{array}{lllll}
        \Y.\fj[0] \doteq \Yj[1],  & \Yj[1].\fj[1] \doteq \Yj[2],  & \ldots, & \Yj[n].\fj[n] \doteq \X & \text{and} \\
        \Z.\fjp[0] \doteq \Zj[1], & \Zj[1].\fjp[1] \doteq \Zj[2], & \ldots, & \Zj[m].\fjp[m] \doteq \X
      \end{array}
    \]
    such that $w = \fj[0]\cdot\fj[1]\cdots\fj[n]$ and
    $w' = \fjp[0]\cdot\fjp[1]\cdots\fjp[m]$. Because
      $\ii,\alpha\models_\beta \phi$ and $\beta>0$, then indeed $w^\ii(\alpha(\Y)) =
    w'^\ii(\alpha(\Z))$ as desired.
    \item Suppose $g = G(\Z:\tops)$ for some variable
      $\Z\notin\osftags(\phi)$. Then $g = w'^\gg(g_{\X})$ for some
      $\X\in\osftags(\phi)$ and some $w'\in\F^*$ such that that $g_{\X}$
      and $w'$
      uniquely determine $\Z$ (see \cref{def:osf_graph_algebra}). This
      case now reduces to the previous one.
  \end{itemize}
  We now prove that the function $\gamma$ is indeed a $\beta$-morphism.
  \begin{itemize}
    \item Let $\f\in\F$ and $g\in\canon$, so that $g$ has shape
      $g=w^\gg(g_{\X})$ as before. Let $w' = w\cdot \f\in\F^*$. Then
      $\gamma(\f^{\canon}(g)) = \gamma(\f^{\canon}(w^\gg(g_{\X}))) =
      \gamma(\f^\gg(w^\gg(g_{\X}))) =
      \gamma(w'^\gg(g_{\X})) =
      w'^\ii(\alpha(\X)) =
      \f^\ii(w^\ii(\alpha(\X))) = \f^\ii(\gamma(g))$.

  \item Now let $\s\in\S$ and we want to show that $\s^\ii(\gamma(g)) \geq
    \beta \land \s^\gg(g)$ for all $g\in\canon$. Let
    $g\in\canon$ be arbitrary. There are two cases to consider.
    \begin{itemize}
      \item Suppose $g = G(\Z:\tops)$ for some variable
        $\Z\notin\osftags(\phi)$. If $\s^\gg(g)=0$ then the result holds.
        Otherwise, if $\s^\gg(g)>0$, then $g$ is labeled by a sort $\su$ such
        that $\fisop(\su,\s)>0$, but since $g$ is labeled by $\su=\tops$, then
        $\s =\tops$, so that $\s^\ii(\gamma(g))=
        \tops^\ii(\gamma(g)) = 1$ and the result holds.
      \item Suppose $g = g_{\X}$ for some $\X\in\osftags(\phi)$ and
        let $\su$ be the label of the root of $g$. By construction of
        $g_{\X}$ then
        $\phi$ must contain a constraint of the form $\X:\su$,
        and by the assumption that $\iimodb\phi$ we have that
        $\su^\ii(\alpha(\X)) \geq\beta$.
        By \cref{def:osf_algebra} then we have that
        $\s^\ii(\gamma(g_{\X}))=\s^\ii(\alpha(\X))
        \geq
        \su^\ii(\alpha(\X))\land\fisa(\su, \s) \geq \beta\land\fisa(\su,\s) =
        \beta\land\s^\gg(g)$. \qedhere
    \end{itemize}
  \end{itemize}
\end{proof}

\thminterpretability*
\begin{proof}[Proof of \cref{thm:interpretability}]%
  \hypertarget{proof:interpretability}{\mbox{}}%
  Let $\X$ be the root variable of $\psi$ and let $d\in\dom$.

  Suppose that $\denot[\psi](d)\geq\beta$. By
  \cref{def:osf_term_denotation,rem:osf_term_denotation} then there is some
  $\alpha$ such that $\denota[\psi](d)\geq \beta$ and $d=\alpha(\X)$.
  By \cref{prop:equivalence_terms_constraints} then $\iimodb\phi$, so that
  by \cref{thm:extracting} there is a $\beta$-homomorphism
  $\gamma:\canon\to\ii$ such that $d = \alpha(\X) = \gamma(G(\phi(\X))) =
  \gamma(G(\psi))$.%

  Now suppose $\gamma:\canon\to\ii$ is a $\beta$-homomorphism such that
  $d=\gamma(G(\psi))$.
  By \cref{thm:satisfiability} we know that $\canon,\alpha\models_1 \phi$
  where $\alpha:\V\to\dom[\canon]$ is such that $\alpha(\Y) = G(\phi(\Y))$
  for all $\Y\in\tags(\phi)$, and thus $d = \gamma(G(\psi)) =
  \gamma(G(\phi(\X))) = \gamma(\alpha(\X))$.
  Then by \cref{thm:extending} $\ii, \gamma\circ\alpha \models_\beta \phi$,
  and thus
  by \cref{prop:equivalence_terms_constraints} it holds that
  $\denota[\psi][\gamma\circ\alpha](\gamma(\alpha(\X))) \geq\beta$, i.e.,
  $\denota[\psi][\gamma\circ\alpha](d)\geq \beta$, so that
  $\denot[\psi](d)\geq \beta$.

  Finally, note that by what we just proved
  (and the fact that, for any
  $S\subseteq [0,1]$, $\sup(S) = \sup(S\setminus\{ 0 \})$):
  \[
    \begin{array}[b]{lll}
      \denot[\psi](d)
        &=&
        \sup(\{ \beta\in[0,1]
        \mid \denot[\psi](d)\geq\beta \})\\
        &=&
        \sup(\{ \beta\in(0,1]
        \mid \denot[\psi](d)\geq\beta \})\\
        &=&
        \sup(\{ \beta\in(0,1]
        \mid\exists~\text{$\beta$-morphism}~
        \gamma:\canon\to\ii~\text{such that}~d=\gamma(G(\psi)) \})\\
        &=&
        \sup(\{ \beta\in[0,1]
        \mid\exists~\text{$\beta$-morphism}~
        \gamma:\canon\to\ii~\text{such that}~d=\gamma(G(\psi)) \}).
    \end{array}\qedhere
  \]
\end{proof}

\section{Proofs for \cref{sec:subsumption}: \nameref{sec:subsumption}}%
\label{proofs:sub}

\proppreorder*
\begin{proof}[Proof of \cref{prop:preorder}]%
  \hypertarget{proof:proppreorder}{\mbox{}}%
  Let $\ii$ be a fuzzy \osf interpretation.

  For all $d\in\ii$ the identity function on $\ii[d]$ is clearly a
  $1$-morphism, so that $\fapproximates[\ii](d, d) = 1$.

  Now suppose $\fapproximates[\ii](d_0, d_1) = \beta_0$ and
  $\fapproximates[\ii](d_1, d_2) = \beta_1$. We want to prove that
  $\fapproximates[\ii](d_0, d_2) \geq \beta_0 \land \beta_1$. Let $\beta_0
  > 0$ and $\beta_1>0$ (otherwise, the desired result follows immediately)
  and let $\gamma_0:\ii[d_0]\to\ii[d_1]$ and $\gamma_1:\ii[d_1]\to\ii[d_2]$
  be, respectively, a $\beta_0$-morphism and a $\beta_1$-morphism. Note
  that these exist because of \cref{prop:homsel}. Then by \cref{prop:homs}
  $\gamma_1\circ\gamma_0$ is a $\beta_0\land\beta_1$ morphism, and
  $\fapproximates[\ii](d_0,d_2)\geq \beta_0\land\beta_1$ follows by
  \cref{def:endomorphic_approximation}.
\end{proof}

\newcommand{\lemmasymmorphisms}{%
  Let $g_0$ and $g_1$ be two \osf graphs. If
  $\gamma_0: \gg[g_0]\to\gg[g_1]$ is a $\beta_0$-morphism ($\beta_0>0$) and
  $\gamma_1: \gg[g_1]\to\gg[g_0]$ is a $\beta_1$-morphism ($\beta_1>0$)
  such that
  $\gamma_0(g_0) = g_1$
  and
  $\gamma_1(g_1) = g_0$, then:
  \begin{enumerate}
    \item for all $g\in\gg[g_0]$, $g$ and $\gamma_0(g)$ are labeled by the
      same sort;
    \item for all $g\in\gg[g_1]$, $g$ and $\gamma_1(g)$ are labeled by the
      same sort;
    \item $\gamma_0\circ\gamma_1 =  \id[{\dom[{\gg[g_1]}]}]$ and
      $\gamma_1\circ\gamma_0 =  \id[{\dom[{\gg[g_0]}]}]$, which implies
      that $\gamma_0$ and $\gamma_1$ are bijections;
    \item $\gamma_0$ and $\gamma_1$ are 1-morphisms.\ifallproofs\else\qedhere\fi
  \end{enumerate}%
}
\ifallproofs
\begin{lemma}[Morphisms between two graphs]%
  \label{prop:symmorphisms}
  \lemmasymmorphisms{}
\end{lemma}
\begin{proof}[Proof of \cref{prop:symmorphisms}]%
  According to the statement of the proposition, let
  \begin{itemize}
    \item $\gamma_0:\gg[g_0]\to\gg[g_1]$ be a
      $\beta_0$-morphism
      such that $\gamma_0(g_0) = g_1$, and
    \item $\gamma_1:\gg[g_1]\to\gg[g_0]$ be a $\beta_1$-morphism
      such that $\gamma_1(g_1) = g_0$
  \end{itemize}
  with $\beta_0>0$ and $\beta_1$>0.

  Let $g\in\gg[g_1]$, so that $g = w^\gg(g_1)$ for some $w\in \F^*$. Then
  \[
    \begin{array}{lllll}
      \gamma_0(\gamma_1(g)) & = & \gamma_0(\gamma_1(w^\gg(g_1)))
                            & = & \gamma_0(w^\gg(\gamma_1(g_1))) \\
                            & = & w^\gg(\gamma_0(\gamma_1(g_1)))
                            & = & w^\gg(\gamma_0(g_0)) = w^\gg(g_1) = g,
    \end{array}
  \]
  i.e., $\gamma_0\circ\gamma_1 = \id[{\dom[{\gg[g_1]}]}]$.

  Let $\s$ be the sort labeling $g$ and $\su$ be the sort labeling
  $g'\defeq\gamma_1(g)$. By \cref{def:osf_algebra_homomorphism} then
  \begin{align*}
    \s^\gg(g) \land \beta_1 \leq \s^\gg(g')~\text{and}~%
    \su^\gg(g') \land \beta_0 \leq \su^\gg(\gamma_0(g')) = \su^\gg(g).
  \end{align*}
  Note that $\s^\gg(g) = 1$ and $\su^\gg(g') = 1$, so that $\s^\gg(g') > 0$
  and $\su^\gg(g)> 0$. Thus
  \begin{align*}
    \fisop(\su, \s) = \s^\gg(g') > 0~\text{and}~%
    \fisop(\s, \su) = \su^\gg(g) > 0,
  \end{align*}
  so that by antisymmetry of $\fisop$ it follows that $\s = \su$, meaning
  that $g$ and $\gamma_1(g)$ are labeled by the same sort.

  Now let $\s$ be an arbitrary sort, and $g\in\gg[g_1]$ be arbitrary. By
  what we just showed, $g$ and $\gamma_1(g)$ are labeled by the same sort,
  so that $\s^\gg(g) = \s^\gg(\gamma_1(g))$, and thus $\s^\gg(g)\land 1\leq
  \s^\gg(\gamma_1(g))$, i.e., $\gamma_1$ is a 1-morphism.

  In a very similar way it can be shown that
  (i) $\gamma_0$ is a 1-morphism,
  (ii) for all $g\in\gg[g_0]$, $g$ and $\gamma_0(g)$ are labeled by the
  same sort, and (iii)
    $\gamma_1\circ\gamma_0 = \id[{\dom[{\gg[g_0]}]}]$.\qedhere
\end{proof}
\fi

\proppartial*
\begin{proof}[Proof of \cref{prop:partial}]%
  \hypertarget{proof:proppartial}{\mbox{}}%
  \ifallproofs
  By \cref{prop:symmorphisms} and \cref{prop:preorder}.
  \else
  By \cref{prop:preorder} and the following claim.
  \begin{claim}
    \lemmasymmorphisms{}
  \end{claim}
  \fi
\end{proof}

\ifallproofs
\propequivalence*
\begin{proof}[Proof of \cref{prop:equivalence}]
  \hypertarget{proof:equivalence}{\mbox{}}%
  Let $g_0 = G(\psi_0)$ and $g_1 = G(\psi_1)$. By assumption we have that
  $g_0\gequiv g_1$.
  Let $\ii$ be arbitrary and $d\in\dom[\ii]$. Then by
  \cref{prop:homs}, \cref{thm:interpretability} and \cref{def:osf_graph_equivalence}:
  \[
    \begin{array}[b]{lll}
      \denot[\psi_0](d) & = &
      \sup(\{
        \beta%
        \mid \exists~\text{$\beta$-morphism}~
        \gamma:\canon\to\ii~\text{such that}~d=\gamma(g_0)
      \})\\
                        & = & \sup(\{
        \beta%
        \mid \exists~\text{$\beta$-morphism}~
        \gamma:\canon\to\ii~\text{such that}~d=\gamma(g_1)
      \})\\
                        & = & \denot[\psi_1](d).
    \end{array}\qedhere
  \]
\end{proof}
\fi

\ifallproofs
\propequivalenceclause*
\begin{proof}[Proof of \cref{prop:equivalence_clause}]
  \hypertarget{proof:equivalence_clause}{\mbox{}}%
  Let $\ii,\alpha_0\models_\beta\phi_0$ and let $\X$ be the root of
  $\phi_0$ and $\Y$ be the root of $\phi_1$. By
  \cref{prop:equivalence_terms_constraints} then
  $\denot[\psi(\phi_0)](\alpha_0(\X))\geq\beta$, and thus by \cref{prop:equivalence}
  $\denot[\psi(\phi_1)](\alpha_0(\X))\geq\beta$.
  Let $\denot[\psi(\phi_1)](\alpha_0(\X))=\beta_1$. If $\beta_1 = 0$, then $\beta=0$ and
  the conclusion follows trivially. Otherwise,
  by \cref{rem:osf_term_denotation} then there is some $\alpha_1$ such that
  $\denota[\psi(\phi_1)][\alpha_1](\alpha_0(\X))=\beta_1\geq\beta$ and thus
  $\alpha_0(\X) = \alpha_1(\Y)$, and thus by
  \cref{prop:equivalence_terms_constraints} we obtain
  $\ii,\alpha_1\models_\beta\phi_1$. The other direction is analogous.
\end{proof}
\fi

\begin{restatable}%
  [{\protect\hyperlink{proof:transparency}{Equivalence of fuzzy \osf orderings}}]%
  {lemma}{thmtransparency}
\label{thm:transparency}
  If
  the normal \osf terms $\psi$ and $\psi'$
  (with roots $\Y$ and $\X$, respectively, and no common variables),
  the \osf graphs $g$ and $g'$, and
  the rooted solved \osf clauses $\phi_{\Y}$ and $\phi'_{\X}$
  respectively correspond to one another though the syntactic mappings,
  then the following are equivalent, for all $\beta\in(0,1]$.
  \begin{enumerate}
    \item There is a $\beta$-morphism $\gamma:\gg[g]\to\gg[g']$ such that
      $\gamma(g) = g'$.
    \item For all \osf interpretations $\ii$ and $d\in\dom$:
      $\denot[\psi'][\ii](d)\land\beta \leq \denot[\psi][\ii](d)$.
    \item For all \osf interpretations $\ii$ and assignments
      $\alpha$ such that $\ii,\alpha\models_{\beta'}\phi'$, there
      is an assignment $\alpha'$ such that $\alpha'(\X) = \alpha(\X)$ and
      $\ii,\alpha'\models_{\beta'\land\beta}\phi[\X/\Y]$.
    \item For all $g\in\dom[\gg]$: $\denot[\psi'][\gg](g)\land\beta \leq \denot[\psi][\gg](g)$.
  \end{enumerate}
\end{restatable}
\begin{proof}[Proof of \cref{thm:transparency}]%
  \hypertarget{proof:transparency}{\mbox{}}%
  \paragraph{\textbf{(1) implies (2)}}%

  Suppose that there is a $\beta$-homomorphism
  $\gamma:\gg[g]\to\gg[g']$ such
  that $g'= \gamma(g)$. We want to show that for all $\ii$ and $d\in\dom$:
  \[
    \denot[\psi'](d)\land\beta\leq\denot[\psi](d).
  \]
  Let $\ii$ be an arbitrary \osf interpretation, let
  $d\in\dom$ and suppose $\denot[\psi'](d) = \beta' > 0$ (otherwise the desired
  result immediately follows). Then by \cref{thm:interpretability} there is
  a $\beta'$-morphism $\gamma':\canon[\phi']\to\ii$ such that $d =
  \gamma'(G(\psi')) = \gamma'(g')$.
  Note that $\gg[g] = \canon$ and $\gg[g'] = \canon[\phi']$, thus we can
  also write $\gamma:\canon\to\canon[\phi']$.
  Then by \cref{prop:homs} $\gamma'\circ\gamma:\canon\to\ii$ is a
  $\beta\land\beta'$-morphism
  and $d = \gamma'(\gamma(g)) = (\gamma'\circ\gamma)(g) =
  (\gamma'\circ\gamma)(G(\psi))$. By \cref{thm:interpretability} then
  $\denot[\psi](d)\geq\beta\land\beta' = \beta\land\denot[\psi'](d)$.

  \paragraph{\textbf{(2) implies (1)}}

  Suppose that, for any $\ii$ and $d\in\dom$:
  $\denot[\psi'](d)\land\beta\leq\denot[\psi](d)$.
  Then in particular for all $g''\in\dom[\gg]$:
  $\denot[\psi'][\gg](g'')\land\beta \leq \denot[\psi][\gg](g'')$.

  Since $g' = G(\psi')$, then $\denot[\psi'][\gg](g') = 1$, so that
  $\denot[\psi][\gg](g') \geq \beta > 0$. Thus by
  \cref{thm:interpretability} there is a $\beta$-morphism $\gamma:
  \canon\to\gg$ such that $g' = \gamma(G(\psi)) = \gamma(g)$. Note that
  $\gg[g] = \canon$, since $g = G(\phi)$. Then by \cref{prop:homsel}
  $\gamma:\gg[g]\to\gg[\gamma(g)]$, i.e., $\gamma:\gg[g]\to\gg[g']$ as
  desired.

  The following claim will be needed in the next two directions.
  \begin{claim}
  \label{claim:substitution}
    Let $\phi$ be an \osf clause, $\beta\in[0,1]$, $\Y\in\tags(\phi)$ and
    $\X\notin\tags(\phi)$: if $\ii, \alpha\models_{\beta'}\phi$, then
    $\ii,\alpha'\models_{\beta'} \phi[\X/\Y]$, where
    \[
      \alpha'(\Z) =
      \begin{cases}
        \alpha(\Y) & \text{if}~ \Z=\X \\
        \alpha(\Z) & \text{otherwise}.
      \end{cases}
    \]
  \end{claim}
  \paragraph{\textbf{(2) implies (3)}}
  Suppose that $\ii,\alpha\models_{\beta'} \phi'$. Then
  $\denot[\psi'](\alpha(\X))\geq\beta'$ by
  \cref{prop:equivalence_terms_constraints}, so that by assumption (2)
  $\denot[\psi](\alpha(\X))\geq\beta\land\beta'$. %
  If $\beta' = 0$ the desired result follows immediately, so suppose
  otherwise.
  By \cref{rem:osf_term_denotation} %
  and \cref{prop:equivalence_terms_constraints}
  then there is some $\alpha'$ such that $\ii,
  \alpha'\models_{\beta'\land\beta} \phi$ and
  $\alpha(\X) = \alpha'(\Y)$ %
  (recall that $\Y$ is the root tag of
  $\phi$ and $\psi$). But then $\alpha''$ defined by letting
  \[
    \alpha''(\Z) =
    \begin{cases}
      \alpha'(\Y) & \text{if}~ \Z=\X \\
      \alpha'(\Z) & \text{otherwise}
    \end{cases}
  \]
  is such that $\ii, \alpha''\models_{\beta'\land\beta} \phi[\X/\Y]$ by
  \cref{claim:substitution}, and $\alpha''(\X) = \alpha'(\Y) = \alpha(\X)$
  as desired.

  \paragraph{\textbf{(3) implies (2)}}

  Let $\ii$ and $d\in\dom[\ii]$ be arbitrary and we want to show that %
  $\denot[\psi'](d)\land\beta\leq\denot[\psi](d)$.
  Suppose that $\denot[\psi'](d) = \beta' > 0$ (otherwise the result
  follows immediately). By \cref{rem:osf_term_denotation} and
  \cref{prop:equivalence_terms_constraints} then
  there is some $\alpha$ such that
  $d=\alpha(\X)$ and $\ii,\alpha\models_{\beta'}\phi'$.

  By assumption (3) then there is some $\alpha'$ such that
  $\alpha'(\X) = \alpha(\X)$ and
  $\ii,\alpha'\models_{\beta\land\beta'} \phi[\X/\Y]$.
  Let $\alpha''$ be defined by letting
  \[
    \alpha''(\Z) =
    \begin{cases}
      \alpha'(\X) & \text{if}~\Z=\Y\\
      \alpha'(\Z) & \text{otherwise}.
    \end{cases}
  \]
  Then $\ii,\alpha''\models_{\beta\land\beta'} \phi[\X/\Y][\Y/\X]$ by
  \cref{claim:substitution}, i.e.,
  $\ii,\alpha''\models_{\beta\land\beta'} \phi$. Since $\alpha''(\Y) = \alpha'(\X) = \alpha(\X)
  = d$, then by \cref{prop:equivalence_terms_constraints}
  $\denot[\psi](d)\geq\beta\land\beta' = \beta\land\denot[\psi'](d)$ as
  desired.

  \paragraph{\textbf{(2) implies (4)}} Obvious.

  \paragraph{\textbf{(4) implies (1)}} Similar to \textit{\textbf{(2) implies (1)}}.
\end{proof}

\thmtransparencybis*
\begin{proof}[Proof of \cref{thm:transparencybis}]%
  \hypertarget{proof:transparencybis}{\mbox{}}%
  Note that, by \cref{thm:transparency}
  (and the fact that, for any
  $S\subseteq [0,1]$, $\sup(S) = \sup(S\setminus\{ 0 \})$):
  \[
    \begin{array}[b]{ll}
      & \sup(\{ \beta\in[0,1] \mid \gamma(g) = g'~\text{for some $\beta$-homomorphism}~\gamma:\gg[g]\to\gg[g'] \}) \\
      \addlinespace[1mm]
      = & \sup\left(\left\{
      \begin{array}{l|l}
        \multirow{2}{*}{$\beta\in[0,1]$}
        & \forall \ii, \forall d\in\dom:
        (\ii,\alpha\models_{\beta'}\phi') \To (\exists \alpha'. \alpha'(\X)
        = \alpha(\X)\\
                      & \text{and}~\ii,\alpha'\models_{\beta'\land\beta}\phi[\X/\Y]
      \end{array}
  \right\}\right) \\
      \addlinespace[1mm]
      = & \sup(\{ \beta\in[0,1] \mid \forall \ii, \forall d\in\dom: \denot[\psi'][\ii](d)\land\beta \leq \denot[\psi][\ii](d) \}).
    \end{array}\qedhere
  \]
\end{proof}

\begin{restatable}%
  [{\protect\hyperlink{proof:lemmasynsem}{Semantic and syntactic subsumption}}]
  {lemma}{lemmasynsem}
\label{lemma:synsem}
  Let $\psi_0$ and $\psi_1$ be two normal $\osf$ terms. Then,
  for all $\beta\in(0,1]$:
  $\fisop(\psi_0, \psi_1)\geq\beta$ if and only if there are two (normal)
  \osf terms $\psi_0'$ and $\psi_1'$ such that
    $\psi_0 \gequiv \psi_0'$,
    $\psi_1 \gequiv \psi_1'$, and
    $\fsynisop(\psi_0', \psi_1')\geq\beta$.
\end{restatable}
\begin{proof}[Proof of \cref{lemma:synsem}]%
  \hypertarget{proof:lemmasynsem}{\mbox{}}%
  \paragraph{\textbf{$(\bm{\oT})$}} Let $\psi_0'$ and
  $\psi_1'$ be as in the statement of the lemma, with
  $h:\tags(\psi_1')\to\tags(\psi_0')$ witnessing
  $\fsynisop(\psi_0', \psi_1')\geq\beta$.
  By \cref{prop:equivalence} it holds that
  $\denot[\psi_0] = \denot[\psi_0']$ and
  $\denot[\psi_1] = \denot[\psi_1']$ for all $\ii$.
  We show that $\fisop(\psi_0, \psi_1)\geq\beta$ by showing that
      $\forall \ii, \forall d\in\dom[\ii]:
      \denot[\psi_0](d)\land\beta\leq\denot[\psi_1](d)$.

  Let $\ii$ be an arbitrary fuzzy interpretation, let $d\in\dom[\ii]$ and
  suppose $\denot[\psi_0](d) = \denot[\psi_0'](d) = \beta_0$.
  Assume $\beta_0 > 0$, as otherwise the result follows immediately.
  Then by \cref{prop:equivalence_terms_constraints,rem:osf_term_denotation} there is some
  $\alpha_0:\V\to\dom$ such that $\ii, \alpha_0\models_{\beta_0} \phi_0'$, where
  $\phi_0' = \phi(\psi_0')$, $d = \alpha_0(\X[X_0'])$, and $\X[X_0']
  = \rtag(\psi_0')$. Define $\alpha_1:\V\to\dom$ as follows, for
  any $\X\in\V$:
  \[
    \alpha_1(\X) =
    \begin{cases}
      \alpha_0(h(\X)) & \text{if}~\X\in\tags(\psi_1'),\\
      \alpha_0(\X) & \text{otherwise}.
    \end{cases}
  \]
  Then $\ii,\alpha_1\models_{\beta\land\beta_0}\phi_1'$, where $\phi_1' = \phi(\psi_1')$:
  \begin{itemize}
    \item Suppose $\phi_1'$ contains a constraint of the form $\Xj[1]:\si[1]$.
      Since $\phi_1'$ is a rooted solved clause constructed from $\psi_1'$,
      then it must be the case that $\si[1] = \sort_{\psi_1'}(\Xj[1])$. Let
      $\si[0] = \sort_{\psi_0'}(h(\Xj[1]))$ and note that by assumption
      that $\fsynisop(\psi_0', \psi_1')\geq\beta$ we have that
      (a) $\fisop(\si[0],\si[1])\geq\beta$.
      By construction $\phi_0'$ contains a
      constraint of the form $h(\Xj[1]):\si[0]$, and since $\ii,
      \alpha_0\models_{\beta_0}\phi_0'$, then
      (b) $\sii[0](\alpha_0(h(\Xj[1])))\geq\beta_0$.
      Thus
      \[
        \begin{array}{llll}
          \sii[1](\alpha_1(\Xj[1])) & =    & \sii[1](\alpha_0(h(\Xj[1])))                             \\
                                    & \geq & \sii[0](\alpha_0(h(\Xj[1]))) \land \fisop(\si[0],\si[1])  & (\text{\cref{def:osf_algebra}}) \\
                                    & \geq & \beta_0\land\beta. & (\text{by (a) and (b)})
        \end{array}
      \]
    \item Suppose $\phi_1'$ contains a constraint of the form $\X.\f \doteq
      \Y$. Then $\X\fto_{\psi'_1}\Y$ and thus by assumption
      that $\fsynisop(\psi_0', \psi_1')\geq\beta$ we have that
      $h(\X)\fto_{\psi_0'}h(\Y)$, so that $\phi_0'$ must contain a
    constraint of the form $h(\X).\f \doteq h(\Y)$.
    Since $\ii, \alpha_0\models_{\beta_0} \phi_0'$ and $\beta_0>0$, then
    $\f^\ii(\alpha_0(h(\X))) = \alpha_0(h(\Y))$, and thus
    $\f^\ii(\alpha_1(\X)) = \alpha_1(\Y)$, so that
    $\ii,\alpha_1\models_{\beta\land\beta_0} \X.\f \doteq \Y$.
  \end{itemize}
  Let $\X[X_1'] = \rtag(\psi_1')$. By \cref{def:syn_osf_term_subsumption}
  then $h(\X[X_1']) = \X[X_0']$, so that $\alpha_1(\X[X_1']) =
  \alpha_0(h(\X[X_1'])) =\alpha_0(\X[X_0']) = d$. Since
  $\ii,\alpha_1\models_{\beta\land\beta_0} \phi_1'$, then by
  \cref{prop:equivalence_terms_constraints}
  $\denot[\psi_1'](d) =
  \denot[\psi_1](d) \geq \beta_0\land\beta = \denot[\psi_0](d) \land \beta$
  as desired.

  \paragraph{\textbf{$(\bm{\To})$ (Sketch)}}
  Let $\phi_0 = \phi(\psi_0)$ and $\phi_1 =
  \phi(\psi_1)$ and $g_0 = G(\psi_0)$ and $g_1 = G(\psi_1)$.
  In the following, we abbreviate $G(\phi_0(\X))$ as
  $g_0^{\X}$ and $G(\phi_1(\X))$ as $g_1^{\X}$ (in other words, $g_i^{\X}$ is
  the subgraph of $g_i$ rooted at the node corresponding to the variable
  $\X\in\tags(\psi_i)$).

  Since $\fisa(\psi_0,\psi_1)\geq\beta$, then $\fapproximates(g_1,
  g_0)\geq\beta$, so there is a $\beta$-morphism
  $\gamma:\gg[g_1]\to\gg[g_0]$ such that $g_0 = \gamma(g_1)$.
  The following steps almost provide a mapping
  $h:\tags(\psi_1)\to\tags(\psi_0)$ that witnesses $\fsynisop(\psi_0,
  \psi_1)\geq\beta$:
  \begin{enumerate}
    \item for $\Y\in\tags(\psi_1)$, consider
      $g_1^{\Y}\in\gg[g_1]$
      and
      $\gamma(g_1^{\Y})\in\gg[g_0]$;
    \item let $\X$ be the variable in $\psi_0$ such that
      $g_0^{\X} =
      \gamma(g_1^{\Y})$;
      and
    \item define $h(\Y) \defeq \X$.
  \end{enumerate}
  Unfortunately, nothing guarantees that such a variable $\X\in\tags(\psi_0)$
  actually exists, because of cases such as the following:
  \begin{center}
    \begin{tabular}{l}
      $\si[0]\fisa_\beta\si[1]$, $\psi_0 = \X[X_0]:\si[0]$, and
      $\psi_1 = \X[Y_0]:\si[1](\f\to \Y[Y_1]:\tops)$
    \end{tabular}
  \end{center}
  where $\gamma(G(\phi_1(\Yj[1]))) = \gamma(G(\Yj[1]:\tops))$ would be the
  trivial graph (with main sort $\tops$) obtained by applying $\f^\gg$ to
  $G(\psi_0)$ (recall \cref{def:osf_graph_algebra}).
  Clearly $\psi_0\fisa_\beta\psi_1$\footnotemark, but we cannot define an
  $h$ that satisfies \cref{def:syn_osf_term_subsumption}.
  \footnotetext{Note that, for any \osf term $\psi = \osfterm$,
    $\denot[\psi] = \denot[\xs(\fti, \ldots, \fti[n], \f\to\tops)]$, for
    any $\f\in\F$.}%
  This is why the statement of the lemma mentions a term $\psi_0'$ that
  is equivalent to $\psi_0$, where the required subterms of $\psi_0$
  corresponding to such trivial graphs would be made explicit (in the latter
  case $\psi_0'$ would be $\X[X_0]:\si[0](\f\to\X:\tops)$). The details
  concerning the general construction of such $\psi_0'$ %
  are left to the reader.
  In the following we simply assume that $\psi_0$ contains
  the variables needed to define $h$ as above.

  Now we show that the function $h$ witnesses the fact that
    $\fsynisa(\psi_0, \psi_1)\geq\beta$.
    \begin{itemize}
      \item Clearly $h(\rtag(\psi_1)) = \rtag(\psi_0)$, since $\gamma(g_1)
        = g_0$,
        and $h$ has been defined accordingly.
      \item Suppose that $\Yj[0]\fto_{\psi_1}\Yj[1]$.
        According to our definition of $h$:
        \begin{itemize}
          \item let $\Xj[0]$ be the variable in $\psi_0$ such that
            $g_0^{\Xj[0]} = \gamma(g_1^{\Yj[0]}))$,
            so that
            $h(\Yj[0]) = \Xj[0]$; and
          \item let $\Xj[1]$ be the variable in $\psi_0$ such that
            $g_0^{\Xj[1]} = \gamma(g_1^{\Yj[1]})$,
            so that
            $h(\Yj[1]) = \Xj[1]$.
        \end{itemize}
        Since $\Yj[0]\fto_{\psi_1}\Yj[1]$, then
        $g_1^{\Yj[1]} = \f^\gg(g_1^{\Yj[0]})$ and,
        since $\gamma$ is a $\beta$-morphism, then
        $\gamma(g_1^{\Yj[1]}) =
        \gamma(\f^\gg(g_1^{\Yj[0]})) =
        \f^\gg(\gamma(g_1^{\Yj[0]}))$,
        and thus
        $g_0^{\Xj[1]} = \f^\gg(g_0^{\Xj[0]})$.
        By construction of
        $g_0$ this means that
        $\Xj[0]\fto_{\psi_0}\Xj[1]$,
        i.e.,.
        $h(\Yj[0])\fto_{\psi_0}h(\Yj[1])$
        as desired.
      \item Now let $\Y\in\tags(\psi_1)$ be arbitrary.
        According to our definition of
        $h$, let $\X$ be the variable in $\psi_0$ such that
        $g_0^{\X} = \gamma(g_1^{\Y})$,
        so that
        $h(\Y) = \X$.
        Let
        $\si[0] = \sort_{\psi_0}(\X)$
        and
        $\si[1] = \sort_{\psi_1}(\Y)$.
        Since $\gamma$ is a $\beta$-morphism, then
        $\sgg[1](g_1^{\Y}) \land \beta \leq
        \sgg[1](\gamma(g_1^{\Y}))
        = \sgg[1](g_0^{\X})$.
        Note that
        $\sgg[1](g_1^{\Y}) = \fisop(\si[1], \si[1]) = 1$
        and
        $\sgg[1](g_0^{\X}) = \fisop(\si[0], \si[1])$,
        so the previous inequation simplifies to
        $\beta\leq\fisop(\si[0],\si[1])$.

    Since $\Y\in\tags(\psi_1)$ was arbitrary, we have shown that
    $\fisop(\sort_{\psi_0}(h(\Y)), \sort_{\psi_1}(\Y)) \geq\beta$ is
    true for all $\Y\in\tags(\psi_1)$,
    and thus
    $\min(\{
      \fisop(\sort_{\psi_0}(h(\Y)), \sort_{\psi_1}(\Y))
      \mid
      \Y\in\tags(\psi_1)
    \}) \geq\beta$,
    showing indeed that
    $\fsynisop(\psi_0, \psi_1)\geq\beta$.\qedhere
    \end{itemize}
\end{proof}

\newcommand{\bsem}{\beta_{\mathit{sem}}}
\newcommand{\bsyn}{\beta_{\mathit{syn}}}
\propsynsem*
\begin{proof}[Proof of \cref{prop:synsem}]%
  \hypertarget{proof:synsem}{\mbox{}}%
  This is a consequence of \cref{lemma:synsem} and the following fact.
  \begin{claim}
    \label{claim:syndegree}
    Let $\psi_0$ and $\psi_1$ be normal \osf terms. If
    \begin{itemize}
      \item  there exist $\psi_0'$ and $\psi_1'$ such that $\psi_0'\equiv
        \psi_0$ and $\psi_1'\equiv \psi_1$ and
        $\psi_0'\fsynisa_{\beta'}\psi_1'$ with $\beta'>0$, and
      \item  there exist $\psi_0''$ and $\psi_1''$ such that $\psi_0''\equiv
        \psi_0$ and $\psi_1''\equiv \psi_1$ and
        $\psi_0''\fsynisa_{\beta''}\psi_1''$ with $\beta''>0$,
    \end{itemize}
    then $\beta' = \beta''$.
  \end{claim}
  The reason why this holds is that, if $\psi_i$, $\psi_i'$ and $\psi_i''$
  ($i\in\{ 0,1 \}$) are equivalent, then they are essentially the same
  term, except that possibly one term may contain a subterm with a feature
  $\f$ pointing at the sort $\tops$, and this subterm is not present in the other
  equivalent terms (see also \cref{def:osf_graph_equivalence} and the proof
  of \cref{lemma:synsem}). The existence of such trivial subterms
  does not however affect the degree of the syntactic subsumption.

  Let us now prove the statement of the theorem. If $\fisop(\psi_0, \psi_1)
  = \beta$ then by \cref{lemma:synsem}
  there exist $\psi_0'\equiv\psi_0$ and $\psi_1'\equiv\psi_1$ such that
  $\fsynisop(\psi_0', \psi_1')\geq\beta$, i.e.,
  $\psi_0'\fsynisa_{\beta'}\psi_1'$ and $\beta'\geq\beta$.
  Then also $\fsynisop(\psi_0', \psi_1')\geq\beta'$,
  and thus by
  \cref{lemma:synsem} $\fisop(\psi_0, \psi_1)\geq\beta'$, i.e.,
  $\beta'\leq\beta$, and thus $\beta'=\beta$.

  Now suppose that
  there exist $\psi_0'\equiv\psi_0$ and $\psi_1'\equiv\psi_1$ such that
  $\psi_0'\fsynisa_{\beta}\psi_1'$, so that
  $\fsynisop(\psi_0', \psi_1')\geq\beta$.
  Then by \cref{lemma:synsem}
  $\fisop(\psi_0, \psi_1)\geq\beta$. To prove that
  $\fisop(\psi_0, \psi_1) \leq \beta$, let us suppose otherwise, i.e.,
  that $\fisop(\psi_0, \psi_1) = \beta' >\beta$. Then by \cref{lemma:synsem}
  there exist $\psi_0''\equiv\psi_0$ and $\psi_1''\equiv\psi_1$ such that
  $\fsynisop(\psi_0'', \psi_1'')\geq\beta'$, i.e.,
  $\psi_0''\fsynisa_{\beta''}\psi_1''$ and $\beta''\geq \beta' > \beta$.
  But by \cref{claim:syndegree} $\beta'' = \beta$, which is the desired
  contradiction.
\end{proof}

\thmcrispfuzzy*
\begin{proof}[Proof of \cref{thm:crispfuzzy}]%
  \hypertarget{proof:crispfuzzy}{\mbox{}}%
  We employ the following notation for the rest of the proof: for an \osf
  graph $g = \ograph$, we let $\s_g \defeq \lambda_N(\X)$, i.e., $\s_g$ is
  the label of the root of $g$.

  In the rest of the proof we rely on definitions, results and
    notation from crisp \osf logic \cite{AitKaci1993b}. In particular,
    $\isa$ denotes the crisp subsumption ordering on sorts and \osf terms,
    and $\approximates[\jj]$ is the crisp approximation ordering on \osf
  graphs.

  To better distinguish between the crisp and fuzzy contexts in the proof,
  we use $\jj$ for the \textit{crisp} \osf graph algebra
  \cite{AitKaci1993b}, and $\gg$ for the \textit{fuzzy} \osf graph algebra.
  Note that the domains of these two structures and the definition of
  $\f^\jj$ and $\f^\gg$ are the same. The only difference is the
  interpretation of sort symbols, since, for each $\s\in\S$:
    $\s^\jj$ is the set of graphs $g$ such that $\s_g\isa \s$,
    while $\s^\gg$ is a fuzzy set defined by letting $\s^\gg(g) =
      \fisop(\s_{g}, \s)$.
  Note that it follows that $g\in\s^\jj$ if and only if $\s^\gg(g) > 0$.

  Also note that, given a graph $g$, the subalgebra $\jj[g]$ and the fuzzy
  subalgebra $\gg[g]$ have the same domain and feature symbols are
  interpreted in the same way.
  The only difference is again the interpretation of sort symbols, since,
  for each $\s\in\S$:
    $\s^{\jj[g]} = \s^\jj\cap\F^*(g)$ and
    $\s^{\gg[g]} = \s^\gg\fcap\cf[\F^*(g)]$.
  Note that it holds that
  \ifallproofs
  \[
    \begin{array}{lll}
      g'\in\s^{\jj[g]} & \Iff & g'\in\s^\jj~\text{and}~g'\in\F^*(g) \\
                       & \Iff & \s^\gg(g')>0~\text{and}~\cf[\F^*(g)](g') = 1 \\
                       & \Iff & \s^\gg\fcap\cf[\F^*(g)](g')>0 \\
                       & \Iff & \s^{\gg[g]}(g') > 0.
    \end{array}
  \]
  \else
    $g'\in\s^{\jj[g]}$ if and only if $\s^{\gg[g]}(g') > 0$.
  \fi

  Let $g_1 = G(\psi_1)$ and $g_2 = G(\psi_2)$.
  Suppose that $\psi_1\isa\psi_2$, so that $g_2\approximates[{\jj}] g_1$
  \cite{AitKaci1993b}, meaning that
  there is a function $\gamma: \dom[{\jj[g_2]}]\to\dom[{\jj[g_1]}]$ such
  that
  \begin{itemize}
    \item $\gamma(g_2) = g_1$;
    \item $\forall \f\in\F$, $\forall g\in\jj[g_2]$:
                       $\gamma(\f^{\jj}(g)) = \f^{\jj}(\gamma(g))$; and
    \item $\forall \s\in\S$, $\forall g\in\jj[g_2]$:
      if $g\in\s^\jj$, then $\gamma(g)\in\s^\jj$.
  \end{itemize}
  Note that, equivalently, the last condition expresses that $\forall
  \s\in\S$, $\forall g\in\gg[g_2]$:
  \begin{align}
    \label{eq:crispifthen}
    \text{if}~\s^\gg(g)>0,~\text{then}~\s^\gg(\gamma(g))>0.
  \end{align}

  We prove that there is a $\beta> 0$ such that
  $\forall \s\in\S$ and $\forall g\in\gg[g_2]$: $\s^\gg(g)\land\beta\leq
  \s^\gg(\gamma(g))$.

  Note that $\gg[g_2]$ contains a finite number of graphs with root label
  different from $\tops$,
  i.e., at most the ones corresponding to the variables of
  $\phi(\psi_2)$.
  Thus the set
  $B \defeq \{ \fisop(\s_{\gamma(g)}, \s_{g})  \mid \s_{g}\neq\tops, g\in\gg[g_2] \}$
  is finite, so that $\beta \defeq \min(B)$ exists.
  Note that
    $\beta = \min(\{ \fisop(\s_{\gamma(g)}, \s_{g})  \mid
      g\in\gg[g_2] \})$,
      since if $\s_{g} = \tops$, then $\fisop(\s_{\gamma(g)},
      \s_{g}) = 1$ by definition.
      Moreover, for every $g\in\gg[g_2]$, $\fisop(\s_{\gamma(g)},
        \s_{g})>0$:
        indeed $\s^\gg_{g}(g) = \fisop(\s_g, \s_g) = 1 > 0$ so that by
        \cref{eq:crispifthen} $\s_{g}^\gg(\gamma(g)) = \fisop(\s_{\gamma(g)}, \s_{g})
        > 0$.
  It follows that $\beta > 0$. Finally, let $g\in\gg[g_2]$ and $\s\in \S$. Then
  \[
    \begin{array}{lllll}
      \s^\gg(g)\land\beta & = & \fisop(\s_{g}, \s) \land \beta
                  & \leq & \fisop(\s_{g}, \s)\land\fisop(\s_{\gamma(g)},
                  \s_{g})\\
                  & \leq & \fisop(\s_{\gamma(g)}, \s)
                  & = & \s^\gg(\gamma(g))
    \end{array}
  \]
  by transitivity of $\fisop$. Thus $\gamma$ is a $\beta$-morphism $\gamma:
  \gg[g_2]\to\gg[g_1]$ such that $\gamma(g_2) = g_1$. Then
  $\fapproximates(g_2, g_1) = \beta' \geq \beta > 0$, so that by
  \cref{thm:transparencybis} $\fisop(\psi_1, \psi_2) > 0$.

  For the other direction, assume that $\psi_1\fisa_\beta \psi_2$ for some
  $\beta>0$. Then $g_2 \fapprel g_1$, meaning that there is a
  function $\gamma: \dom[{\gg[g_2]}]\to\dom[{\gg[g_1]}]$ such that
  \begin{itemize}
    \item $\gamma(g_2) = g_1$;
    \item $\forall \f\in\F$, $\forall g\in\gg[g_2]$:
                       $\gamma(\f^{\gg}(g)) = \f^{\gg}(\gamma(g))$; and
    \item $\forall \s\in\S$, $\forall g\in\gg[g_2]$:
      $\s^\gg(g)\land\beta\leq \s^\gg(\gamma(g))$.
  \end{itemize}
  Let $\s$ and $g$ be arbitrary and suppose that
  $g\in\s^\jj$,
  so that $\s^\gg(g)>0$. Since
  $\s^\gg(g)\land\beta\leq \s^\gg(\gamma(g))$, then
  $\s^\gg(\gamma(g))>0$,
  and so
  $\gamma(g)\in\s^{\jj}$.
  Thus $\gamma$ is an
  \osf algebra morphism \cite{AitKaci1993b} $\gamma:\jj[g_2]\to\jj[g_1]$
  such that $\gamma(g_2) = g_1$, i.e., $g_2\fapproximates[\jj] g_1$, and
  equivalently $\psi_1\isa\psi_2$ as desired.
\end{proof}

\section{Proofs for \cref{sec:unification}: \nameref{sec:unification}}%
\label{proofs:unif}

\thmalgo*
\begin{proof}[Proof of \cref{thm:algo}]%
  \hypertarget{proof:algo}{\mbox{}}%
  The fact that $\psi = \psi_1\fmeet\psi_2$ is given by
  \cref{thm:fuzzy_osf_term_unification}. Let us assume that $\psi$ is not
  the inconsistent clause, as otherwise $\beta=1$ and the result follows
  immediately.

  For $i\in\{ 1, 2 \}$, let
  $h_i: \tags(\psi_i) \to\tags(\psi)$ be the
  mapping defined by letting
  $h_i(\X) = \Z_{[\X]}$ for all $\X\in\tags(\psi_i)$.
  Then $h_1$ witnesses the syntactic subsumption
  $\psi \fsynisa_{\beta_1} \psi_1$,
  as it satisfies the three conditions of
  \cref{def:syn_osf_term_subsumption}.
  \begin{enumerate}
    \item
      Let $\rtag(\psi_1) = \Xj[1]$.
      The fact that
      $h_1(\rtag(\psi_1)) =h_1(\Xj[1]) = \Z_{[\Xj[1]]} = \rtag(\psi)$
      is clear by the construction of $\psi = \psi(\phi')$ on \cref{line:13}
    and the fact that $\phi'$ is rooted in $\Z_{[\Xj[1]]}$.
    \item
      Suppose that $\X\fto_{\psi_1}\Y$. Then $\phi$ on \cref{line:2}
      contains the constraint $\X.\f \doteq \Y$. Throughout the application of
      the constraint normalization rules of \cref{fig:osf_normalization}
      the variables $\X$ and $\Y$ of this constraint may be substituted
      with other variables, possibly multiple times.
      All of these substitutions are witnessed by equality constraints
      contained in $\phi$ after the loop of
      \cref{line:3,line:4}.
      By the definition of $\osfeq$ on \cref{line:9},
      after the loop of \cref{line:11,line:12} the clause
      $\phi'$ must contain the constraint $\Z_{[\X]}.\f\to \Z_{[\Y]}$.
      Finally, by the construction of $\psi = \psi(\phi')$ on
      \cref{line:13}, it holds that
      $\Z_{[\X]}\fto_{\psi}\Z_{[\Y]}$, i.e.,
      $h_1(\X)\fto_{\psi}h_1(\Y)$, as required.
    \item
      Since $h_1(\X) = \Z_{[\X]}$ for all $\X\in\tags(\psi_1)$, the
      computation of $\beta_1$ on \cref{line:14} is carried out in the same
      way as the definition of the syntactic subsumption degree in
      \cref{def:syn_osf_term_subsumption}.
  \end{enumerate}
  It can be proved analogously that $h_2$ witnesses the syntactic subsumption
  $\psi \fsynisa_{\beta_2} \psi_2$.
  It follows by \cref{prop:synsem} that
  $\fisop(\psi, \psi_1) = \beta_1$
  and
  $\fisop(\psi, \psi_2) = \beta_2$, and, by
  \cref{def:fuzzy_osf_term_unification}, that $\beta=\min(\beta_1,
  \beta_2)$ is the unification degree of $\psi_1$ and $\psi_2$.
\end{proof}

\bibliography{abrv,bibliography}

\begin{thebibliography}{32}
\expandafter\ifx\csname natexlab\endcsname\relax\def\natexlab#1{#1}\fi
\providecommand{\url}[1]{\texttt{#1}}
\providecommand{\href}[2]{#2}
\providecommand{\path}[1]{#1}
\providecommand{\DOIprefix}{doi:}
\providecommand{\ArXivprefix}{arXiv:}
\providecommand{\URLprefix}{URL: }
\providecommand{\Pubmedprefix}{pmid:}
\providecommand{\doi}[1]{\href{http://dx.doi.org/#1}{\path{#1}}}
\providecommand{\Pubmed}[1]{\href{pmid:#1}{\path{#1}}}
\providecommand{\bibinfo}[2]{#2}
\ifx\xfnm\relax \def\xfnm[#1]{\unskip,\space#1}\fi
\bibitem[{A{\"i}t-Kaci(1984)}]{AitKaci1984}
\bibinfo{author}{H.~A{\"i}t-Kaci}, \bibinfo{title}{A Lattice Theoretic Approach
  to Computation Based on a Calculus of Partially Ordered Type Structures},
  Ph.D. thesis, \bibinfo{address}{University of Pennsylvania},
  \bibinfo{year}{1984}.
\bibitem[{Baader et~al.(2008)Baader, Horrocks, and Sattler}]{DLKRHB}
\bibinfo{author}{F.~Baader}, \bibinfo{author}{I.~Horrocks},
  \bibinfo{author}{U.~Sattler},
\newblock \bibinfo{title}{Description logics},
\newblock in: \bibinfo{editor}{F.~{van Harmelen}},
  \bibinfo{editor}{V.~Lifschitz}, \bibinfo{editor}{B.~Porter} (Eds.),
  \bibinfo{booktitle}{Handbook of Knowledge Representation},
  volume~\bibinfo{volume}{3} of \textit{\bibinfo{series}{Foundations of
  Artificial Intelligence}}, \bibinfo{publisher}{Elsevier},
  \bibinfo{year}{2008}, pp. \bibinfo{pages}{135--179}. \URLprefix
  \url{http://www.sciencedirect.com/science/article/pii/S1574652607030039}.
  \DOIprefix\doi{https://doi.org/10.1016/S1574-6526(07)03003-9}.
\bibitem[{A{\"i}t-Kaci(2007)}]{AitKaci2007b}
\bibinfo{author}{H.~A{\"i}t-Kaci},
\newblock \bibinfo{title}{Data models as constraint systems: a key to the
  semantic web},
\newblock \bibinfo{journal}{Constraint Processing Letters} \bibinfo{volume}{1}
  (\bibinfo{year}{2007}).
\bibitem[{Smolka(1988)}]{Smolka1988}
\bibinfo{author}{G.~Smolka},
\newblock \bibinfo{title}{A feature logic with subsorts},
\newblock \bibinfo{journal}{LILOG-Report} \bibinfo{volume}{33}
  (\bibinfo{year}{1988}).
\bibitem[{Carpenter(1992)}]{Carpenter1992}
\bibinfo{author}{R.~L. Carpenter}, \bibinfo{title}{The Logic of Typed Feature
  Structures: With Applications to Unification Grammars, Logic Programs and
  Constraint Resolution}, Cambridge Tracts in Theoretical Computer Science,
  \bibinfo{publisher}{Cambridge University Press}, \bibinfo{year}{1992}.
  \DOIprefix\doi{10.1017/CBO9780511530098}.
\bibitem[{A{\"i}t-Kaci and Nasr(1986)}]{AitKaci1986b}
\bibinfo{author}{H.~A{\"i}t-Kaci}, \bibinfo{author}{R.~Nasr},
\newblock \bibinfo{title}{Login: a logic programming language with built-in
  inheritance},
\newblock \bibinfo{journal}{J. Log. Program.} \bibinfo{volume}{3}
  (\bibinfo{year}{1986}) \bibinfo{pages}{185--215}. \URLprefix
  \url{http://www.sciencedirect.com/science/article/pii/0743106686900130}.
  \DOIprefix\doi{https://doi.org/10.1016/0743-1066(86)90013-0}.
\bibitem[{A{\"i}t-Kaci and Podelski(1993)}]{AitKaci1993b}
\bibinfo{author}{H.~A{\"i}t-Kaci}, \bibinfo{author}{A.~Podelski},
\newblock \bibinfo{title}{Towards a meaning of life},
\newblock \bibinfo{journal}{J. Log. Program.} \bibinfo{volume}{16}
  (\bibinfo{year}{1993}) \bibinfo{pages}{195--234}. \URLprefix
  \url{http://www.sciencedirect.com/science/article/pii/074310669390043G}.
  \DOIprefix\doi{https://doi.org/10.1016/0743-1066(93)90043-G}.
\bibitem[{Mukai(1987)}]{Mukai1987}
\bibinfo{author}{K.~Mukai}, \bibinfo{title}{Anadic tuples in Prolog},
  \bibinfo{type}{Technical Report} \bibinfo{number}{TR-239}, ICOT,
  \bibinfo{address}{Tokyo, Japan}, \bibinfo{year}{1987}.
\bibitem[{A{\"i}t-Kaci and Amir(2017)}]{AitKaciAmir2017}
\bibinfo{author}{H.~A{\"i}t-Kaci}, \bibinfo{author}{S.~Amir},
\newblock \bibinfo{title}{Classifying and querying very large taxonomies with
  bit-vector encoding},
\newblock \bibinfo{journal}{J. Intell. Inf. Syst.} \bibinfo{volume}{48}
  (\bibinfo{year}{2017}) \bibinfo{pages}{1--25}. \URLprefix
  \url{https://doi.org/10.1007/s10844-015-0383-2}.
  \DOIprefix\doi{10.1007/s10844-015-0383-2}.
\bibitem[{Amir and A{\"i}t-Kaci(2017)}]{AmirAitKaci2017}
\bibinfo{author}{S.~Amir}, \bibinfo{author}{H.~A{\"i}t-Kaci},
\newblock \bibinfo{title}{An efficient and large-scale reasoning method for the
  semantic web},
\newblock \bibinfo{journal}{J. Intell. Inf. Syst.} \bibinfo{volume}{48}
  (\bibinfo{year}{2017}) \bibinfo{pages}{653--674}. \URLprefix
  \url{https://doi.org/10.1007/s10844-016-0435-2}.
  \DOIprefix\doi{10.1007/s10844-016-0435-2}.
\bibitem[{A{\"i}t-Kaci et~al.(1989)A{\"i}t-Kaci, Boyer, Lincoln, and
  Nasr}]{AitKaci1989}
\bibinfo{author}{H.~A{\"i}t-Kaci}, \bibinfo{author}{R.~Boyer},
  \bibinfo{author}{P.~Lincoln}, \bibinfo{author}{R.~Nasr},
\newblock \bibinfo{title}{Efficient implementation of lattice operations},
\newblock \bibinfo{journal}{ACM Trans. Program. Lang. Syst.}
  \bibinfo{volume}{11} (\bibinfo{year}{1989}) \bibinfo{pages}{115--146}.
  \URLprefix \url{https://doi.org/10.1145/59287.59293}.
  \DOIprefix\doi{10.1145/59287.59293}.
\bibitem[{A{\"i}t-Kaci and Pasi(2020)}]{AitKaciPasi2020}
\bibinfo{author}{H.~A{\"i}t-Kaci}, \bibinfo{author}{G.~Pasi},
\newblock \bibinfo{title}{Fuzzy lattice operations on first-order terms over
  signatures with similar constructors: A constraint-based approach},
\newblock \bibinfo{journal}{Fuzzy Sets Syst.} \bibinfo{volume}{391}
  (\bibinfo{year}{2020}) \bibinfo{pages}{1--46}. \URLprefix
  \url{http://www.sciencedirect.com/science/article/pii/S016501141930199X}.
  \DOIprefix\doi{https://doi.org/10.1016/j.fss.2019.03.019},
  \bibinfo{note}{computer Science}.
\bibitem[{Straccia(2014)}]{Straccia2014}
\bibinfo{author}{U.~Straccia}, \bibinfo{title}{Foundations of Fuzzy Logic and
  Semantic Web Languages}, \bibinfo{publisher}{Chapman and Hall},
  \bibinfo{address}{New York}, \bibinfo{year}{2014}.
  \DOIprefix\doi{https://doi.org/10.1201/b15460}.
\bibitem[{Borgwardt and Pe{\~n}aloza(2017)}]{Borgwardt2017}
\bibinfo{author}{S.~Borgwardt}, \bibinfo{author}{R.~Pe{\~n}aloza},
\newblock \bibinfo{title}{Fuzzy description logics--a survey},
\newblock in: \bibinfo{booktitle}{International Conference on Scalable
  Uncertainty Management}, \bibinfo{organization}{Springer},
  \bibinfo{year}{2017}, pp. \bibinfo{pages}{31--45}.
\bibitem[{Dubois and Prade(1980)}]{DuboisPrade1980}
\bibinfo{author}{D.~Dubois}, \bibinfo{author}{H.~Prade}, \bibinfo{title}{Fuzzy
  Sets and Systems. Theory and Applications.}, volume \bibinfo{volume}{144} of
  \textit{\bibinfo{series}{Mathematics in Science and Engineering}},
  \bibinfo{publisher}{Elsevier}, \bibinfo{year}{1980}. \URLprefix
  \url{https://www.sciencedirect.com/bookseries/mathematics-in-science-and-engineering/vol/144/}.
\bibitem[{Arcelli-Fontana and Formato(2002)}]{Arcelli2002}
\bibinfo{author}{F.~Arcelli-Fontana}, \bibinfo{author}{F.~Formato},
\newblock \bibinfo{title}{A similarity-\-based resolution rule},
\newblock \bibinfo{journal}{International Journal of Intelligent Systems}
  \bibinfo{volume}{17} (\bibinfo{year}{2002}) \bibinfo{pages}{853--872}.
\bibitem[{Gerla and Sessa(1999)}]{Gerla1999}
\bibinfo{author}{G.~Gerla}, \bibinfo{author}{M.~I. Sessa},
  \bibinfo{title}{Similarity in Logic Programming},
  \bibinfo{publisher}{Springer US}, \bibinfo{address}{Boston, MA},
  \bibinfo{year}{1999}, pp. \bibinfo{pages}{19--31}. \URLprefix
  \url{https://doi.org/10.1007/978-1-4615-5261-1_2}.
  \DOIprefix\doi{10.1007/978-1-4615-5261-1_2}.
\bibitem[{Sessa(2002)}]{Sessa2002}
\bibinfo{author}{M.~I. Sessa},
\newblock \bibinfo{title}{Approximate reasoning by similarity-based {SLD}
  resolution},
\newblock \bibinfo{journal}{Theor. Comput. Sci.} \bibinfo{volume}{275}
  (\bibinfo{year}{2002}) \bibinfo{pages}{389--426}. \URLprefix
  \url{http://www.sciencedirect.com/science/article/pii/S0304397501001888}.
  \DOIprefix\doi{https://doi.org/10.1016/S0304-3975(01)00188-8}.
\bibitem[{Juli{\'a}n-Iranzo and Rubio-Manzano(2015)}]{Iranzo2015b}
\bibinfo{author}{P.~Juli{\'a}n-Iranzo}, \bibinfo{author}{C.~Rubio-Manzano},
\newblock \bibinfo{title}{Proximity-based unification theory},
\newblock \bibinfo{journal}{Fuzzy Sets Syst.} \bibinfo{volume}{262}
  (\bibinfo{year}{2015}) \bibinfo{pages}{21 -- 43}. \URLprefix
  \url{http://www.sciencedirect.com/science/article/pii/S0165011414003133}.
  \DOIprefix\doi{https://doi.org/10.1016/j.fss.2014.07.006},
  \bibinfo{note}{theme: Logic and Computer Science}.
\bibitem[{Juli{\'a}n-Iranzo and S{\'a}enz-P{\'e}rez(2020)}]{Iranzo2020}
\bibinfo{author}{P.~Juli{\'a}n-Iranzo},
  \bibinfo{author}{F.~S{\'a}enz-P{\'e}rez},
\newblock \bibinfo{title}{Proximity-based unification: an efficient
  implementation method},
\newblock \bibinfo{journal}{IEEE Trans. Fuzzy Syst.} \bibinfo{volume}{29}
  (\bibinfo{year}{2020}) \bibinfo{pages}{1238--1251}.
\bibitem[{Kutsia and Pau(2020)}]{Kutsia2020}
\bibinfo{author}{T.~Kutsia}, \bibinfo{author}{C.~Pau},
\newblock \bibinfo{title}{Solving proximity constraints},
\newblock in: \bibinfo{editor}{M.~Gabbrielli} (Ed.),
  \bibinfo{booktitle}{Logic-Based Program Synthesis and Transformation},
  \bibinfo{publisher}{Springer International Publishing},
  \bibinfo{address}{Cham}, \bibinfo{year}{2020}, pp. \bibinfo{pages}{107--122}.
\bibitem[{Julián-Iranzo and Sáenz-Pérez(2023)}]{Iranzo2023}
\bibinfo{author}{P.~Julián-Iranzo}, \bibinfo{author}{F.~Sáenz-Pérez},
\newblock \bibinfo{title}{{Bousi{$\sim$}Prolog}: Design and implementation of a
  proximity-based fuzzy logic programming language},
\newblock \bibinfo{journal}{Expert Syst. Appl.} \bibinfo{volume}{213}
  (\bibinfo{year}{2023}) \bibinfo{pages}{118858}. \URLprefix
  \url{https://www.sciencedirect.com/science/article/pii/S0957417422018760}.
  \DOIprefix\doi{https://doi.org/10.1016/j.eswa.2022.118858}.
\bibitem[{Juli{\'a}n-Iranzo et~al.(2020)Juli{\'a}n-Iranzo, Moreno, and
  Riaza}]{Iranzo2020fasill}
\bibinfo{author}{P.~Juli{\'a}n-Iranzo}, \bibinfo{author}{G.~Moreno},
  \bibinfo{author}{J.~A. Riaza},
\newblock \bibinfo{title}{The fuzzy logic programming language {FASILL}: Design
  and implementation},
\newblock \bibinfo{journal}{Int. J. Approx. Reason.} \bibinfo{volume}{125}
  (\bibinfo{year}{2020}) \bibinfo{pages}{139--168}.
\bibitem[{Cohn(1989)}]{Cohn1989}
\bibinfo{author}{A.~G. Cohn},
\newblock \bibinfo{title}{Taxonomic reasoning with many-sorted logics},
\newblock \bibinfo{journal}{Artif. Intell. Rev.} \bibinfo{volume}{3}
  (\bibinfo{year}{1989}) \bibinfo{pages}{89--128}.
\bibitem[{Reynolds(1970)}]{Reynolds1970}
\bibinfo{author}{J.~C. Reynolds},
\newblock \bibinfo{title}{Transformational systems and algebraic structure of
  atomic formulas},
\newblock \bibinfo{journal}{Mach. Intell.} \bibinfo{volume}{5}
  (\bibinfo{year}{1970}) \bibinfo{pages}{135--151}.
\bibitem[{Plotkin(1969)}]{Plotkin1969}
\bibinfo{author}{G.~D. Plotkin},
\newblock \bibinfo{title}{A note on inductive generalization},
\newblock \bibinfo{journal}{Mach. Intell.} \bibinfo{volume}{5}
  (\bibinfo{year}{1969}) \bibinfo{pages}{153--163}.
\bibitem[{Milanese and Pasi(2022)}]{Milanese2022}
\bibinfo{author}{G.~C. Milanese}, \bibinfo{author}{G.~Pasi},
\newblock \bibinfo{title}{{Fuzzy Order-Sorted Feature Term Unification}},
\newblock in: \bibinfo{booktitle}{{36th International Workshop on
  Unification}}, \bibinfo{year}{2022}. \URLprefix
  \url{https://www.cs.cas.cz/unif-2022/Papers/UNIF_2022_paper_3.pdf}.
\bibitem[{Aho and Hopcroft(1974)}]{Aho1974}
\bibinfo{author}{A.~V. Aho}, \bibinfo{author}{J.~E. Hopcroft},
  \bibinfo{title}{The design and analysis of computer algorithms},
  \bibinfo{publisher}{Pearson Education India}, \bibinfo{year}{1974}.
\bibitem[{Milanese and Pasi(2021)}]{Milanese2021a}
\bibinfo{author}{G.~C. Milanese}, \bibinfo{author}{G.~Pasi},
\newblock \bibinfo{title}{Conjunctive reasoning on fuzzy taxonomies with
  order-sorted feature logic},
\newblock in: \bibinfo{booktitle}{2021 IEEE International Conference on Fuzzy
  Systems (FUZZ-IEEE)}, \bibinfo{organization}{IEEE}, \bibinfo{year}{2021}, pp.
  \bibinfo{pages}{1--7}.
\bibitem[{Chon(2009)}]{Chon2009}
\bibinfo{author}{I.~Chon},
\newblock \bibinfo{title}{Fuzzy partial order relations and fuzzy lattices},
\newblock \bibinfo{journal}{Korean J. Math.} \bibinfo{volume}{17}
  (\bibinfo{year}{2009}) \bibinfo{pages}{361--374}.
\bibitem[{{Mezzomo} et~al.(2013){Mezzomo}, {Bedregal}, and
  {Santiago}}]{Mezzomo2013}
\bibinfo{author}{I.~{Mezzomo}}, \bibinfo{author}{B.~{Bedregal}},
  \bibinfo{author}{R.~H.~N. {Santiago}},
\newblock \bibinfo{title}{Operations on bounded fuzzy lattices},
\newblock in: \bibinfo{booktitle}{2013 Joint IFSA World Congress and NAFIPS
  Annual Meeting (IFSA/NAFIPS)}, \bibinfo{year}{2013}, pp.
  \bibinfo{pages}{151--156}. \DOIprefix\doi{10.1109/IFSA-NAFIPS.2013.6608391}.
\bibitem[{{Mezzomo} and {Bedregal}(2016)}]{Mezzomo2016}
\bibinfo{author}{I.~{Mezzomo}}, \bibinfo{author}{B.~C. {Bedregal}},
\newblock \bibinfo{title}{On fuzzy {$\alpha$}-lattices},
\newblock in: \bibinfo{booktitle}{2016 IEEE International Conference on Fuzzy
  Systems (FUZZ-IEEE)}, \bibinfo{year}{2016}, pp. \bibinfo{pages}{775--781}.
  \DOIprefix\doi{10.1109/FUZZ-IEEE.2016.7737766}.

\end{thebibliography}

\end{document}